\documentclass[final,leqno,onefignum,onetabnum]{siamltexmm}

\usepackage{graphics}
\usepackage{graphicx}
\usepackage{amsfonts}
\usepackage{amssymb}
\usepackage{amsmath}
\usepackage{color}
\usepackage{hyperref}
\usepackage{caption}
\usepackage{subcaption}
\usepackage{epsfig}

\newcommand{\bx}{\mathbf{x}}
\newcommand{\by}{\mathbf{y}}

\newcommand{\bff}{\mathbf{f}}
\newcommand{\bg}{\mathbf{g}}

\newcommand{\bu}{\mathbf{u}}

\newcommand{\bp}{\mathbf{p}}

\newcommand{\bd}{\mathbf{d}}
\newcommand{\bn}{\mathbf{n}}

\newcommand{\bI}{\mathbf{I}}

\newcommand{\bF}{\mathbf{F}}

\newcommand{\bSigma}{\boldsymbol{\Sigma}}
\newcommand{\bxi}{\boldsymbol{\xi}}
\newcommand{\bk}{\boldsymbol{\kappa}}

\newcommand{\bomega}{\boldsymbol{\omega}}
\newcommand{\btheta}{\boldsymbol{\theta}}

\newcommand{\balpha}{\boldsymbol{\alpha}}
\newcommand{\bbeta}{\boldsymbol{\beta}}
\newcommand{\bDelta}{\boldsymbol{\Delta}}

\newcommand{\calG}{\mathcal{G}}
\newcommand{\calD}{\mathcal{D}}
\newcommand{\calN}{\mathcal{N}}

\newcommand{\calF}{\mathcal{F}}
\newcommand{\calH}{\mathcal{H}}

\newcommand{\calY}{\mathcal{Y}}
\newcommand{\calK}{\mathcal{K}}
\newcommand{\calU}{\mathcal{U}}

\newcommand{\R}{\mathbb{R}}
\newcommand{\N}{\mathbb{N}}
\newcommand{\E}{\mathbb{E}}

\newcommand{\nothree}[1]{\left[\operatorname{NO}_3^-\right]}

\title{Efficient Bayesian experimentation using an expected information
  gain lower bound} 

\author{Panagiotis Tsilifis\thanks{Department of Mathematics,
    University of Southern California, Los Angeles, CA 90089
    (\email{tsilifis@usc.edu}).} \and Roger
G. Ghanem\thanks{Department of Civil Engineering, University of
  Southern California, Los Angeles, CA 90089 (\email{ghanem@usc.edu})}
  \and Paris Hajali\thanks{MUREX Environmental Inc., 15375 Barranca
    Parkway, Suite K-101, Irvine, CA 92618 (\email{parishajali@murexenv.com})}}

\begin{document}
\maketitle
\newcommand{\slugmaster}{%
\slugger{juq}{xxxx}{xx}{x}{x--x}}

\begin{abstract}
Experimental design is crucial for inference where limitations in
the data collection procedure are present due to cost or other restrictions. Optimal
experimental designs determine parameters that in some appropriate
sense make the data the most informative possible. In a Bayesian
setting this is translated to updating to the \emph{best} possible
posterior. Information theoretic arguments have led to the formation
of the expected information gain as a design criterion. This can
be evaluated mainly by Monte Carlo sampling and maximized by using
stochastic approximation methods, both known for being computationally
expensive tasks. We propose a framework where a lower
bound of the expected information gain is used as an alternative design
criterion.  In addition to alleviating the computational burden, this
also addresses issues concerning estimation bias. 
The problem of permeability inference in a large contaminated area is
used to demonstrate the validity of our
approach where we employ the massively parallel version of the multiphase
multicomponent simulator TOUGH2 to simulate contaminant transport and
a Polynomial Chaos approximation of the forward model that further
accelerates the objective function evaluations. 
The proposed methodology is demonstrated to a setting where field measurements are 
available.
\end{abstract}



\pagestyle{myheadings}
\thispagestyle{plain}
\markboth {PANAGIOTIS TSILIFIS, ROGER G. GHANEM AND PARIS HAJALI}{EFFICIENT BAYESIAN EXPERIMENTATION}

\section{Introduction}

Remediation of a polluted subsurface presents an increasingly common
need in most urban areas undergoing accelerated expansions.  Risks
associated with these pollutants range from health consequences to
financial costs of both monitoring, remediation and insurance.  The
fact that the subsurface can never be completely characterized 
is a key contributor to these risks, and credible procedures for
improving this characterization have far reaching consequences
across the spectrum of constituency.  What is ultimately needed is
a {\it sufficient} assessment of the subsurface which depends on the
physics of subsurface flow and a knowledge of the initial conditions,
and an assessment of the flow functionals that are relevant to risk
assessment.  As a first step, in the present paper, we address the
issue of optimal subsurface characterization under conditions of
limited resources. 

A vast amount of research works have been dedicated to this challenge
and have offered application-specific answers throughout the years with
a Bayesian decision analysis framework often being present. Among the
earliest ones, the works by \cite{freeze,james}, discussed the worth of
correlated hydraulic conductivity
measurements that was evaluated depending on the revenue due to the resulting
uncertainty reduction and the cost of obtaining them. In the excellent
work by \cite{gorelick}, a sequential sampling methodology was developed and
the issues of where to obtain the next sample and when to cease
sampling were addressed using an analogous worth-of-information
framework.  More recently \cite{Haddad-2013}, a Kullback-Liebler
formalism that maximizes the worth of information of an experiment was
developed, and the worth of additional observations
characterized. While that work ignored the physics of flow in porous
media by developing a kriging model for the measured concentrations in
the subsurface, it does lay the foundations for the present work.
A common characteristic in all the above works and many that
followed, is that the data-worth analyses,
Bayesian in nature, are divided in three main phases: the prior, the
preposterior and the posterior analyses with the first and the last
being the well known steps of any Bayesian method and the second being
the phase where the decisions about the design parameters are to be
made. This typically involves the suggestion of a utility function,
expressed often in monetary terms that quantifies the worth of
collecting some specific data and then evaluating the average worth by
taking the expectation of this
function. The above mentioned works and numerous others
\cite{massmann, norberg, bau} verified
a general conclusion: that the location or any other design parameters characterizing
the best samples may depend
on the model uncertainty but also on the decision to be made with the
latter factor usually adding an economic flavor in the data
worth approach. Although in most
real-world problems concerning the geosciences community, this might
be an efficient way to address such issues, it is of critical
importance, from a mathematical perspective, to be able to distinguish
the role of model uncertainty in optimal designs and not much attention has been received on this
direction. 

It is the purpose of this paper to present an approach to the optimal
design problems arising in contaminant transport applications that
meets each problem's objectives from an information theoretic
perspective, that is to explore the use of other design criteria as
candidates in the preposterior phase of any data-worth framework,
that provide maximum information about model uncertainty without being
concerned about economic costs, but still providing low risk results
for decision making. An extensive statistical literature is
available on experimental design methodologies for linear
regression models summarized in \cite{atkinson, fedorov} with criteria stated as
functionals of the Fisher information matrix. Bayesian
counterparts of these criteria as well as additional design criteria
for nonlinear models have also been suggested (see \cite{chaloner} for
a review). Typically these are taken as the expectation of some
appropriately chosen utility
functions or approximations of them. 

\begin{figure}
\begin{center}
\includegraphics[width = 7.5cm]{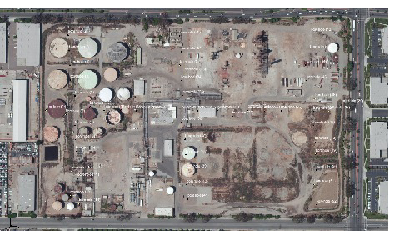}
\includegraphics[width = 8cm]{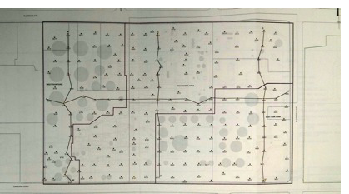}
\caption{Aerial photo showing the location of storage tanks (left) and
  soil boring map (right). \label{fig:world_map}}
\end{center}
\end{figure}

\begin{figure}
\centering
\includegraphics[width = 0.55\textwidth]{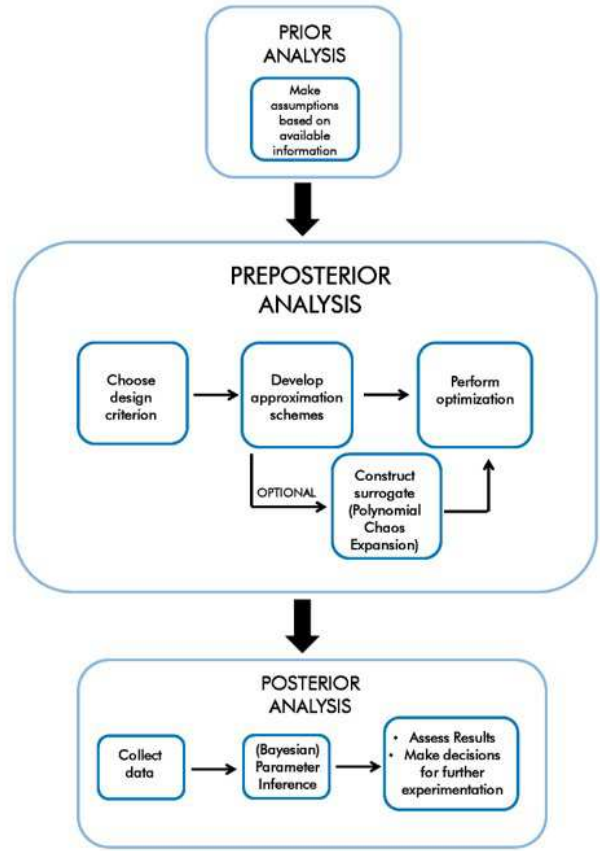}
\caption{A flowchart outlining the steps in prior, preposterior and
 posterior stages of the optimal design methodology.\label{fig:flowchart}}
\end{figure}

This is the direction followed in
this work. We choose a design criterion for data collection procedure
according to an objective function that maximizes information gain
about the uncertain parameters.  As a
natural consequence of dealing with realistic, complex and often
high-dimensional non-linear models, computationally intensive techniques such as the use
of model surrogates and Monte Carlo methods are involved in order to
evaluate and optimize the design criterion. The Bayesian parameter inference results that are
obtained from such optimal designs can then provide a rational basis for
decision making.  
All the above mentioned steps can be summarized as shown in the flowchart
in Fig.~\ref{fig:flowchart} .
The methodology developed in this paper is
demonstrated on an actual site shown in Fig.~\ref{fig:world_map}.
Soil-type data was collected at this site at various depth at boreholes located
along the solid lines that criss-cross the site, as shown in the soil boring map in Fig.~\ref{fig:world_map}.  Concentration values
were collected at the black dots throughout the site.
With reference to this specific case study, the sequence of action
that we investigate consists of the following steps:
\begin{enumerate}
\item Initial spills are assumed to have occurred at the location of
  the storage tanks shown in Figure \ref{fig:world_map}.
\item Mean and variances of permeability values are associated to soil types
  throughout the site, resulting in lognormal initial probabilistic models of
  permeabilities.
\item Physics of flow through porous media yield a corresponding
  distribution for the concentration of pollutants throughout the
  site.
\item For a given monitoring design consisting of a set of specific
  five spatial locations where concentrations are to be observed, a
  distribution of these observables can be obtained from the previous
  step.
\item From this distribution of observables, the likelihood for each
  possible value of observable is calculated, and the Bayesian posterior
  distribution for that observable is evaluated. Note that at this
  step, no concentration data is assumed to have been collected yet.
\item The Kullback-Leibler (KL) distance between the prior and the posterior
  distributions is evaluated as a measure of information gain.
\item The KL distance is integrated over all possible values attained
  by the observables at the fixed design locations.
\item The optimal monitoring locations are determined by maximizing
  this averaged KL distance.
\item Measurements of pollutants are taken from the actual field data at the
  optimal locations.
\item The Bayesian posterior distributions for permeabilities are calculated based
  on these observations. These posteriors provide an updated
  description of the permeabilities throughout the site and can be used
  to improve the prediction of concentration maps over time. 
\end{enumerate}
While a stopping criterion for this Bayesian update has been
introduced elsewhere \cite{Haddad-2013}, it is outside the scope of
the present investigation.  The paper addresses computational and
algorithmic issues required for the implementation of the above steps
and provides insight into the computed optimal monitoring strategies.
and is organized as follows. In Section~\ref{sec:exp_design} we
introduce the design criterion used for experimental design
that provides optimal Bayesian inference results and a lower bound that
is used as an alternative in the optimization procedure. The stochastic approximation
framework required for maximizing the objective function, is described in
Section~\ref{sec:stoch_opt}. Section~\ref{sec:toy_example} provides a toy
example in order to justify the use of the lower bound estimate instead of
the actual expected information gain as the design criterion. Then our
methodology is applied on a real site in
Section~\ref{sec:case_study}. Precisely, the model is described in subsections
\ref{sec:description} and \ref{sec:tough2}, then a
surrogate is constructed in subsection \ref{sec:surrogate} in order to
accelerate simulations of computationally intensive procedures. The
experimental design analysis is performed in subsection
\ref{sec:design} and the results are validated in subsection
\ref{sec:inference} by estimating the Bayesian update of the uncertain
parameters based on data that is generated both for a
hypothetical scenario and from field measurements. Our conclusions are
summarized in Section~\ref{sec:conclusions}.

\section{Bayesian experimental design}
\label{sec:exp_design}

\subsection{Design criterion}

Our main interest in the present work is to define and evaluate a
particular experiment which, constrained by fixed resources, will reduce the
prediction uncertainty in some well-defined optimal sense.
Let $\bd$ denote what we refer to as the set of design parameters.
For each choice of these parameters, the experiment design is fixed.
Thus $\bd$ could, for instance, consist of smoothing kernels that
parameterize measurement instruments. Alternatively, as we do in the
present work, we construe $\bd \in \mathbb{R}^{2m}$ as a vector that
defines the spatial coordinates in the horizontal plane of $m$ points
at which observations will be obtained.  Clearly, $\bd$
could also consist of nonlinear functionals. The numerical values of the
observations will be denoted by $\by\in\mathbb{R}^m$.  Prior to
conducting the experiments, $\by$ is a random vector.  Following the
experiment, the numerical values attained by this vector will be used
to condition the inference.  We make the restrictive assumption that
optimal reduction of uncertainty in model-based prediction is attained
through an optimal reduction of uncertainty in model parameters which
we denote by $\btheta$.  Relaxing this
assumption, while numerically tedious, does not present conceptual
difficulties.

\subsubsection{Expected information gain}

We are thus interested in inferring the unknown parameters $\btheta$ that
govern the behavior of a physical process.  We model these parameters
as a vector of random variables.  In a Bayesian setting, this
inference is carried out by updating the prior distribution $p(\btheta)$
with a posterior one, namely $p(\btheta | \by, \bd)$, given that some
specific data $\by$ was observed for particular design parameters
$\bd$. The posterior distribution is updated according to the rule 
\begin{equation}
\label{eg:bayes}
p(\btheta|\by, \bd) =  \frac{p(\by | \btheta, \bd)p(\btheta)}{p(\by| \bd)}
\end{equation}
where the multi-dimensional integral, $p(\by) = \int p(\by|\btheta,
\bd)p(\btheta)d\btheta$, is referred to as the \emph{evidence}.  Using a Shannon
information approach we can define the information gain after updating
the distribution of $\btheta$ to be the Kullback-Leibler (KL) divergence
\cite{kullback} of the prior and the posterior distributions, that is  
\begin{equation}
D_{KL} \left[ p(\cdot | \by, \bd) || p(\cdot) \right] = \int_{\Theta} p(\btheta |
    \by, \bd) \log \left[ \frac{p(\btheta | \by, \bd)}{p(\btheta)} \right] d\btheta.
\end{equation}
We are interested in quantifying the information gain before the data
is actually collected. In the same spirit we define the expected
information gain, first proposed by \cite{lindley} as 
\begin{equation}
\label{eq:eig}
U(\bd) = \int_{\calY} D_{KL} \left[ p(\cdot | \by, \bd) || p(\cdot) \right]
p(\by | \bd)
d\by = \int_{\calY}
\int_{\Theta} p(\btheta | \by, \bd) \log \left[ \frac{p(\btheta | \by,
    \bd)}{p(\btheta)}\right] d\btheta\  p(\by | \bd) d\by
\end{equation}
This can be interpreted as follows: The contribution of any possible output that
is used as a set of data
$\by$ to update to the posterior distribution is given as the KL
divergence of the two distributions. Before the data is collected,
$U(\bd)$ provides us with a measure of the average information to be
gained. This is a function only of the design parameters $\bd$. It is therefore
natural to assume that the choice of
$\bd^*$ that offers \emph{on average} the most informative
observations and thus, is the optimal experimental design, 
is the one that maximizes $U(\bd)$ and so it satisfies
\begin{equation}
\bd^* = \arg \max_{\bd \in \calD} U(\bd).
\end{equation} 

Evaluation of the above objective function and therefore its optimization is not a trivial
task. At first, one can see that the posterior, being unknown
beforehand, needs to be evaluated or approximated. In \cite{long_quan}, the
posterior was replaced by its Laplace approximation and $U(\bd)$ was
estimated with a sparse quadrature rule. In \cite{ryan} an alternative form
of the objective function was derived by using Bayes' rule to express the
posterior in terms of the evidence, the likelihood and the prior
distributions and estimates were proposed via Monte Carlo
methods. In both approaches, the maxima were identified by using an
exhaustive grid search over the whole design space
and limitations due to computational
expense were reported. Evaluation of the design criterion on all
points of the design space can easily become infeasible in applications where
either higher dimensional design parameters are involved or an
expensive forward solver is incorporated, hence the need for iterative
search strategies is necessary to detect the optimal value. In \cite{marzouk_simul,
  marzouk_gradient} it was demonstrated that stochastic
approximation methods \cite{spall_alg} are well adapted to the present situation, when
the Monte Carlo estimate of the objective is 
used and this is the approach we follow in this paper. However, instead of
simply adapting the methodology developed in these works, we further
derive a lower bound of the expected information gain and its
corresponding Monte Carlo estimate to be maximized. The reason for
doing so is to overcome computational obstacles that arise from our
application: The direct Monte Carlo estimator entails a discretization
of the double integral appearing in equation
(\ref{eq:eig}) and has
been shown to have a bias that is inversely proportional to
the number of samples in the inner sum of the estimate, requiring a
very large number of inner loop samples for the bias to be
negligible \cite{ryan} (see also Appendix B in \cite{marzouk_simul}
for a numerical study). Furthermore, the variance
of that estimator is controlled only by the number of the outer loop
samples. Expressions for both the bias and variance are given in \ref{sec:appendix_a}. For applications such as the one included in this 
paper, using tens or hundreds of thousands samples can easily become prohibitive.  On the contrary, the lower bound derived below can be easily
seen to be unbiased and its variance is controlled by the product of
the number of samples used in both loops, something that
allows us to achieve the same level of accuracy in our estimate with a
much lower number of samples.

\subsubsection{Derivation of a lower bound}

By substituting $p(\btheta| \by, \bd)$ from eq. \ref{eq:bayes}, we
write
\begin{eqnarray*}
U(\bd) & = &\int_{\calY} \int_{\Theta} \log \left[ \frac{p(\btheta | \by,
    \bd)}{p(\btheta)}\right] p(\btheta, \by | \bd) d\btheta d\by\\ & = &\int_{\calY}
\int_{\Theta} \log \left[ p(\by | \btheta, \bd) \right] p(\by |
\btheta | \bd) p(\btheta)
d\btheta d\by - \int_{\calY} \log \left[ p(\by) \right] p(\by | \bd)
d\by
\end{eqnarray*}
and by denoting with $\calH[q(\bomega)] = -\int \log
\left[ q(\bomega)\right] q(\bomega) d\bomega$ the entropy of a
distribution $q(\bomega)$ we take 
\begin{eqnarray*}
U(\bd) = - \int_{\Theta} \calH[p(\by | \btheta, \bd)] p(\btheta) d\btheta +
\calH[p(\by | \bd)]. 
\end{eqnarray*}
In what follows we take the likelihood as a Gaussian
distribution with density
\begin{equation*}
p(\by | \btheta, \bd) = (2\pi)^{-m/2} |\bSigma|^{-1/2} \exp\left\{-
  \frac{1}{2}\left[\by - \calG(\btheta, \bd)\right]^T \bSigma^{-1}\left[\by -
    \calG(\btheta, \bd) \right]  \right\}.
\end{equation*}
where the mean is the model output $\calG(\btheta,\bd)$ and the covariance
matrix is $\bSigma$. This is a common choice for models where the
observables $\by$ are defined as the model prediction plus some
measurement errors
\begin{equation*}
\by = \calG(\btheta, \bd) + \epsilon
\end{equation*}
with the latter being normally distributed with density $\calN(0,
\bSigma)$. In general, the measurement error can be related to the model
output through a proportionality factor or even a more complex relation
which is typically incorporated in the covariance matrix
$\bSigma$. This implies that $\bSigma$ can depend implicitly on both
the uncertain and the design parameters. For our purposes we make the
rather simple assumption that no such dependence is involved and
$\bSigma$ has fixed entries. Further details on the exact values of the covariance
matrix including independence among the measurement errors is
provided later in our applications. 

For the above choice of the likelihood function the entropy can
be calculated and is simply 
\begin{equation*}
\calH[p(\by | \btheta, \bd)] = \frac{1}{2}\left\{m+\log\left[ (2\pi)^m
    |\bSigma|\right] \right\}
\end{equation*}
and we finally get 
\begin{equation}
\label{eq:eig_2}
U(\bd) = -\frac{1}{2}\left\{m + \log\left[ (2\pi)^m |\bSigma|\right] \right\} + \calH[p(\by
| \bd)]. 
\end{equation}
At last, using Jensen's inequality, we derive a lower bound for
$U(\bd)$. Namely we have 
\begin{equation*}
\calH[p(\by | \bd)] = \int_\calY -\log\left[p(\by | \bd)\right] p(\by
| \bd) d\by \geq -\log\left[ \int_\calY p^2(\by | \bd) d\by \right]
\end{equation*}
and we define the lower bound of $U(\bd)$ as 
\begin{equation}
\label{eq:eig_lbo}
U_L(\bd) =  -\frac{1}{2}\left\{ m + \log\left[ (2\pi)^m |\bSigma|\right]
  \right\} - \log\left[ \int_{\calY} p^2(\by | \bd) d\by \right]\ .
\end{equation}
Note that since the common first term of $U(\bd)$ and $U_L(\bd)$, as they
appear in equations (\ref{eq:eig_2}) and (\ref{eq:eig_lbo}), are independent of
$\bd$, one eventually needs only to maximize the second term, namely
the entropy of the marginal distribution of the data or its lower
bound. This idea is not new and has been previously 
used in linear regression models \cite{sebastiani} and in geophysical applications
\cite{vandenberg}.

\subsubsection{Estimation of the lower bound}

As mentioned above, the optimization problem is equivalent to minimizing
\begin{equation*}
U^*_L(\bd) =  \int_{\calY} p^2(\by | \bd) d\by.
\end{equation*}
After expanding the evidence function and writing 
\begin{equation*}
U^*_L(\bd) = \int_\calY \int_\Theta \int_\Theta p(\by | \btheta_1, \bd) p(\by |
\btheta_2, \bd ) p(\btheta_1) p(\btheta_2) d\btheta_1 d\btheta_2 d\by
\end{equation*}
we can see that a Monte Carlo estimate of the above is
\begin{equation}
\label{eq:lbo_estimate}
\hat{U}^*_L(\bd) =   \frac{1}{NM} \sum_{i,j = 1}^{N,M}
  p(\by^{i} | \btheta^{j}, \bd)
\end{equation}
where $\{\btheta^i\}$, $\{\btheta^{j}\}$, $i=1,..., N$, $j=1,..., M$ are i.i.d. samples from $p(\btheta)$ and
$\{\by^{i}\}$ are i.i.d. samples from $p(\by | \btheta^i, \bd)$.

\section{Stochastic optimization}
\label{sec:stoch_opt}

\subsection{Simultaneous perturbation stochastic approximation}

Maximization of the objective function derived in the above section
can in general be a difficult task. In cases where the design
parameters are of high dimension or an expensive forward solver is
involved in the evaluation of $U^*_L(\bd)$, a direct grid search can
easily become prohibitive. Since only a noisy estimate of the actual
objective function to be maximized is available, we turn our attention
to stochastic approximation methods. These are algorithms that
approximate roots of noisy functions and can be used for optimization
problems to find the roots of the gradient of the objective function
to be optimized. This is done with an iterative procedure of
the general form 
\begin{equation}
\bd_{k+1} = \bd_k - a_k g(\bd_k), \ \ \ k \geq 0
\end{equation}
where $\{a_k\}_{k \geq 0}$ is a sequence of positive pre-specified
deterministic constants, known also as the \emph{learning rate} of the
algorithm and $g(\bd) = \nabla_\bd U(\bd)$ is the gradient of the
objective function with respect to $\bd$. Algorithms that use an explicit
expression for the gradient are called \emph{Robbins-Monro algorithms}
\cite{monro} and those where a finite-difference scheme is
employed, are called \emph{Kiefer-Wolfowitz algorithms}
\cite{kiefer}. For our purposes we use an algorithm from the
\emph{Kiefer-Wolfowitz} family, namely the \emph{Simultaneous
  Perturbation Stochastic Approximation} method (SPSA), proposed by
Spall \cite{spall_alg, spall_cookbook}. What makes this methods
attractive versus others is the fact that only two forward evaluations
are required for the gradient estimation. The updating step of the
method is given by 
\begin{equation}
\bd_{k+1} = \bd_k - a_k \hat{g}_k(\bd_k),
\end{equation}
where 
\begin{equation}
\hat{g}_k(\bd_k) = \frac{\hat{U}^*_L(\bd_k + c_k\bDelta_k) -
  \hat{U}^*_L(\bd_k - c_k\bDelta_k)}{2 c_k} \left[ \begin{array}{c}
    \Delta^{-1}_{k, 1} \\ \vdots \\ \Delta^{-1}_{k, n_d} \end{array} \right],
\end{equation}
\begin{equation}
a_k = \frac{a}{(A + k + 1)^{\alpha}}, \ \ c_k = \frac{c}{(k + 1)^{\gamma}}
\end{equation}
and $a_k$, $c_k$, $\alpha$, $\gamma$ are positive parameters whose values
affect the convergence rate and the gradient estimate. General
instructions on how to choose the appropriate values for each problem
are given in \cite{spall_cookbook}. The vectors $\bDelta_k$ are random
vectors with coefficients $\Delta_{k, i}$ drawn independently from any
zero-mean probability distribution such that the expectation of $|\Delta^{-1}_{k,i}|$
exists \cite{spall_alg}. A common choice, which we used in our
analysis, corresponds to $\Delta_{k,i} = 2 ( Z - 1/2)$ where $Z\sim
\mbox{Bernoulli} (p)$ with success probability $p = \frac{1}{2}$.

\section{Example: Nonlinear model}
\label{sec:toy_example}

\subsection{The model and experimental scenarios}

The main goal in this example is to explore the accuracy of the lower bound
estimate as a substitute for the direct maximization of
the expected information gain \cite{marzouk_simul}
and provide a numerical comparison of the two approaches. This is
achieved by evaluating the two objective functions for a
simple algebraic model which is inexpensive to evaluate and is nonlinear with
respect to both the uncertain and design parameters.  Consider thus the
model where the observable quantity $y$ is dependent on $\kappa$ and
$d$ as 
\begin{eqnarray}
y(\kappa, d) & = & G(\kappa, d) + \epsilon \\ 
& = & \kappa^3 d^2 + \kappa e^{-|0.2 - d|} + \epsilon
\end{eqnarray}
where $G(\kappa, d)$ is the model output and the noise is taken to be
$\epsilon \sim \calN(0, 10^{-4})$. For prior we choose initially $\kappa \sim
\calU (0,1)$ and later on we discuss a few more cases. Suppose we have
control over $d \in \calD$ where $\calD = [0,1]$ is the design space
and we are interested in inferring $\kappa$. We explore two cases,
first the case where inference is carried out using a
single observation of $y$ and second the case where two observations
of $y$ can be obtained, corresponding to different values of $d$. One can
think of the design parameter $d$ as the location where $y$ is
observed. Before observing $y$, we would like to know
the value of $d$ that would make our observations the most informative
ones. Both cases of this example have been
studied in \cite{marzouk_simul} using the direct Monte Carlo estimate
of $U(d)$. The first case was also studied in \cite{long_quan} using the Laplace approximation of
the posterior distribution of $\kappa$ and then performing the integration with
sparse quadratures.

\begin{figure}[h]
\centering
\includegraphics[width=0.40\textwidth]{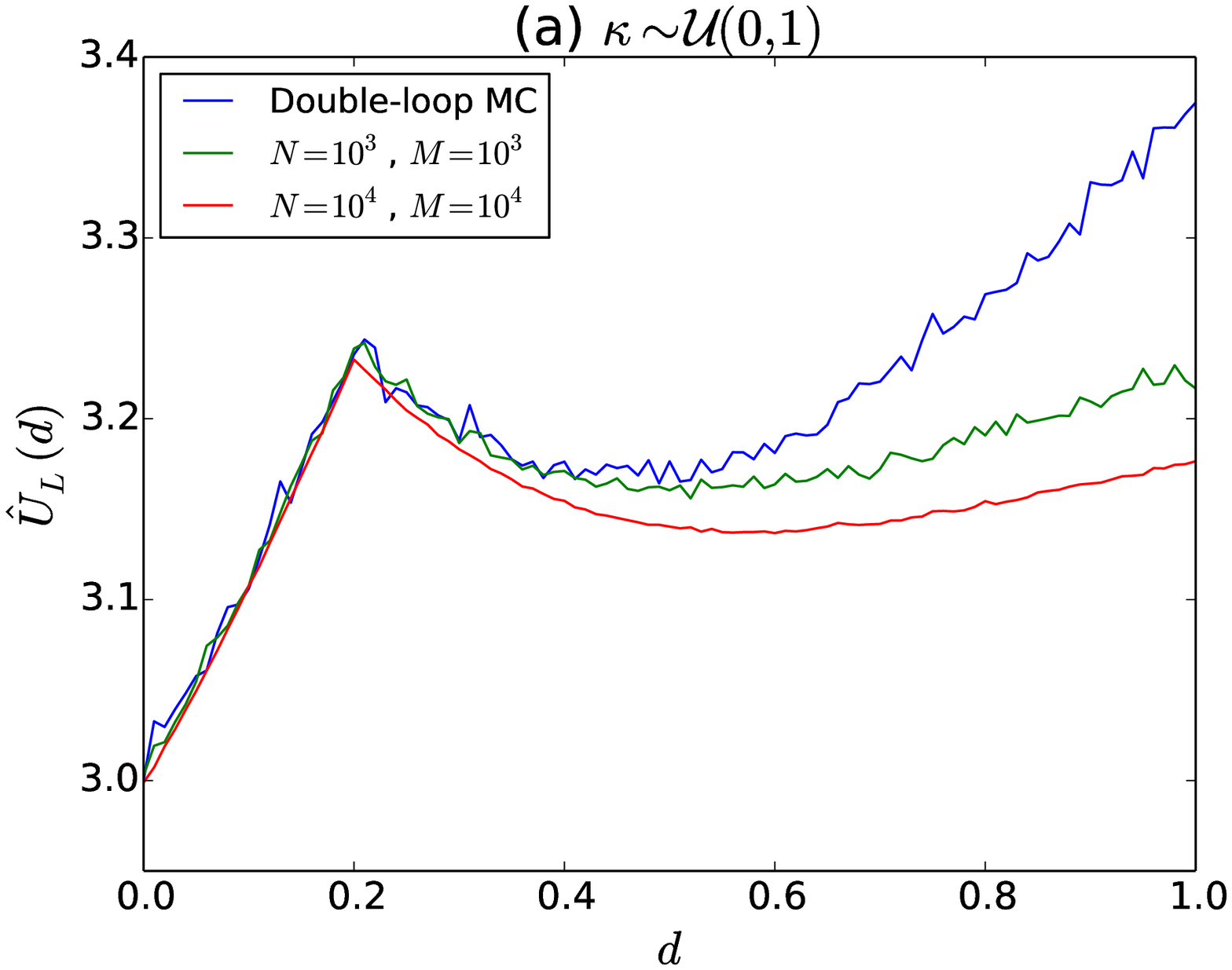}\\
\includegraphics[width=0.40\textwidth]{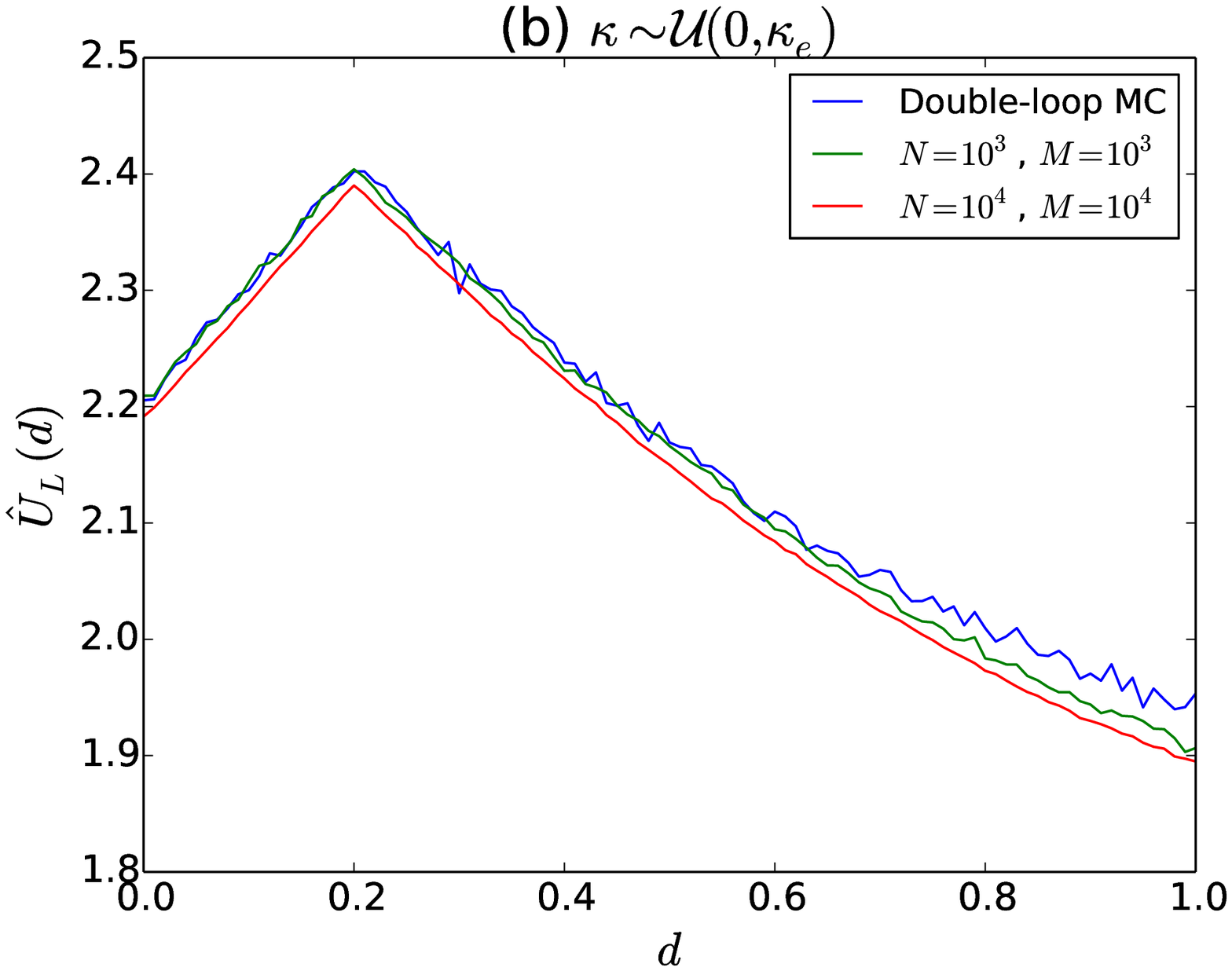}
\includegraphics[width=0.40\textwidth]{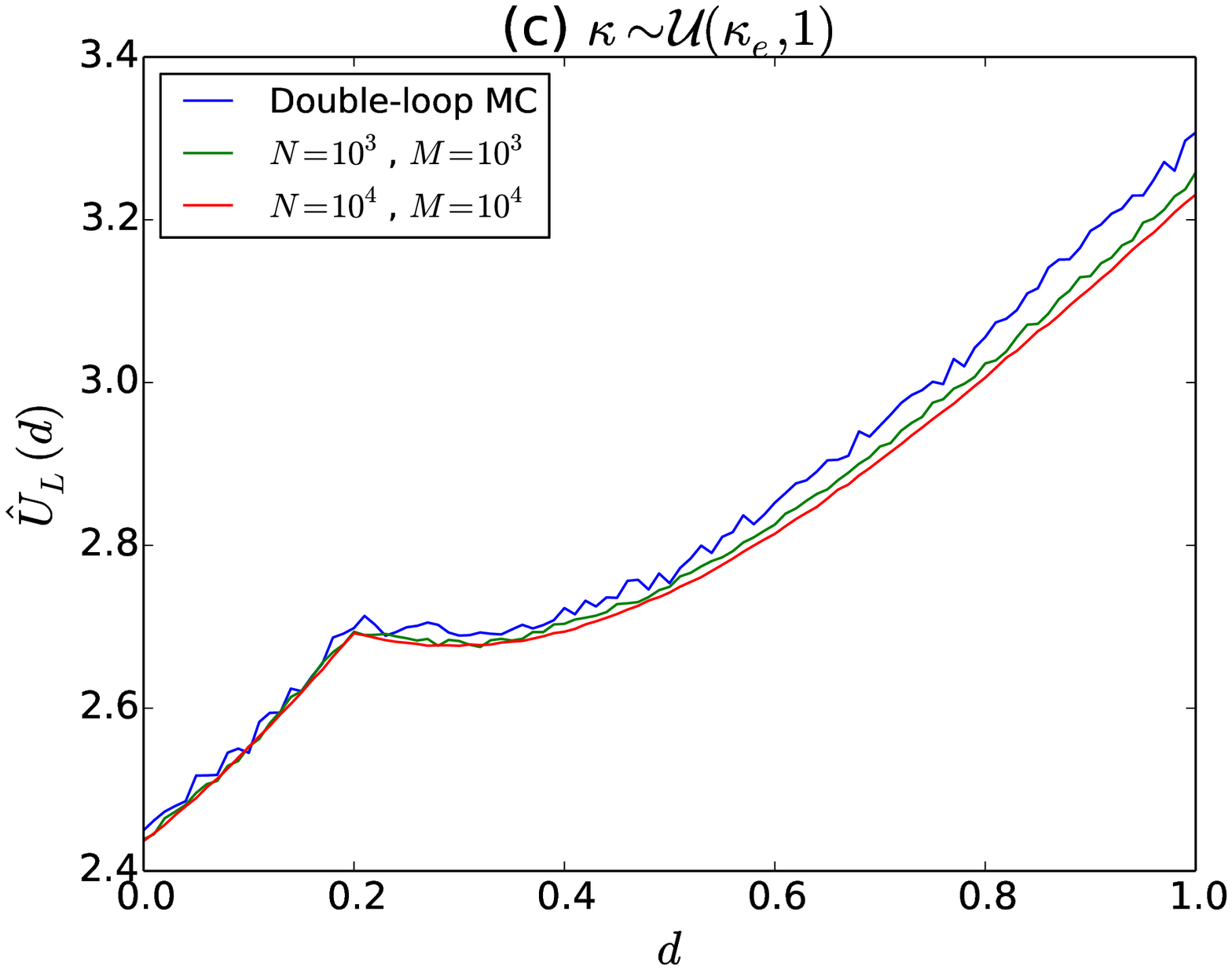}
\caption{Expected information gain lower bounds and dlMC with $N =
  10^4$, $M = 10^4$ estimated over $\calD =
  [0,1]$ for three different priors. \label{fig:toy_estimate}}
\end{figure}

\subsection{Results}

We are using the expected information gain lower bound estimate as
given in (\ref{eq:lbo_estimate}) as our criterion to determine the
optimal design $d$ for inferring $\kappa$. For comparison, we also
reproduce the results of \cite{marzouk_simul} using the expected
information gain estimate which from now on we call
\emph{double loop Monte Carlo} (dlMC) estimate. The exact
expression of the dlMC estimate is given in \ref{sec:appendix_a}. Our
computations are performed using an
ensemble of samples with $N = M = 10^4$
for all different priors that we consider below, while keeping the
error variance fixed. We demonstrate that the two estimate share the
same maxima and their graphs have good quantitative agreement with the
lower bound providing slightly less noisy values. 

\subsubsection{Single observation}
Fig.~\ref{fig:toy_estimate} shows the values of estimates of $U_L^*(d)$ for
the design of one experiment after using a $101$-point
uniform partition on the design space $\calD = [0,1]$ for $N = M =
10^3$ and for $N = M = 10^4$ as well as the estimates of dlMC. All cases
show the existence of two local maxima at $d = 0.2$ and $d = 1$.
As expected, our estimate always takes values slightly smaller than dlMC,
as the former is a lower bound of the latter. Due to the fact that the
first term in our estimate (the Gaussian entropy) is a constant, while
the first term of the dlMC estimate is a Monte-Carlo approximation of the very same
constant, we observe that the lower bound is a smoother curve for the
same values of $N$, $M$. Following
the slope analysis of \cite{marzouk_simul} we also present the results
for two different choices of prior, namely when $\kappa \sim \calU(0,
\kappa_e)$ and when $\kappa \sim \calU(\kappa_e,1)$ where $\kappa_e =
\left[(1 - e^{-0.8}) / 2.88 \right]^{1/2}$ which is the point where the slope of $G(\kappa,
1)$ becomes greater than the slope of $G(\kappa, 0.2)$. For the former
case we obtain a global maximum at $d = 0.2$ whereas for the latter,
the maximum is at $d = 1$. At some particular points we can observe that although our estimate is only
a lower bound of the dlMC estimate, the value of the dlMC might fall
below the value of $U^*_L(d)$. This, in addition to the fact that we
compare noisy estimates of a deterministic quantity, can also emerge
as the result of the bias of dlMC which is not negligible for our choice
of $M$ (number of inner loop samples). For a more detailed discussion on the
bias of dlMC for this example see \cite{marzouk_simul} and for the
analytic form of the bias see \cite{ryan}.

\subsubsection{Two observations}

\begin{figure}
\centering
\begin{subfigure}[b]{\textwidth}
\includegraphics[width = 0.32\textwidth]{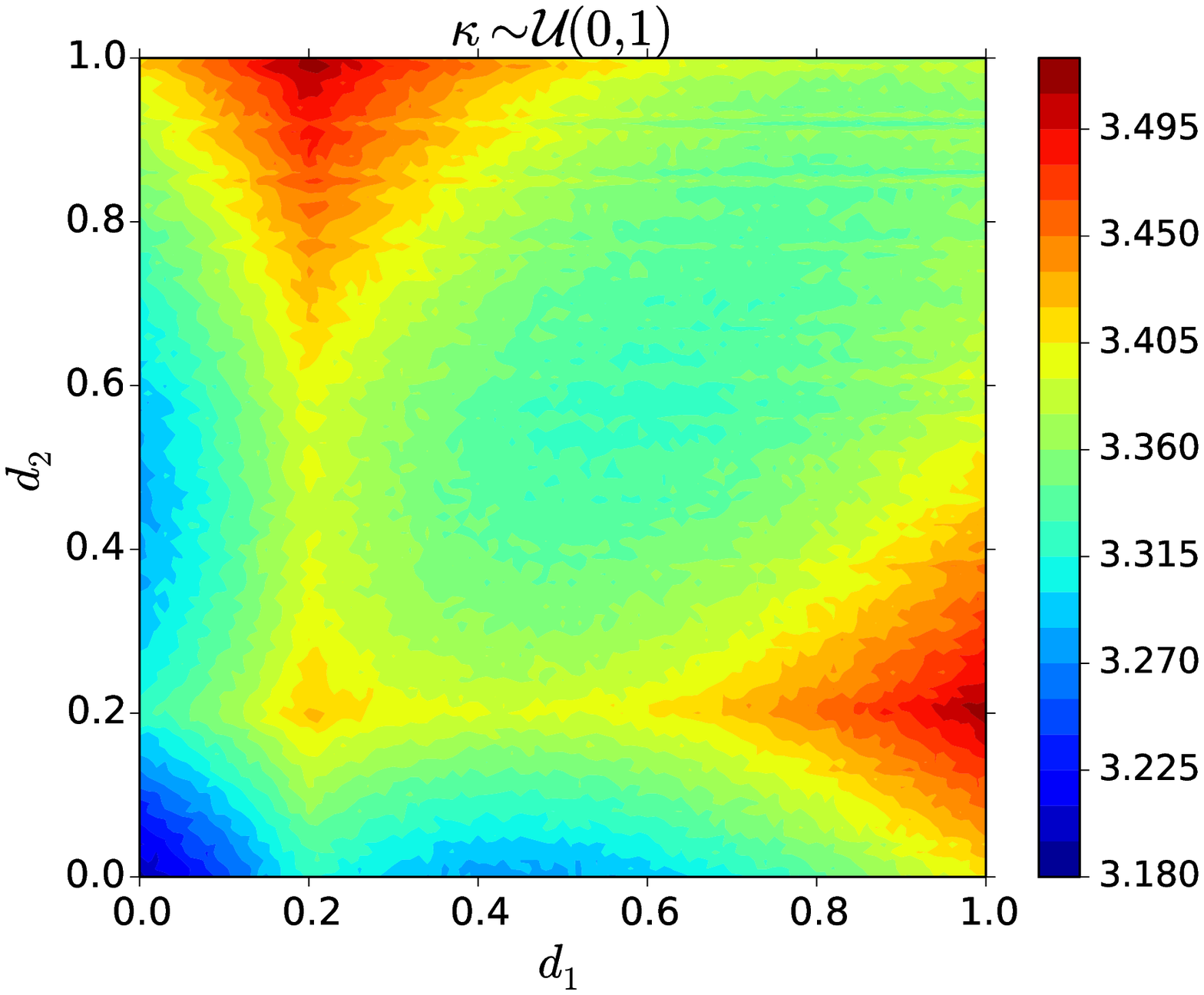}
\includegraphics[width = 0.32\textwidth]{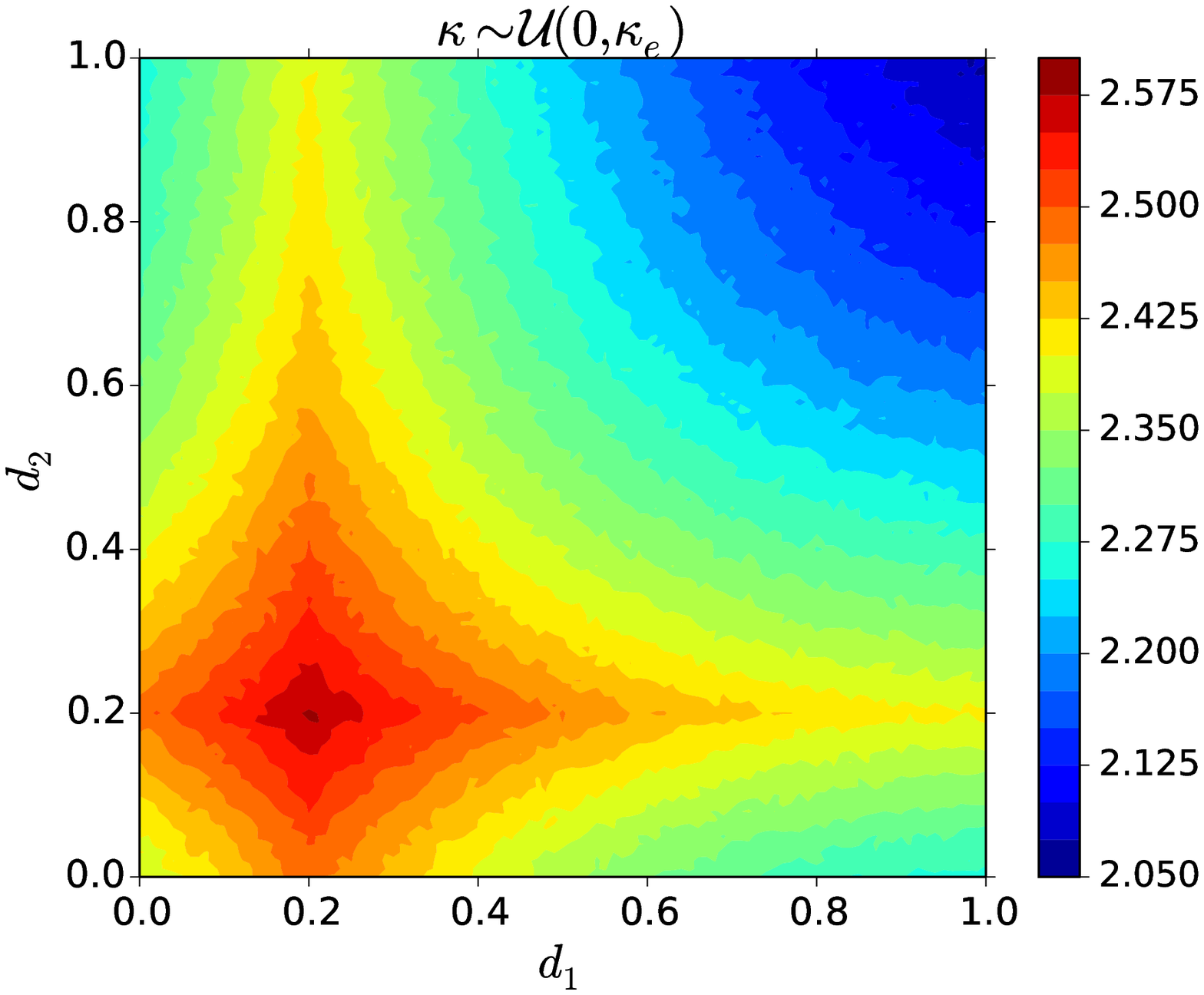}
\includegraphics[width = 0.32\textwidth]{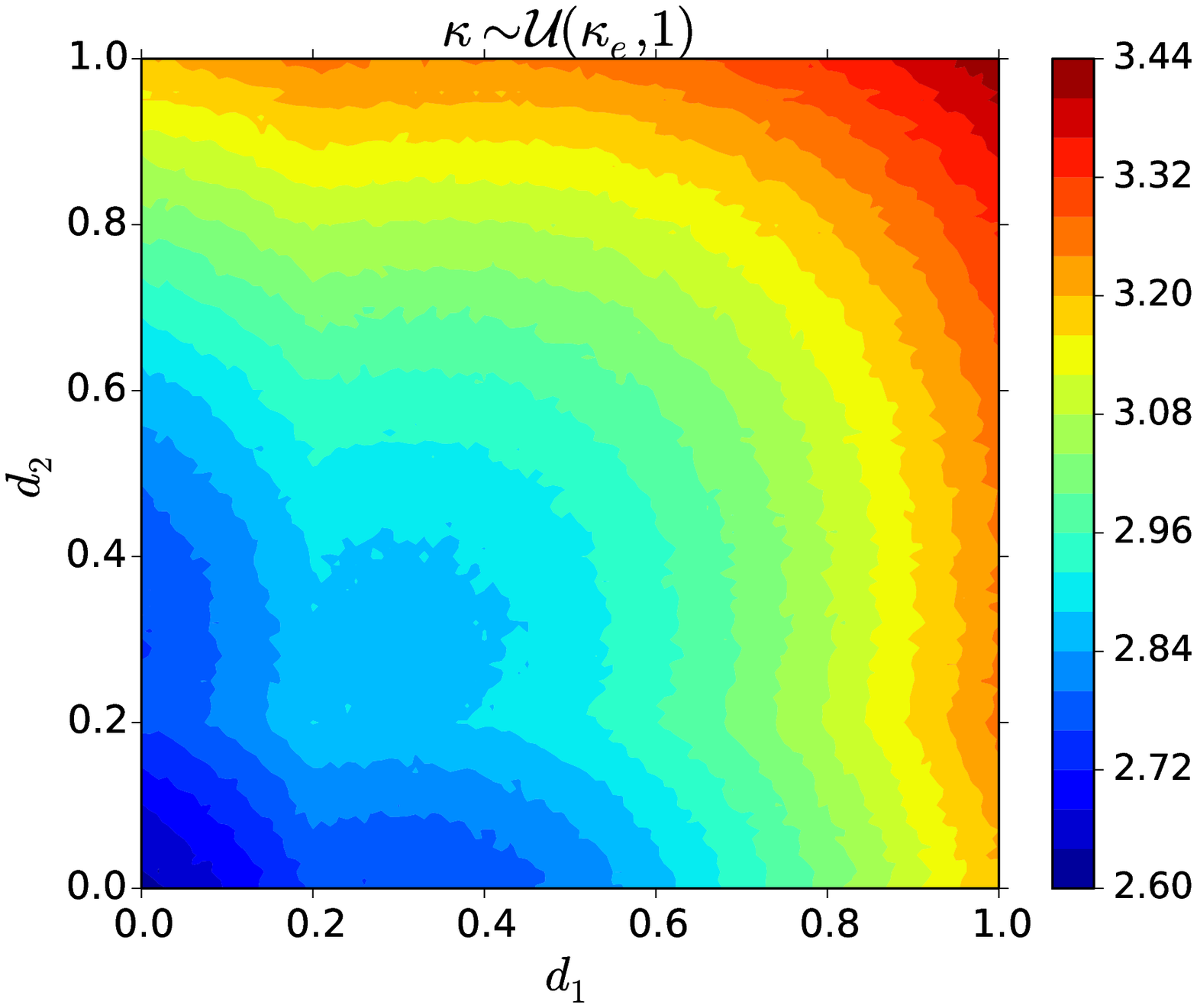}
\caption{Lower bound estimate $\hat{U}_L(\bd)$}
\end{subfigure}
\begin{subfigure}[b]{\textwidth}
\includegraphics[width = 0.32\textwidth]{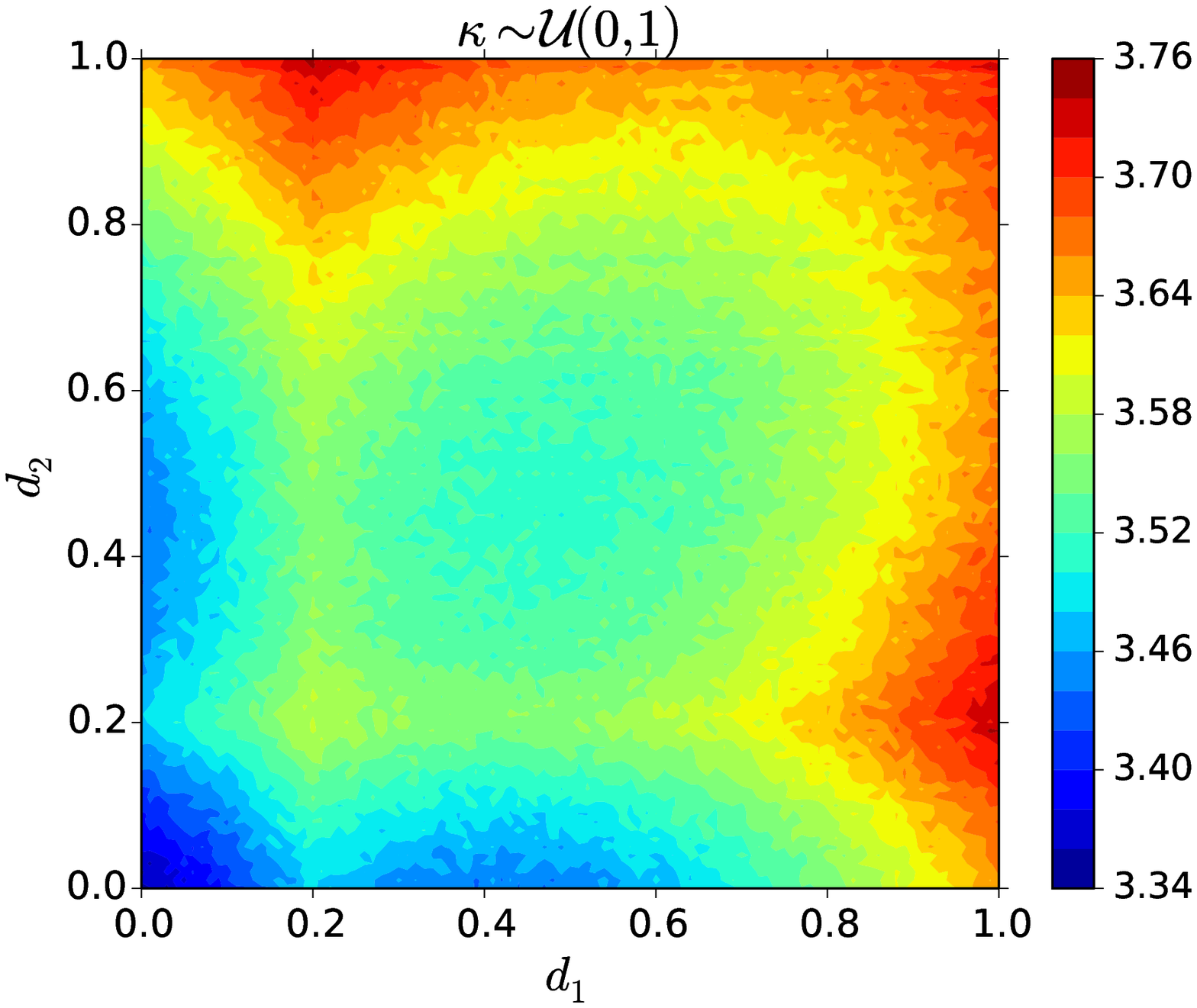}
\includegraphics[width = 0.32\textwidth]{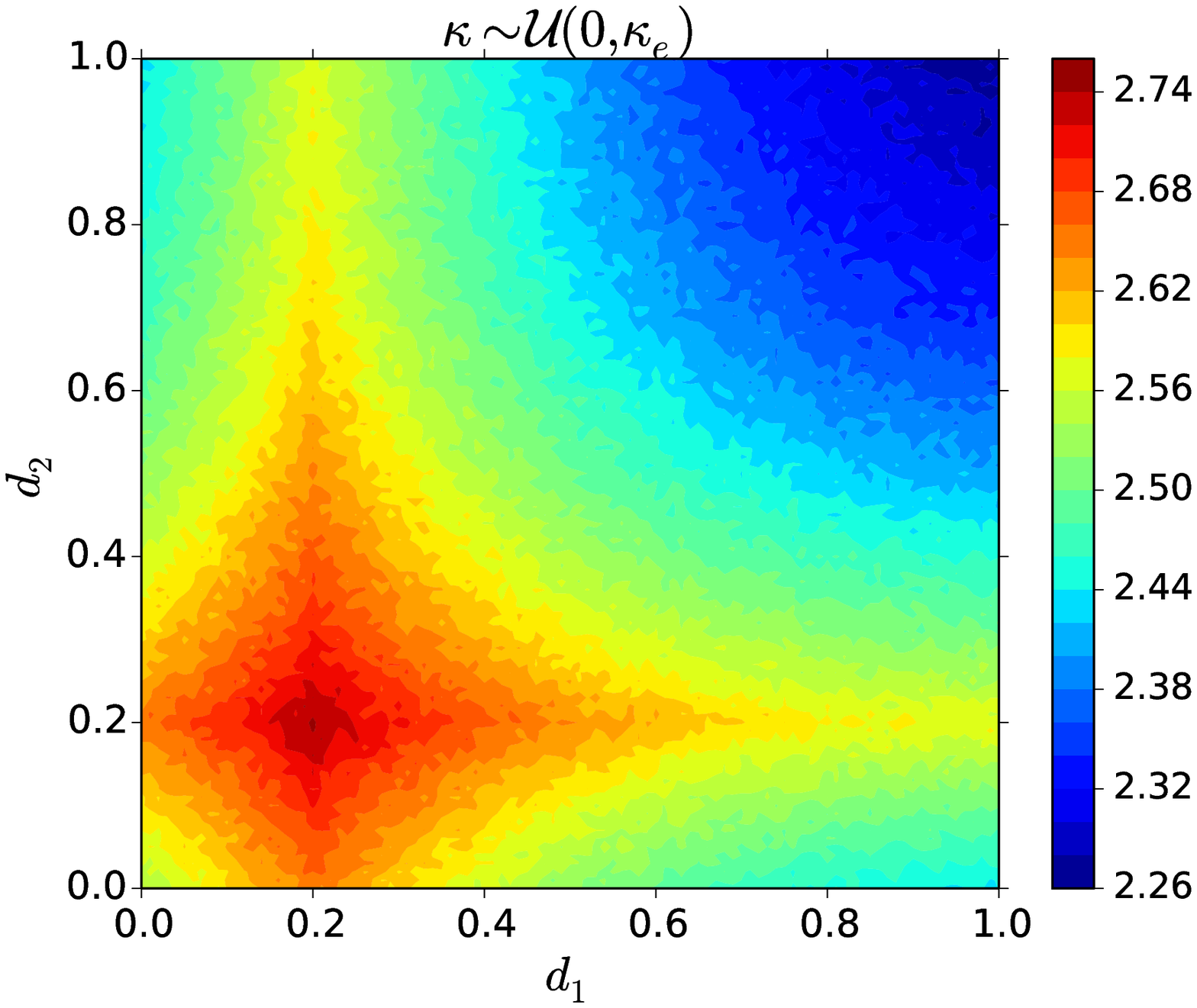}
\includegraphics[width = 0.32\textwidth]{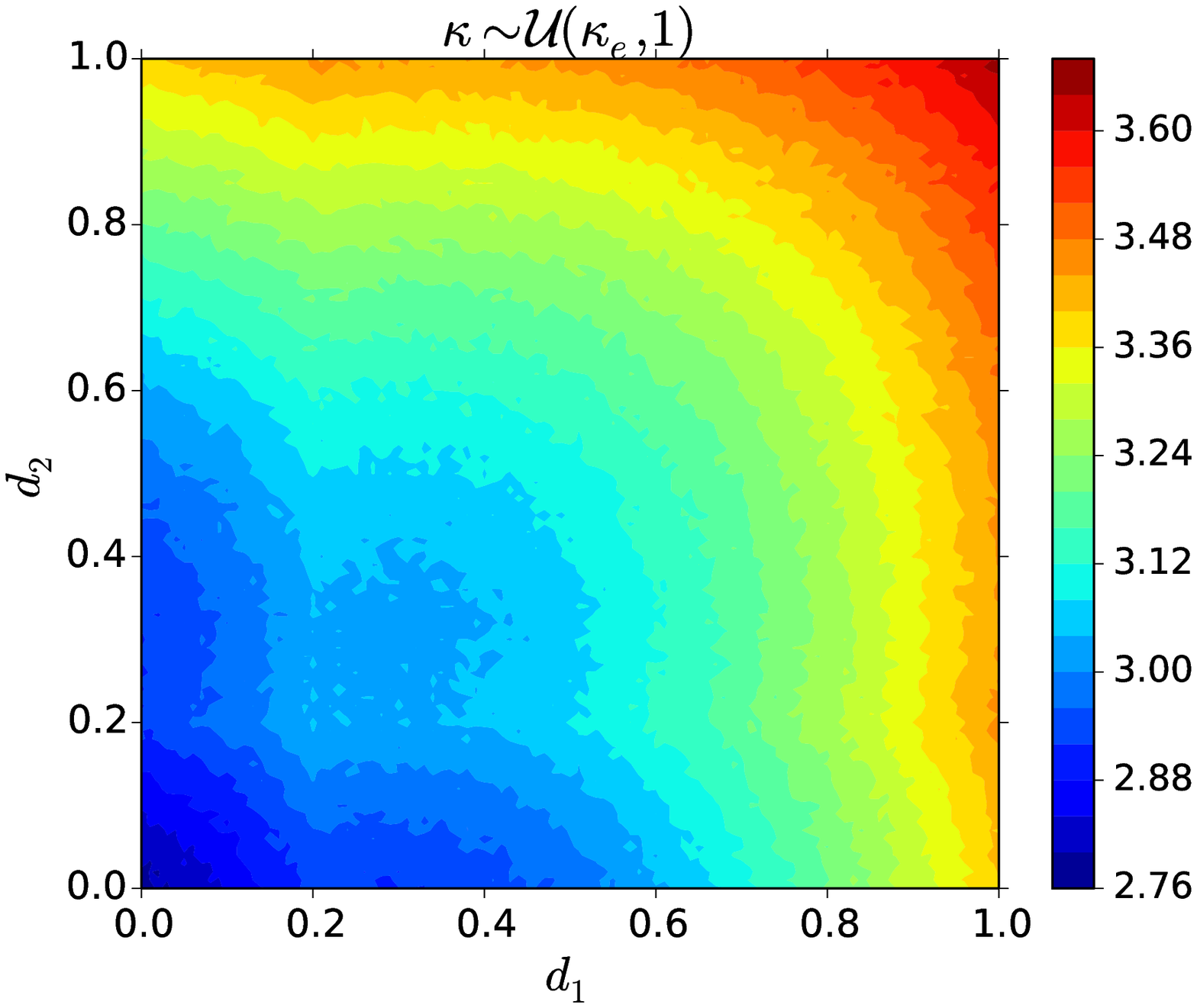}
\caption{dlMC estimate $\hat{I}(\bd)$}
\end{subfigure}

\caption{Expected information gain lower bound estimate and dlMC estimate with $N =
  10^4$, $M = 10^4$ for the 2-dimensional design, estimated over $\calD =
  [0,1]^2$ for different priors. \label{fig:toy_estimate_2d}}
\end{figure}

We calculate the values of $U^*_L(d_1, d_2)$ for $N = M = 10^4$
over $\calD \times \calD  = [0,1]^2$ and we
compare them with those of dlMC in Fig.~\ref{fig:toy_estimate_2d}. Similar
qualitative results as in case 1 can be observed. For $\kappa \sim
\calU(0,1)$, the points where maximum is attained are $(d_1, d_2) = (0.2, 1)$
and $(d_1, d_2) = (1, 0.2)$ which are combinations of the local maxima in the
1-dimensional case. That means that, if two observations can be
afforded, they should be taken at the points where local maxima exist
for the one observation scenario. The order is insignificant since, as
we see, the $U^*_L(d_1, d_2)$ surface is symmetric with respect to the
$d_1 = d_2$ line. Similar conclusions are drawn for $\kappa \sim
\calU(0,\kappa_e)$ and $\kappa \sim \calU(\kappa_e,1)$ where the
maxima are at $(d_1, d_2)= (0.2, 0.2)$ and $(d_1, d_2) = (1, 1)$
respectively, as already expected from the results of case 1. 

\section{Example: Case study}
\label{sec:case_study}

This second example demonstrates the application of the formalism to
a subsurface pollution characterization problem for an actual site
where field data for permeability and concentrations are available. 

\subsection{Description}
\label{sec:description}

\begin{figure}[th]
\begin{center}
\includegraphics[width = 7.5cm]{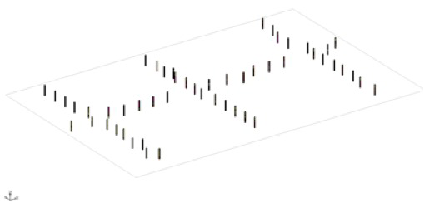}
\includegraphics[width = 6cm]{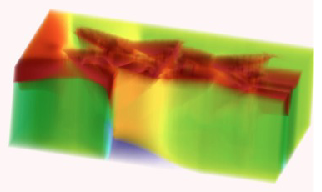}
\caption{Boreholes (left) and final domain
  stratigraphy (right) with the $z$-axis being stretched for the sake
  of illustration. \label{fig:boreholes}}
\end{center}
\end{figure}

\begin{figure}[tbh]
\centering
\begin{subfigure}[b]{0.30\textwidth}
\includegraphics[width = \textwidth]{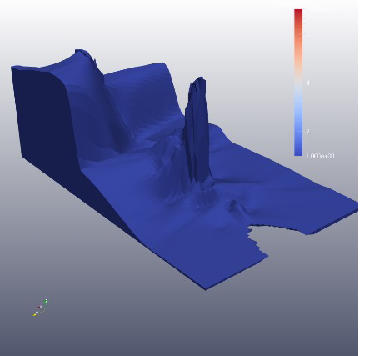}
\caption{Clay}
\end{subfigure}
\begin{subfigure}[b]{0.30\textwidth}
\includegraphics[width = \textwidth]{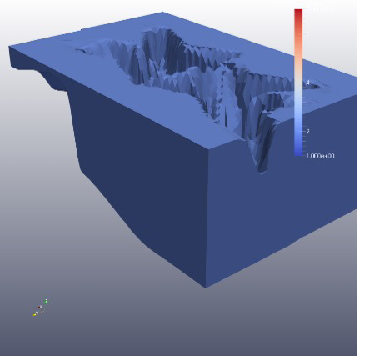}
\caption{Silt}
\end{subfigure}
\begin{subfigure}[b]{0.30\textwidth}
\includegraphics[width = \textwidth]{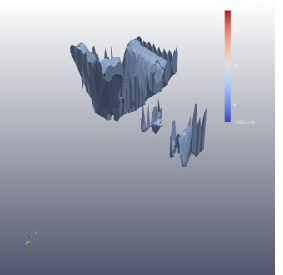}
\caption{Clayey sand}
\end{subfigure}
\begin{subfigure}[b]{0.30\textwidth}
\includegraphics[width = \textwidth]{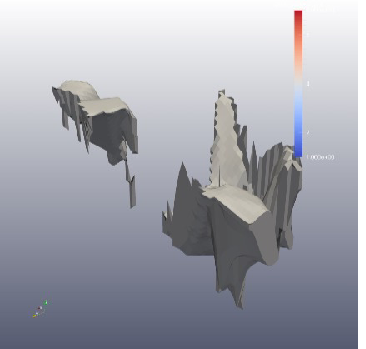}
\caption{Silty sand}
\end{subfigure}
\begin{subfigure}[b]{0.30\textwidth}
\includegraphics[width = \textwidth]{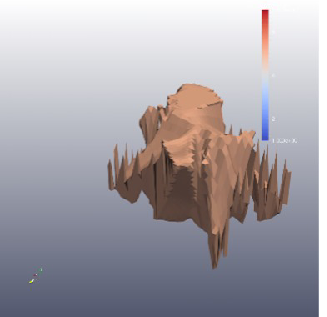}
\caption{Sand with 10\% silt}
\end{subfigure}
\begin{subfigure}[b]{0.30\textwidth}
\includegraphics[width = \textwidth]{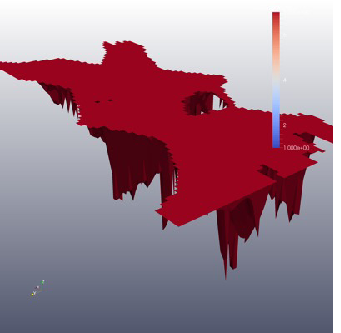}
\caption{Poorly graded sand}
\end{subfigure}
\caption{Soil morphology of the domain. The $z$-axis has been
  stretched for the sake of illustration. \label{fig:soils}}
\end{figure}

We are interested in performing Bayesian inference on the uncertain
parameters that affect the transport of pollutants in a contaminated
area. Such a procedure will decrease the uncertainty regarding the
extent and location of the plume and will further affect the cost and
other important aspects of soil remediation. When limited resources
are available for experimentation and data collection, it is of great
importance to develop experimental design strategies that will enhance the
quality of data. Our current study concerns a
contaminated site located in Santa Fe Springs, California. Previous
data consisting of soil types is available from boreholes located along
four cross sections across the site. Each borehole reaches a depth of
20m.  Accordingly, soil type is defined as a categorical variable
with six possible states. 
The field and the locations of the boreholes are shown in
Figs.~\ref{fig:world_map} and~\ref{fig:boreholes} (left). The available soil data is used to construct a
domain that can be used in our forward model to simulate
the transport of pollutants. The construction of a domain that can be
regarded as a good approximation of the real situation, is achieved by
the stratigraphic modeling capabilities of the Groundwater
Modeling System (GMS) software \cite{aquaveo}. More specifically,
using a standard inverse-distance weighting based interpolation scheme, we
assign soil types for all locations in the area of our interest. The stratigraphy of the resulting
domain as well as the extent of each soil are shown in
Figs.~\ref{fig:boreholes} (right) and~\ref{fig:soils}. Next, the
domain was discretized to a $40 \times 60 \times 10$ grid where each cell has
dimensions $15.4$m.$\times 6.7$m.$\times 2$m. and carries the information of the
soil type present in that location. The grid is used as our
finite volume discretization in the subsequent TOUGH2 simulations and
each of the six different soil types present in our domain are
assigned different permeability values which, in our study, are
considered to be random and are the only source of uncertainty.

\begin{figure}
\begin{center}
\includegraphics[width = 0.37\textwidth]{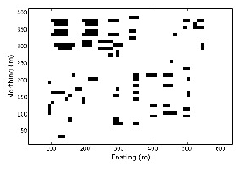}
\caption{Locations where initial contaminant injections are placed. \label{fig:sources}}
\end{center}
\end{figure}

\subsection{Simulating flow and transport using TOUGH2}
\label{sec:tough2}

\subsubsection{EOS7r module}

We employed the multiphase simulator TOUGH2 \cite{pruess} and its
massively parallel version TOUGH2-MP \cite{pruess_mp}  with the EOS7r
module to simulate groundwater flow and contaminant transport. TOUGH2
provides us with a finite volume solver which discretizes the mass and
energy equations over space and time and updates a set of primary
variables that consist the solution of the governing equations by
estimating the increments at each time step with a Newton-Raphson
method. The common mathematical form of the equations for multiphase
fluid flow include several thermophysical parameters such as density,
viscosity, enthalpy etc. which are determined by the various "EOS"
(equation-of-state) modules. The EOS7r module \cite{olden} used here, is mainly intended
to provide radionuclide transport capability
but can be as well used for transport of volatile and water soluble
organic chemicals (VOCs). Change in concentrations is caused in
general for three reasons: transport, decay and adsorption. 
Volatilization of the VOCs in all phases is modeled by Henry's law and occurs by advection
and diffusion. Decay is modeled by a first-order decay law. In case of radionuclide
transport, this is interpreted as radioactive decay but in the case
of organic contaminants it can be explained as
biodegradation. Adsorption is dependent on the distribution
coefficient that characterizes each rocktype. EOS7r in total can
simulate transport of five components in a two-phase flow, namely water,
air, brine, a parent contaminant and a daughter contaminant in aqueous
and gaseous phases.

\subsubsection{Santa Fe site}

For our problem we are interested in simulating the spreading of
organic contaminants in a field located in Santa Fe Springs, CA which
is approximately  $600 m. \times 400 m.$ and only $20 m.$ deep. This
is discretized in TOUGH2 to a grid consisting of $40 \times 60 \times 10 =
24000$ active cells where each cell has dimension $15.4 m. \times 6.7
m. \times 2 m.$, as explained in the previous paragraph and is assigned a rocktype according
to its soil type. We focus on investigating the propagation of uncertainty emerging from
the unknown permeabilities $\bk = (\kappa_1, ..., \kappa_6)$ of the
six different soil types the are present
on our site, through the transport process of the VOCs. The porosity
is taken to be uniform $\phi = 0.35$ all over the domain. Although we
are interested in the transport mainly of petroleum hydrocarbons whose
weathering processes are in general known to include adsorption
and biodegradation effects, for the purposes of this study we will
consider these effects to be of negligible importance by assigning the
distribution coefficients for adsorption to be zero and the half-life
parameter $\sim 10^{50}$ so that we have no decay effects. Thus our model
focuses only on the volatilization properties of the VOCs.  We choose the values for the molecular
weight and inverse Henry's constant parameters to correspond to those
used for describing transport of petroleum hydrocarbons and
specifically those that are mostly detected using gas chromatography
techniques, named Gasoline Range Organics (GRO). Typically GRO
are the subsets of hydrocarbons in the C5-C12 range and include
hydrocarbons such as Benzene, Toluene, Ethylbenzene, m-, o- and
p-Xylenes, Naphthalene and Acenaphthene. Their molecular weights are
in the $78-154$ grams range. In our case we arbitrarily set the
molecular weight to be $112.4$ grams which is a value close to the average
GRO hydrocarbon.

\begin{table}[tbh]
\caption{Material and initial parameters used in our simulations.}
\label{tab:tough2_params}
\centering
\begin{tabular}{l | c | c}
Parameter & Symbol & Value \\
\hline
Porosity & $\phi$ & $0.35$ \\
Tortuosity factor & $\tau_0$ & $0.1$ \\
Relative permeability parameters: & $\lambda$ & $0.457$\\
 -  & $S_{lr}$ & $0.15$ \\
 -  & $S_{ls}$ & $1$ \\
 -  & $S_{gr}$ & $0.1$ \\
Capillary pressure parameters: & $\lambda$ & $0.457$ \\
 - & $S_{lr}$ & $0$ \\
 - & $1 / P_0 = \alpha/ \rho_w g$ & $5.105e-4$\\
 - & $P_{\max}$ & $10^7$\\
 - & $S_{ls}$ & $1$\\
Diffusivities (all $k$): gas phase& $ d_{g}^k $ & $10^{-6}$ \\
aqueous phase & $d_{l}^k$ & $10^{-10}$\\
Molecular weight & - & $112.4$ \\
Inverse Herny's constant & - & $2.1\cdot 10^{-8}$ \\
Half-life parameter & - & $10^{50}$ \\
Initial pressure & $P(0)$ & $1.013\cdot 10^{5}$ \\
Initial gas saturation & $S_g$ & $0.75$ \\
Temperature (constant) & $T$ & $20^{\circ}$ \\
\end{tabular}
\end{table}

For our forward evaluations of the model with TOUGH2 we consider that only $3$ components
are present by assigning the brine and daughter radionuclide mass fractions to be zero. 
We assume isothermal conditions with the temperature being
constant at $20 ^{\circ}$C and no-flux boundary conditions. The
mathematical formulation of the flow and transport model is described
in detail in Appendix~\ref{sec:appendix_b} and the values of the model
parameters that are assumed known are given in Table~\ref{tab:tough2_params}.
Since we want the pressure and gas saturation to be close to
steady-state conditions by the time the contaminants are released to the ground,
we first run the model for a time period until $T = 13.3$ years, where the initial mass
fractions for the contaminants is zero. The outputs of pressure and
gas saturation are subsequently used as initial conditions for the transport simulation. We
rerun the model after assigning initial aqueous phase solubilities for
the VOC. We denote these initial conditions with $\by_0$. Their
locations at time $t = 0$ are taken to be approximately in the areas were the
storage tanks are located as can be seen in
Fig.~\ref{fig:sources}. For our purposes we add $164$ inactive
(zero volume) cells to the surface layer of our grid and initial injections are assigned. The exact
values were chosen to be $\by_0 = 0.1 + U$ ppb (parts per billion) where $U$ are numbers
randomly generated from a Uniform distribution with support on
$(5\cdot 10^{-5}, 10^{-2})$ and are considered known. The transport
simulation runs for the same time period as the flow simulation.

\subsection{Developing a surrogate model}
\label{sec:surrogate}

Due to the complexity of our model, only a single simulation of the
transport flow with TOUGH2 requires several minutes to finish. Thus,
implementing our experimental design
framework, including the minimization of the expected information gain
lower bound with SPSA which requires thousands of
objective function evaluations, becomes impractical. It is therefore
necessary to create a surrogate model that would provide
us with forward evaluations of the model that are significantly
cheaper to obtain than running TOUGH2. 

\subsubsection{The prior and input transformation}

The unknown physical parameters of our problem are the permeabilities $\bk
= (\kappa_1, ..., \kappa_6)$ of the six materials making-up the
subsurface at the site of interest. We choose all
$\kappa_i$ to be independent with a lognormal prior distribution, that is
\begin{equation}
p(\log \kappa_i) = \calN( -23.5, 4)
\end{equation}
The choice of the mean $\mu = -23.5$ and variance $\sigma^2 = 4$ are made
such that our prior covers a range for permeability values in the
order of magnitude $10^{-8}$cm$^2$ to $10^{-12}$cm$^2$ which corresponds to
semi-pervious materials. We find this a rather general prior that
corresponds solely to our knowledge that the materials present in our
domain are silt, clay, silty sand, clayey sand, sand with 10$\%$ silt
and poorly graded sand. 

Now if we let $F(x) = P(\kappa_i \leq x)$ to be the
cumulative distribution function of $\kappa_i$, then we have that
$\xi_i := F(\kappa_i) \sim U(0,1)$ and for $\bxi = (\xi_1,...,\xi_6)$ we
define $\hat{\calG}(\bxi) = \calG(\mathbf{F}^{-1}(\bxi),\bd)$ where
$\mathbf{F}(\bxi) =\prod_{i=1}^6 F(\xi_i) $ to be our model output
where the input is a set of $6$ independent standard uniform random
variables.

\subsubsection{Polynomial chaos expansion}

We make use of the property
that our random output $\hat{\calG}(\bxi)$ is a physical process with
finite variance, therefore it is a square-integrable random field $\hat{\calG}(\bxi) \in
L^2(\R^{m})$, defined on the probability space
$([0,1]^6, \calF, P)$ and admits a
polynomial chaos representation of the form \cite{ghanem_sfem,ghanem_soize,xiu}
\begin{equation}
\label{eq:gen_pce}
\hat{\calG}(\bxi) = \sum_{\balpha, |\balpha| < \infty} \bp_{\balpha} \Psi_{\balpha}(\bxi)
\end{equation}
where $\balpha = (\alpha_1,..., \alpha_d)$ and $\alpha_i \in \N^m$ for
$i = 1,..., n$ is a multi-index
with modulus $|\balpha| = \alpha_1 + ... + \alpha_n$, each
$\bp_{\balpha}$ is a vector in $\R^m$, $\bxi$ is a second-order random
variable defined on $([0,1]^6, \calF, P)$ with values in $\R^m$ and the
functions $\Psi_{\alpha}$ form a complete set of orthonormal functions
that satisfy 
\begin{equation}
\E[\Psi_{\balpha}(\bxi) \Psi_{\bbeta}(\bxi)] = \delta_{\balpha\bbeta} =
\delta_{\alpha_1\beta_1} \times \cdots \times \delta_{\alpha_d\beta_d}
\end{equation}
Typically the random variable $\bxi$ has independent components that
follow a Gaussian, Uniform, Gamma or Beta distribution. The basis
function then is chosen to consist of multidimensional polynomials
$\Psi_{\balpha}(\bxi) = \psi_{\alpha_1}(\xi_1) \times \cdots \times
\psi_{\alpha_d}(\alpha_n)$, where $\psi_{\alpha_i}$ is respectively Hermite,
Legendre, Laguerre or Jacobi polynomials of order
$\alpha_i$. According to the input transformation in the previous
paragraph, we take the components of $\bxi$ to be $\calU (0,1)$
distributed and therefore the polynomials in the expansion will be Legendre.
For computational purposes we work with a truncated version of the
expression (\ref{eq:gen_pce}) by writing 
\begin{equation}
\label{eq:trunc_pce}
\hat{\calG}_r(\bxi) = \sum_{\balpha, |\balpha| \le r} \bp_{\balpha}\Psi_{\balpha}(\bxi).
\end{equation}
where the number of terms in the above expansions is 
\begin{equation}
N_{\bp} = |\{\balpha \in \N^n, 0 \leq |\balpha| \leq r \}| = \sum_{j=0}^r
\frac{(j + n - 1)!}{j! (n-1)!}.
\end{equation}
Eq.~(\ref{eq:trunc_pce}) provides an accurate approximation
of the true model output $\by$ as long as the
coefficients $\bp_{\balpha}$ are estimated in a fashion that they also
incorporate a transformation of $\bxi$ to the uncertain parameters of
the problem of interest.

\subsubsection{Coefficient estimation}

In order for the expression (\ref{eq:trunc_pce}) to be of use, we need
to estimate the coefficients $\bp_{\alpha}$. This in general can be
done with various methods, mainly categorized as intrusive \cite{xiu_nonintrusive} and
nonintrusive methods. We use non-intrusive methods
which are easier to implement and more convenient when the forward
simulation is seen as a black box. Popular nonintrusive methods
include approximating the coefficients by orthogonal projection of the
output on the basis functions
\cite{ghiocel}, which involves calculating multidimensional
integrals. Other methods calculate
the coefficients by solving a linear system of equations constructed
after evaluating the model on a set of samples and then either
interpolating these points (by choosing collocation points of the polynomial roots
\cite{li}) or minimizing the least squares error \cite{sahinidis}. 
The last method is the one adapted here, namely we estimate the
coefficients by linear regression taking advantage of the
linear dependence of $\hat{\calG}_r(\bxi)$ on $\bp_{\alpha}$. This requires the
evaluation of $\hat{\calG}(\bxi)$ at $N_{\calG}$ points $\{\bxi^{j}\}_{j=1}^{N_{\calG}}$
and then for each component $y_c$, $c = 1,..., 24000$ of the output
vector $\hat{\calG} = (y_1, ..., y_{24000})^T$, we solve the system 
\begin{equation}
\label{eq:regression}
\by_c = \Psi \bp_c, 
\end{equation}
where $\by_c = [y_c^1, ..., y_c^{N_{\calG}}]^T$,  $\Psi$ is the
$N_{\calG} \times N_{\bp}$ matrix formed by evaluating the polynomial
basis functions at the $N_{\calG}$ selected points and $\bp_c$ are the
vector with the unknown coefficients. The least squares solution of (\ref{eq:regression}) is
the one that minimizes $||\Psi \bp_c - \by_c ||^2$ and is given by 
\begin{equation}
\hat{\bp}_c = (\Psi^T \Psi)^{-1} \Psi^T \by_c
\end{equation}
provided that $(\Psi^T \Psi)^{-1}$ exists. This requires that $N_\calG
\geq N_{\bp}$ so that our system will be overdetermined. Again, due to
the computationally demanding TOUGH2 forward simulator it is
impractical to obtain tens or hundreds of thousand samples so we work with
an ensemble of $N_{\calG} = 1000$ samples which, even on the massively
parallel version of TOUGH2 (TOUGH2-MP), with a moderate number of
processors used, takes around one week to
be generated. This choice of $N_{\calG}$ allows us to
calculate the coefficients of an expansion of order up to
$r = 5$.  The points $\{\bxi^{j}\}_{j=1}^{N_{\calG}}$ were randomly selected using Latin
Hypercube sampling. In fact, the $5$-order polynomial chaos expansion
included $462$ coefficients. This is close to $N_{\calG} \approx
2N_{\bp}$ which has been suggested as a good choice 
\cite{hosder}. New results concerning the stability and $L^2$-convergence of Polynomial
Chaos approximations obtained via least-squares solutions have been
recently reported by \cite{cohen, doostan}, however such an
analysis falls beyond the scope of this work.

\subsubsection{Goodness of fit and truncation error}
The next step after obtaining the expansion coefficients is to test
how well $\hat{\calG}_r$ performs as a surrogate. For validation purposes, we estimate the
coefficients for $r = 1, 2, 3, 4$ and $5$ using all $1000$
samples. We want to test how \emph{well} the least squares solution
fits the samples but also how close $\hat{\calG}_r$ is to $\hat{\calG}$ in the $L^2$
sense.

 For the first test we employ the common $R^2$ statistic known also as
\emph{coefficient of determination} \cite{seber} and provides a goodness-of-fit
measure for our linear model. The
statistic is defined as 
\begin{equation}
R^2_c = 1 - \frac{RSS_c}{\sum_{j=1}^{N_{\calG}} (y_c^i - \bar{y}_c)^2 }
\end{equation}
where the residuals sum of squares is $RSS_c = \sum_{j=1}^{N_{\calG}}
(y_c^{j} - y_c^{r,j})^2$, $c = 1,..., 24000$.

For the second test, we define the expected relative truncation error as 
\begin{equation}
e_c = \frac{\E[ |\calG_c(\bxi) - y_c(\bxi)|^2]}{ \E[|\calG_c(\bxi)|^2]} =
\frac{\int_{[0,1]^n} |\calG_c(\bxi) - y_c(\bxi)|^2p(\bxi) d\bxi}{
  \int_{[0,1]^n}|\calG_c(\bxi)|^2 p(\bxi)d\bxi}, \ \ \ c = 1,..., 24000
\end{equation}
where $p(\bxi) = 1$ is the joint distribution of independent
$\calU(0,1)$'s. The above can be estimated as
\begin{equation}
e_c \approx \frac{ \sum_{j=1}^{1000}|\calG_c(\bxi^j) -
  y_c(\bxi^j)|^2}{ \sum_{j=1}^{1000}|\calG_c(\bxi^j)|^2}, \ \ \ c =
1, ..., 24000.
\end{equation}

\begin{figure}[tbh]
\centering
\includegraphics[width = 0.4\textwidth]{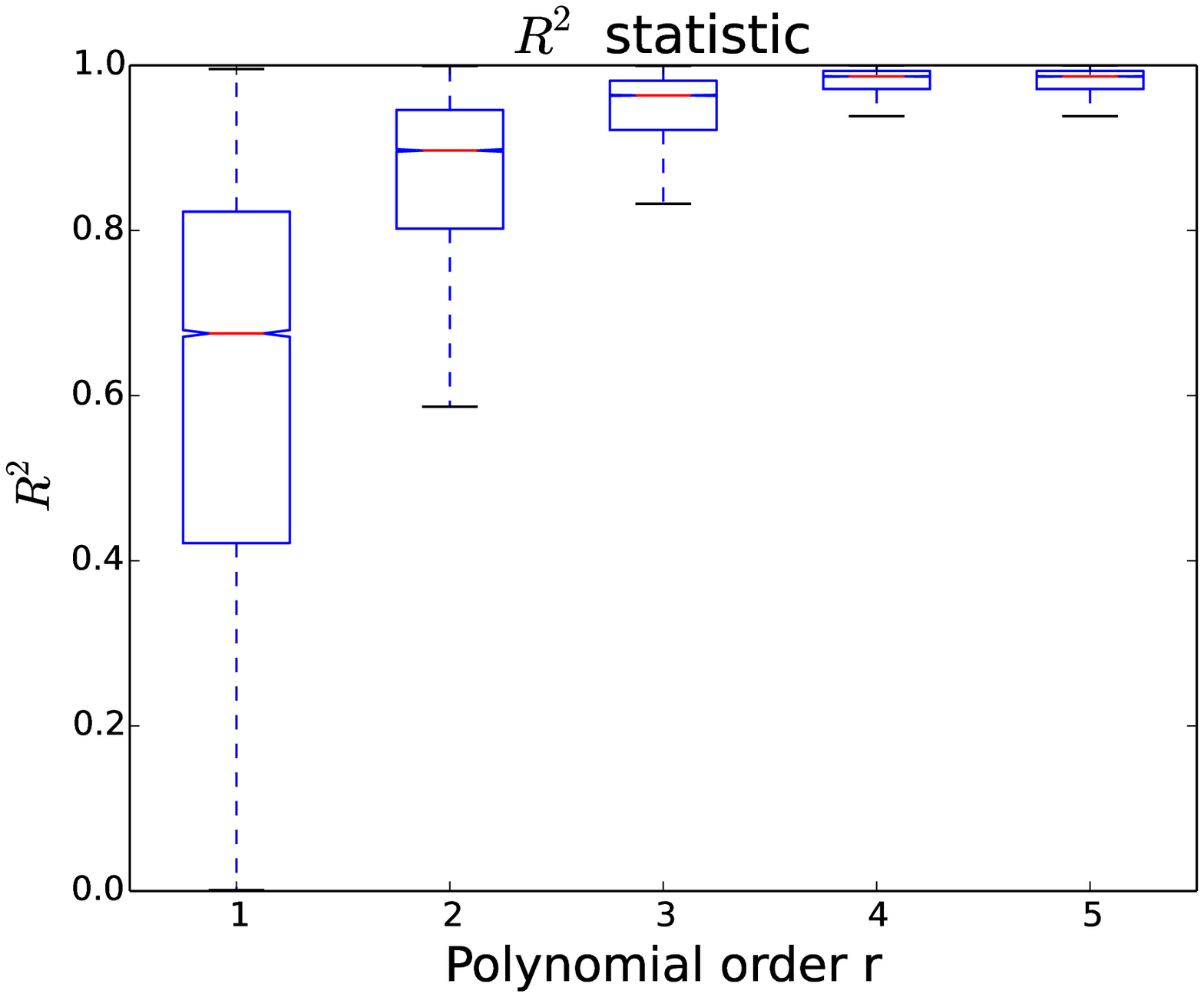}
\includegraphics[width = 0.4\textwidth]{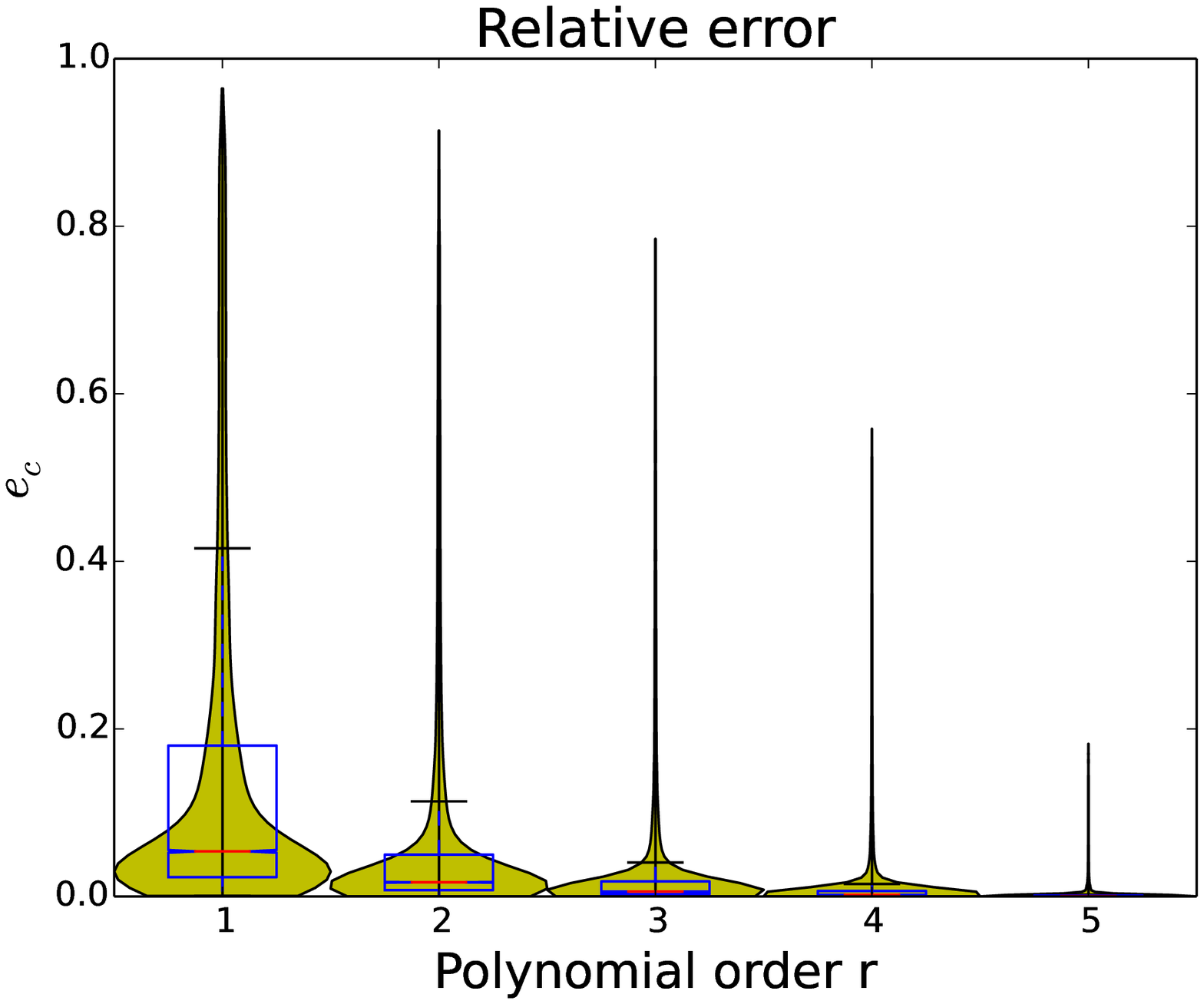}
\caption{Boxplots of the $R^2$ statistic (left) and  relative errors (right) of the polynomial chaoses
  for all components of the output $\calG(\bxi)$, of order $r = 1, ..., 5$. \label{fig:boxplots}}
\end{figure}

Fig.~\ref{fig:boxplots} shows the
boxplots of the $R^2$ statistic, obtained from all components $\calG$,
of the expansions of order $r = 1,...5$ (left) and the boxplots of the relative errors
for each order $r = 1, ..., 5$ and for all the components of
$\calG$ (right). Regarding the $R^2$ statistic and particularly for $r=5$ the
median is $0.986$ and the lower quartile is above $0.96$ giving us
a good fit for the $75\%$ of the expansions along the domain. A
thorough look showed that the remaining outputs including outliers for which
the fit is not stable correspond to points of the domain that are in
the bottom layer and along the
east boundary. Regarding the expected truncation error and
specifically for the expansions of order $5$ the median is
$3.1\cdot 10^{-4}$ and the upper quantile is $10^{-2}$ giving us good
approximations in the $L^2$ sense. Again the truncation error
increases for the points in the bottom layers. To ensure that our
experimental design methodology implemented in the next paragraph is
unaffected by possible instabilities of the Polynomial Chaos
expansions, the model outputs produced for the bottom layer of our
domain were not involved in our study.

\subsection{Experimental design}
\label{sec:design}

\subsubsection{Design settings}

As mentioned above, our main goal is to perform Bayesian inference on
the permeability parameters that will provide us a tighter
posterior that will decrease the uncertainty about the true values of
the soil permeability. This will further enhance our current knowledge
about the plume location and extent that can potentially result in
developing cost-efficient remediation strategies. In order to achieve
such a goal, we are seeking the best locations from where data should
be collected by employing the expected information gain lower bound as
our design criterion.

Unlike the analysis of the nonlinear algebraic model it is clearly
understood that in this case, design of a single experiment, that is
design for the collection of a single datum would not
have important effects in the inference procedure and it is never
performed in practice. In our setting we consider that for any
location in the $(x,y)$-plane we can observe the concentrations $y$
from the first $5$ layers of our domain, that is up to $10$m. depth,
so $5$ data points are available from each location. Without any loss
of generality in our method we choose to find the design of
experiments consisted of the $5$ best locations in the $(x,y)$-plane
from where data will be collected simultaneously to be used for
inference. This is a moderate choice which appears to be satisfactory
in order to validate our method. Further evaluation of the optimal number of data
points is beyond the scope of the present study. The design parameters
therefore are $\bd = (d_{1,x}, d_{1,y}, d_{2,x}, d_{2,y}, ..., d_{5,x},
d_{5,y})$ where $(d_{i,x}, d_{i,y}) \in \calD_0 =  [0, 600] \times
[0,400]$, $i = 1,..., 5$ and the design space for our problem is
$\calD = \calD_0^5 $. 

The observations $\by$ are subject to additional measurement noise
as indicated from our Gaussian likelihood function $\calN( \by |
\hat{\calG}(\bxi), \bSigma)$ involved in the expected information gain
lower bound derivation. Here we have substituted our forward
solver $\calG(\bk, \bd)$ with the polynomial chaos surrogate
$\hat{\calG}(\bxi)$. The covariance matrix has been chosen to be
$\bSigma = \sigma^2 \bI_{25}$ with $\sigma^2 = 10^{-2}$. This is a
rather simple choice that guarantees that measurement errors are
independent. A more sophisticated choice would be to take the standard
deviation to be proportional to the observable quantity $\sigma =
a\hat{\calG}_c(\bxi, \bd)$ where $a$ is some proportionality factor. We
avoid this choice as it implies the dependence of the error on the
design parameters which contradicts our
assumptions for the lower bound derivation.

\begin{figure}
\centering
\includegraphics[width = 0.49\textwidth]{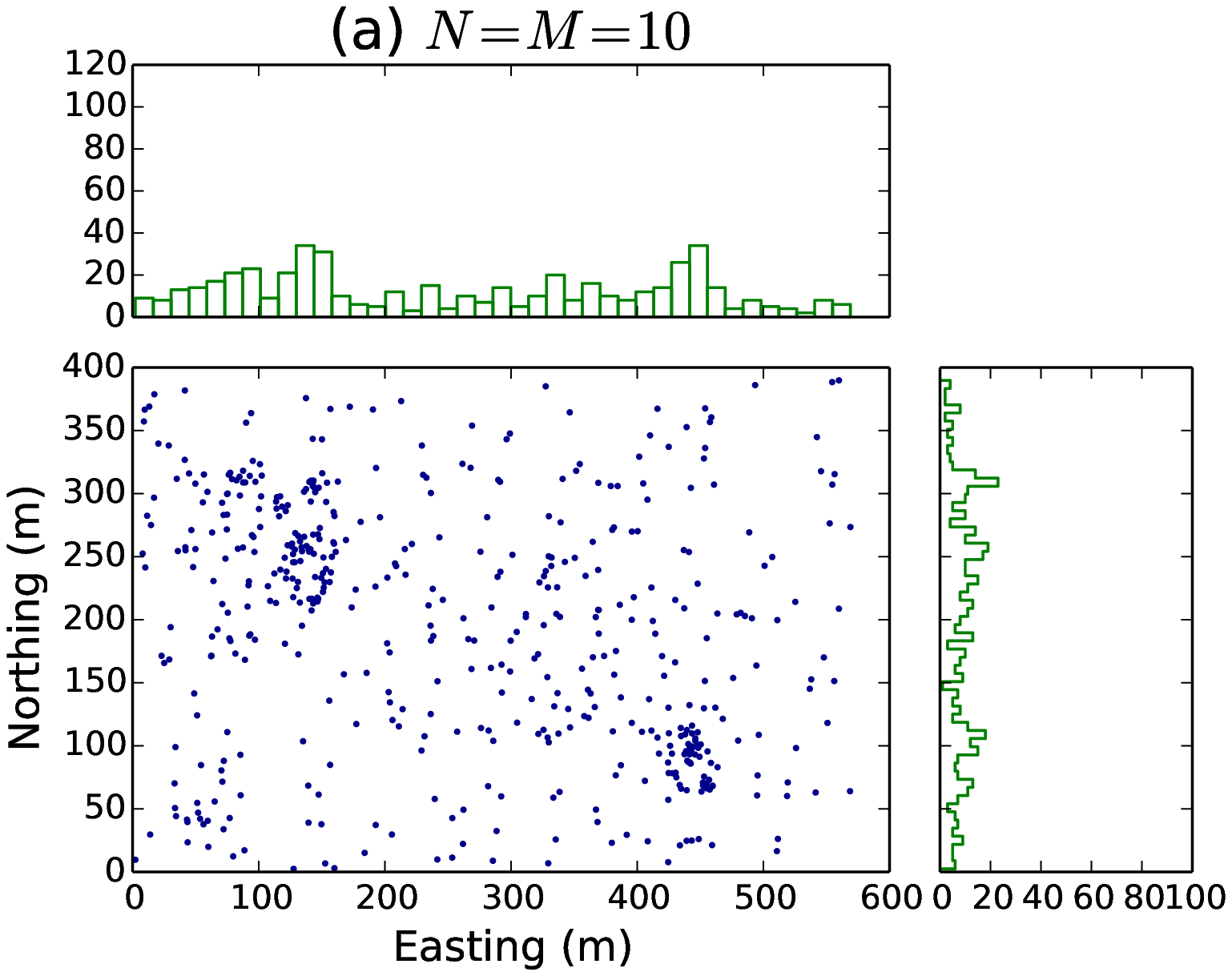}
\includegraphics[width = 0.49\textwidth]{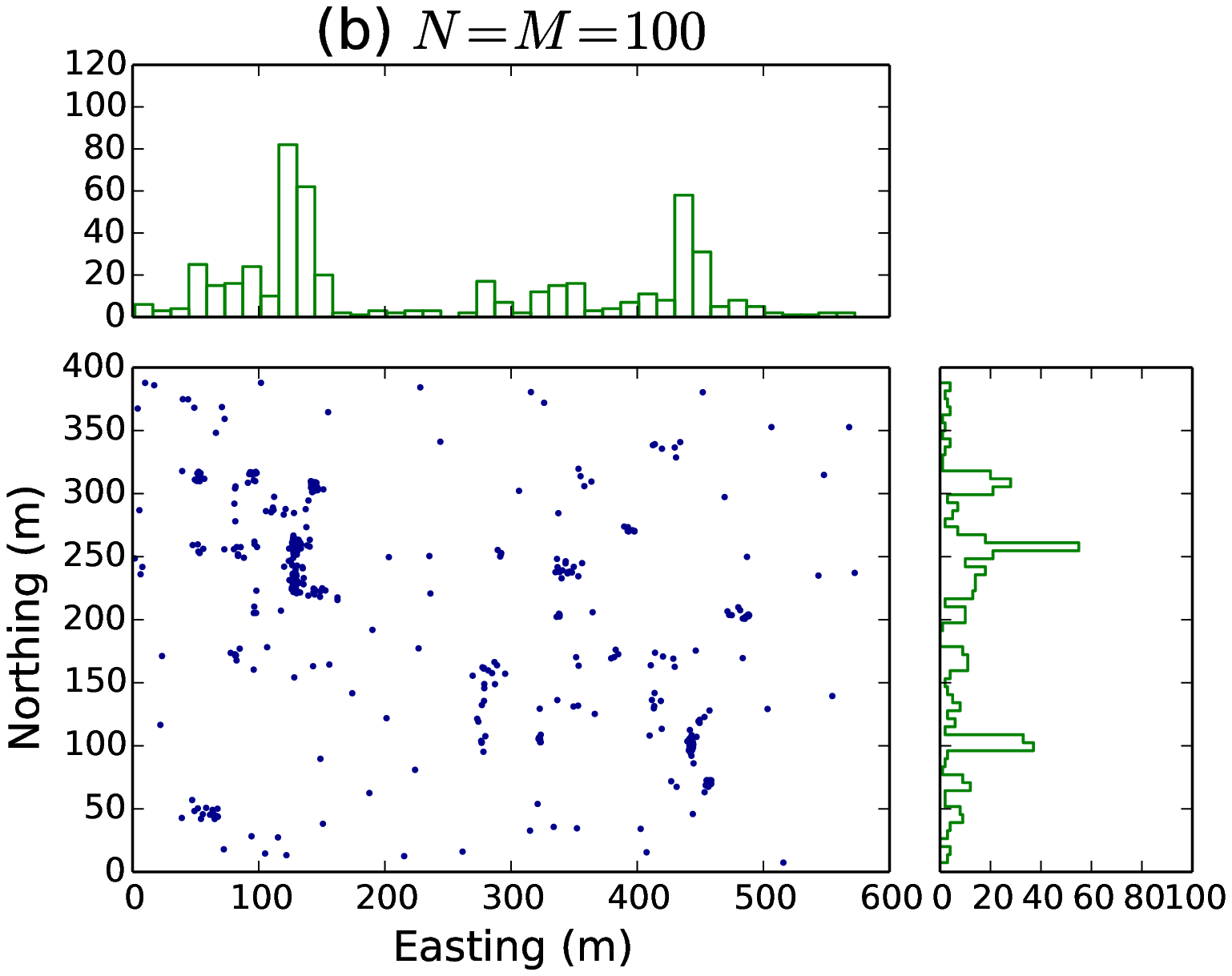}
\includegraphics[width = 0.49\textwidth]{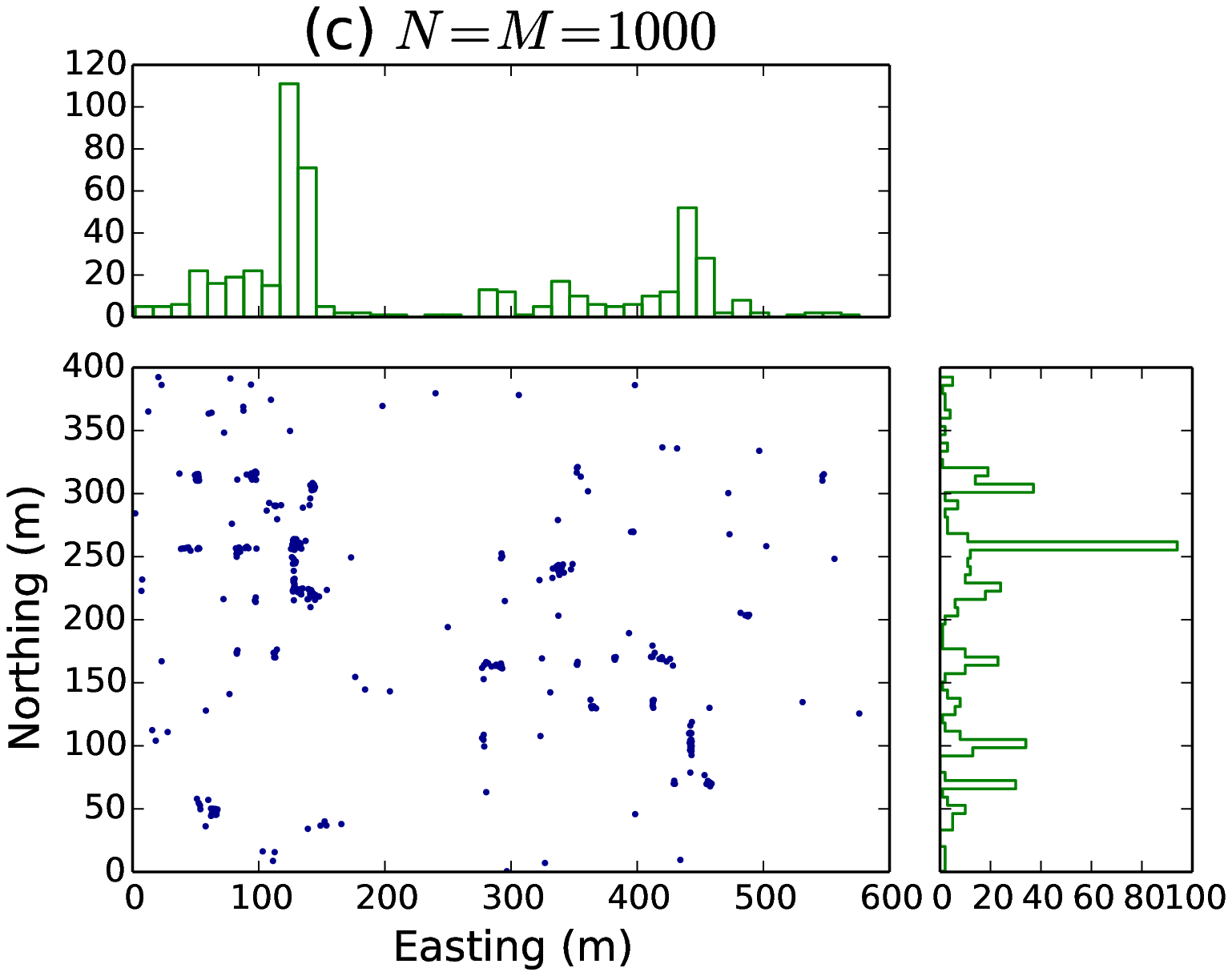}
\caption{Scatter plots of the optimal points $d_i$ for all optimization
  solutions $\bd^*$ of the SPSA algorithm. The red transparent areas
  denote the locations of the sources, plotted for comparison of the patterns. \label{fig:scatter_plots}}
\end{figure}

\subsubsection{Results}

We run the SPSA algorithm in order to find $\bd$ that
minimizes $\hat{U}_L^*$, therefore maximizes the lower bound
estimate. To assess the performance of the algorithm, we run several
cases where the choices of $N$ and $M$ in the Monte Carlo loop
vary. It is easy to see that the variance of our estimator is 
\begin{equation}
var\left[ \hat{U}^*_L(\bd) \right] = \frac{1}{NM}var\left[ p(Y| \bk, \bd)\right]
\end{equation}
which shows that $N$ and $M$ have the same contribution in controlling the
variance, therefore we take them equal. As stated previously, our estimator
is unbiased, unlike the dlMC estimator, so an extremely large value
for $M$ is not necessary as it is for dlMC to maintain a low bias. We
considered three different cases where $N = M = 10, 100$ and
$1000$. Since the results of SPSA are noisy, in order to better
evaluate the performance of the algorithm we obtain the results of
$100$ independent runs for each case. For the first two cases we run
the algorithm for a total of $10^{4}$ evaluations of the objective 
function and for the last case we run for $5\cdot 10^{3}$
evaluations. It appears that these choices are rather high since the
iterates are stabilized much earlier maintaining a low iteration
error. 

The resulting $5$-tuples of points in $\calD_0$ are shown all together
in a total of $500$ points in Fig.~\ref{fig:scatter_plots}, for all three
cases. This gives us a qualitative idea about where the expected
information gain is maximized and so data should be collected
from. The formation of a certain pattern as $N$, $M$ are increased
is obvious and the optimal points appear to be very close to where the contaminant
sources were placed. At the same time various points that appear to be
outliers are also present, a characteristic behavior of the SPSA
algorithm. The case $N = M = 10$ particularly shows all points widely
spread all over $\calD_0$ and one can only distinguish the two areas around
$(x, y) = (150,250)$ and around $(x,y) = (450, 100)$ where more points are
accumulated. The cases $N = M =
100$ and $N = M = 1000$ provide very similar conclusions with the
majority of the runs having converged at very specific locations, close
to the actual contaminant sources, forming a clear pattern with the
main difference that in the third case the points are even more concentrated
and less outliers are present. 

\begin{figure}
\centering
\includegraphics[width = 5.5cm]{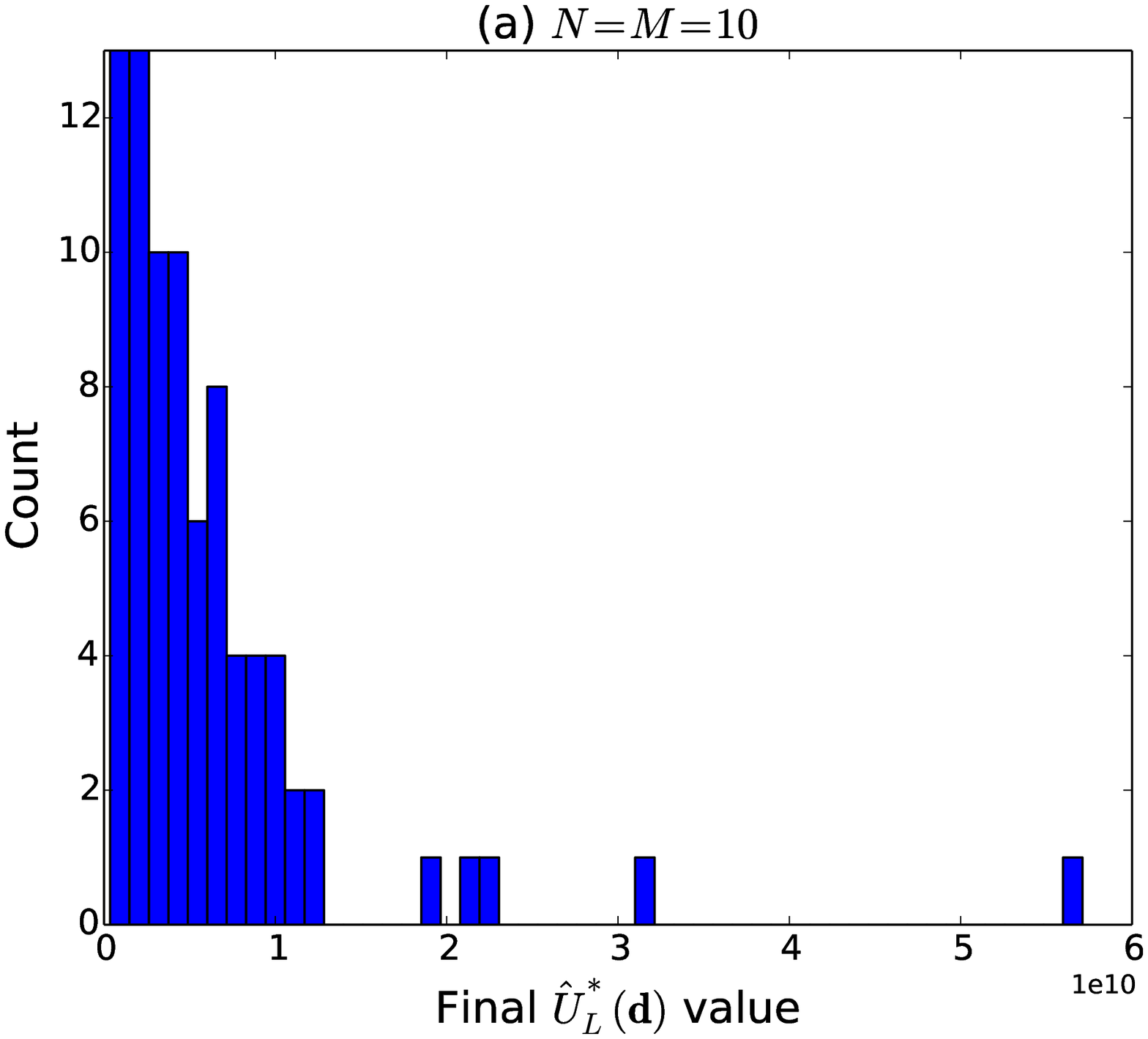}
\includegraphics[width = 5.5cm]{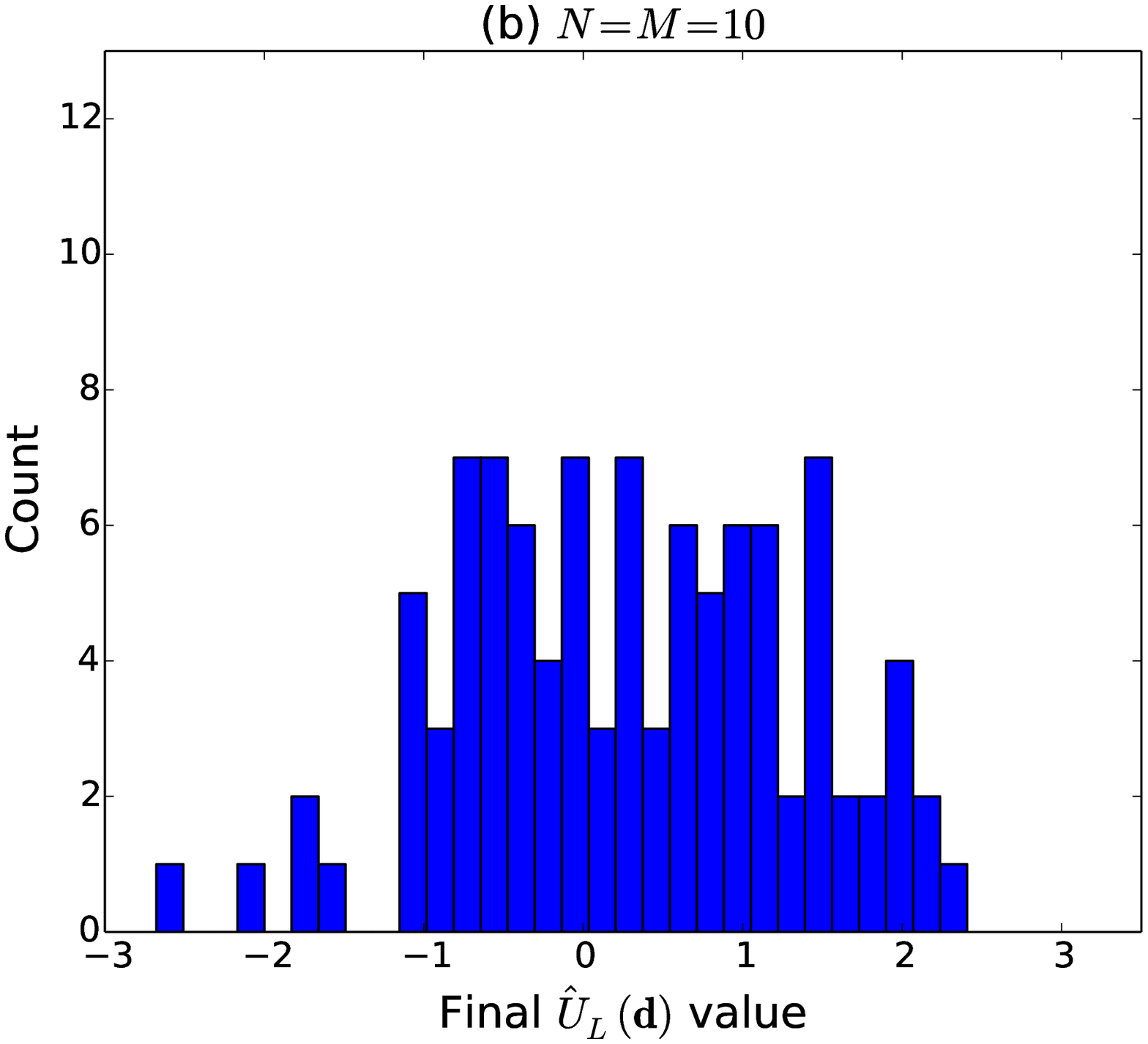}
\includegraphics[width = 5.5cm]{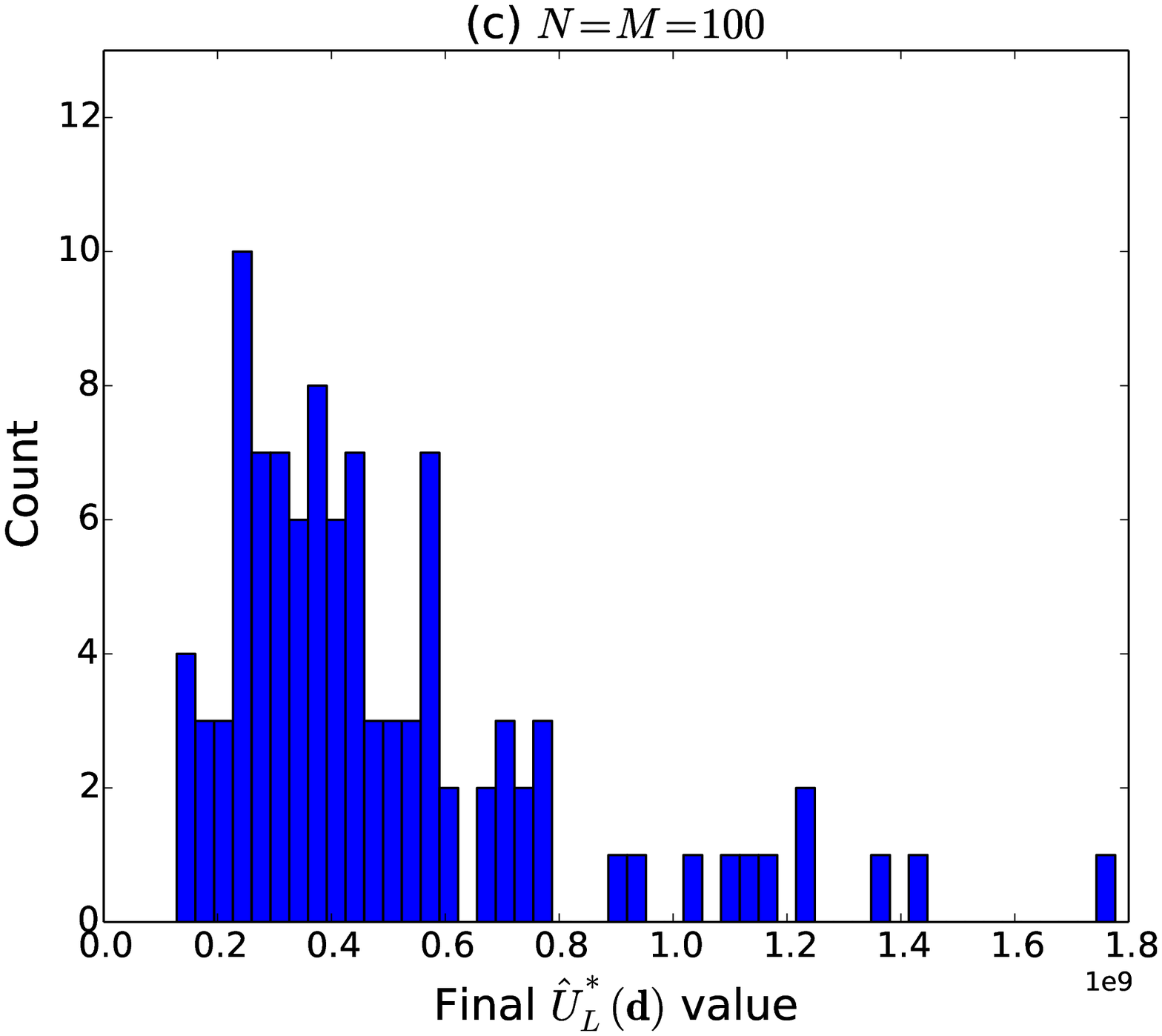}
\includegraphics[width = 5.5cm]{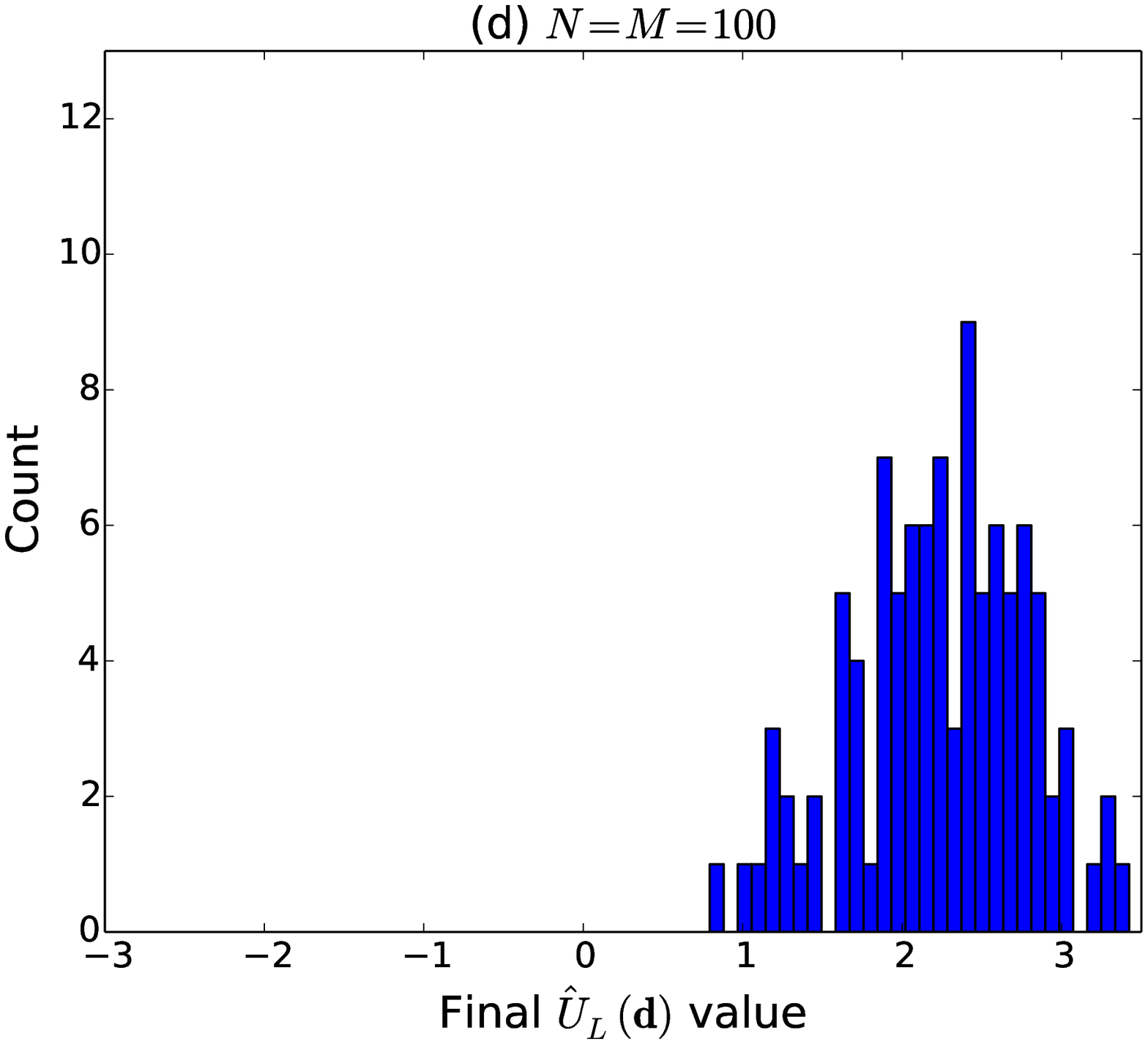}
\includegraphics[width = 5.5cm]{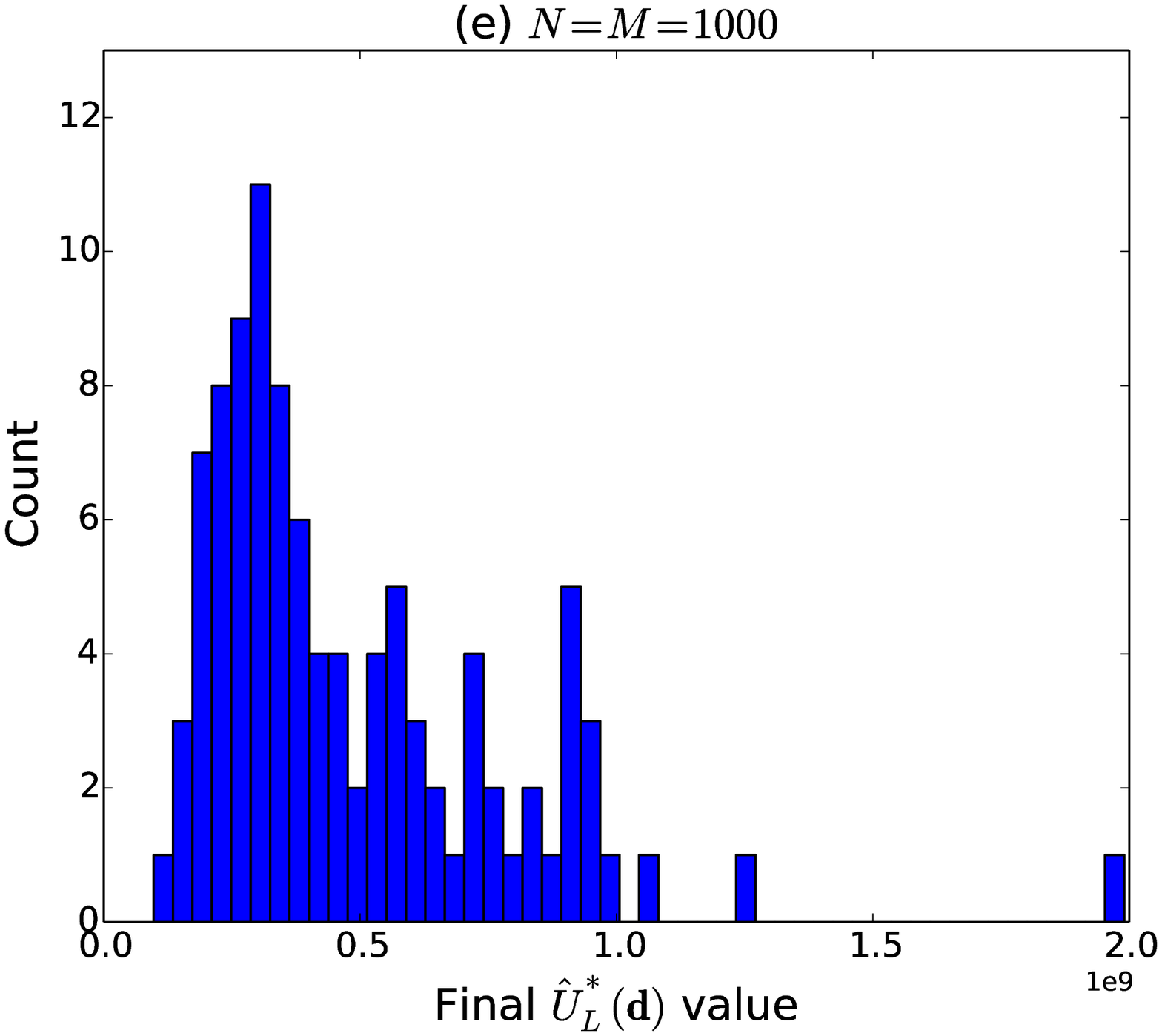}
\includegraphics[width = 5.5cm]{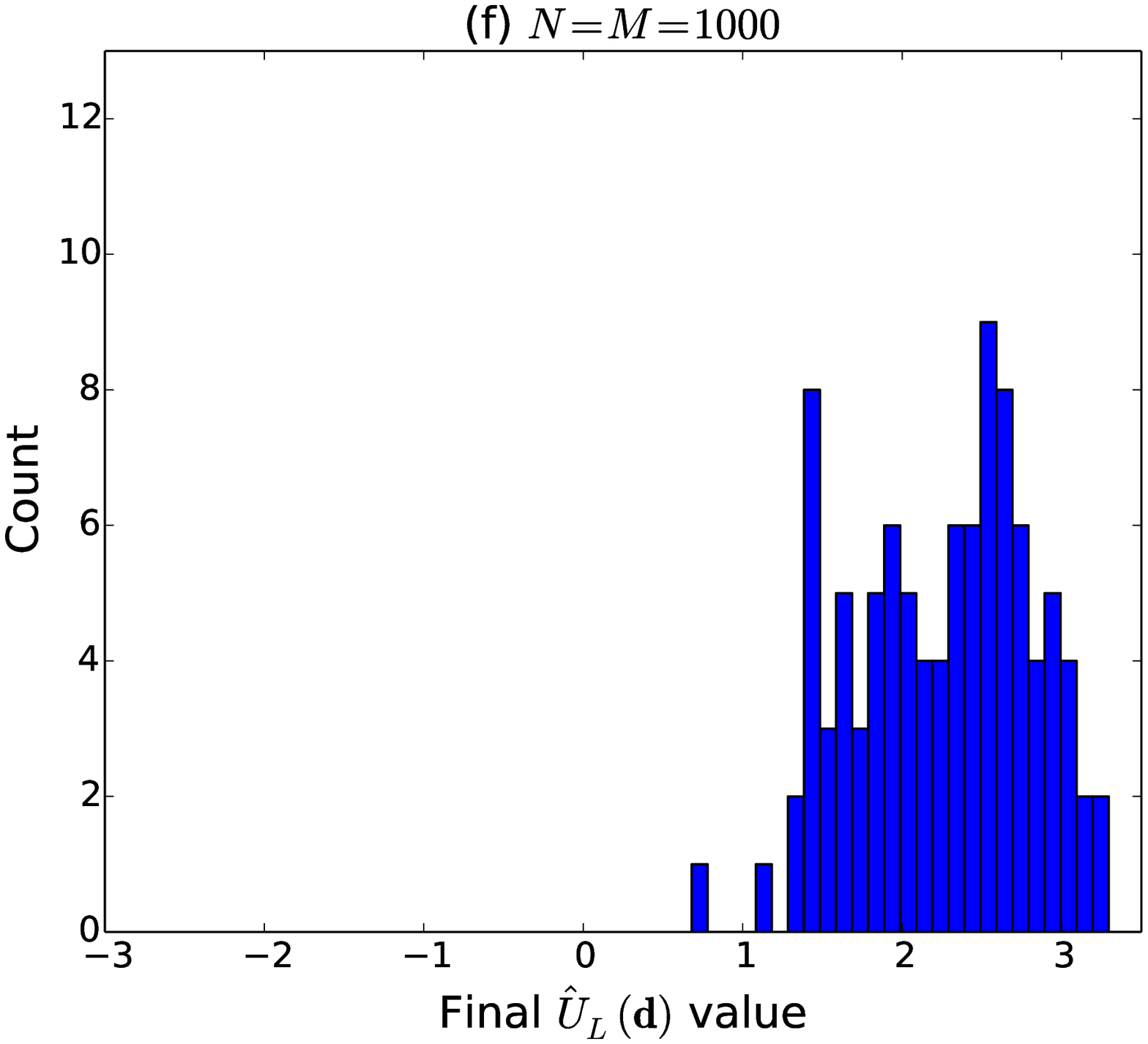}
\caption{Histograms of the $\hat{U}^*_L(\bd)$ (left) and
  $\hat{U}_L(\bd)$ (right) values for all optimal
  points $\bd^*$. \label{fig:histograms}}
\end{figure}

For a more quantitative argument we also present the exact values of
$\hat{U}^*_L(\bd)$ as well as those of 
\begin{equation}
\label{eq:u_hat_lower}
\hat{U}_L(\bd) =  -\frac{1}{2}\left\{ m + \log\left[ (2\pi)^m |\bSigma|\right]
  \right\} - \log\left[  \frac{1}{NM} \sum_{i,j = 1}^{N,M}
  p(\by^{i} | \bk^{ij}, \bd) \right]
\end{equation}
calculated at all outcomes $\bd^*$ of our runs. These values quantify the
information gain that each design is expected to offer and at the
end, if we had to choose only one design, this should be the design
that provides the largest value. This
time, for diagnostic reasons and in order to provide a measure of
comparison, we want to have a fixed
accuracy for the estimates and we
use the same number of samples $N = M = 1000$ for the evaluation of
the two objective functions on all points. Their values are all shown in
the histograms presented in Fig.~\ref{fig:histograms}. Again there is a significant difference
between the quality of the results of the first case with that of the last
two cases for both functions. For $\hat{U}^*_L(\bd)$, the first case
gives us designs that whose values vary in a range from $0.1\cdot
10^8$ to $6\cdot 10^{10}$ (observe that the $x$-axis of the histogram
is an order of magnitude larger than the rest) with a large number of
them being away from the minimum value. The impression one might get
for this case is that chances are few that the optimization will yield
an optimal design solution and not just an outlier. The other two
cases provide similar results with the third being, as expected, better
than the second, in the sense that the optimal points are even more
concentrated around the minimum value. For completeness we present
also $\hat{U}_L(\bd)$ which is
technically nothing else but the negative logarithm of
$\hat{U}^*_L(\bd)$, shifted by a constant. In the
first case, the values cover a range
from $-3$ to $2.5$ with a mean around $0$. The last two cases give us
again similar results with both
histograms covering values from around $1$ to $3.5$, a much narrower interval
than the first case. Again we can see that the third case is slightly better than the
second where we see that the unimodal structure of the histogram is more concentrated
on higher values close to $3$ which implies that a few more runs eventually
converged to the global maximum.  

\subsection{Bayesian update of $\log (\kappa)$}
\label{sec:inference}

In order to validate our methodology and demonstrate the
significance of the experimental design analysis developed above, we
perform Bayesian inference based on two different
designs that are selected according to the results of the previous
section. More precisely, data from two different sets of $5$ points,
denoted with $\bd_\circ$ and $\bd_\diamond$, is collected and used to update the
probability distribution of $\bk$. Specifically, $d_\circ$
is selected among the $100$ optimal designs that were approximated by SPSA
in the previous section, for $N = M = 1000$ and is the one that achieved the maximum
$\hat{U}_L(\bd)$ value, so based on the previous analysis it is the design
that is expected to yield the best posterior. The second design
$\bd_\diamond$, chosen for comparison only, was arbitrarily selected among the optimal designs that
were returned by SPSA using the poor estimate with $N = M = 10$ and
has a much lower $\hat{U}_L(\bd)$ value, therefore it is expected to
update to a wider posterior. The exact location of the points
for each set and their $\hat{U}_L(\bd)$ values are shown in
Table~\ref{tab:designs}.

\begin{table}
\caption{The $(x,y)$-coordinates of the 5-point designs $\bd_\circ$ and
  $\bd_\diamond$ and their $\hat{U}_L(\bd)$ values. }
\label{tab:designs}
\centering
\begin{tabular}{c | c c c c c | c }
& $d_{1}$ & $d_{2}$ & $d_{3}$ & $d_{4}$ & $d_5$ &  $\hat{U}_L(\bd)$\\
\hline
$\bd_\circ$ & $(127.96, 259.65)$ & $(442.59, 97.05)$ & $(143.22, 304.41)$ &
$(337.11, 279.02)$ & $(412.26, 135.86)$ &  $3.698$\\ 
$\bd_\diamond$ & $(22.95, 171.37)$ & $(75.36 , 300.00)$ & $(318.58, 169.27)$ &
$(425.06, 109.96)$ & $(441.00, 253.70)$ & $0.286$\\
\end{tabular}
\end{table}

Since the posterior distributions cannot be calculated explicitly, we
use \emph{Markov Chain Monte Carlo (MCMC)} methods to draw samples
from them. MCMC methods rely on generating a sequence of random
variables based on a Markov Chain that converges to a stationary
distribution $\pi(y)$, called the target distribution and therefore
the sequence generated after a burn-in
period, can be thought of as following the
stationary distribution $\pi$. To avoid strong autocorrelation among the samples,
appropriate thinning and tuning of the algorithm is required. A
powerful MCMC method is the Metropolis-Hastings (MH) algorithm that
was first proposed by Metropolis
\cite{metropolis} and later generalized by Hastings \cite{hastings}. The MH
algorithm at an arbitrary step generates a new sample $y$ from a proposal
distribution $q(x,y)$ given that $x$ is the last accepted sample and
accepts it with probability 
\begin{equation}
\alpha(x,y) = \min \left\{ 1 , \frac{\pi(y) q(x,y)}{\pi(x)q(y,x)}\right\}.
\end{equation} 

In our case we use the adaptive version of the MH algorithm as developed in
\cite{haario} with a Gaussian as the proposal distribution where its
covariance matrix is updated at each step taking into account all the previous
samples that have been drawn. This implies that the chain is
non-Markovian, however it has been proved that it has the correct
ergodic properties. The adaptive MH (aMH) algorithm provides faster
convergence to the target distribution and achieves lower autocorrelation among the
samples without very large thinning. 

Below we explore two cases where in the first data is observed from a
reference field that is associated with one specific realization of the model
output while in the second, data is observed from a reference field 
that is constructed using Gaussian process (GP) regression on
measurements taken from the real site. 

\begin{figure}[tbh]
\centering
\includegraphics[width = 6cm]{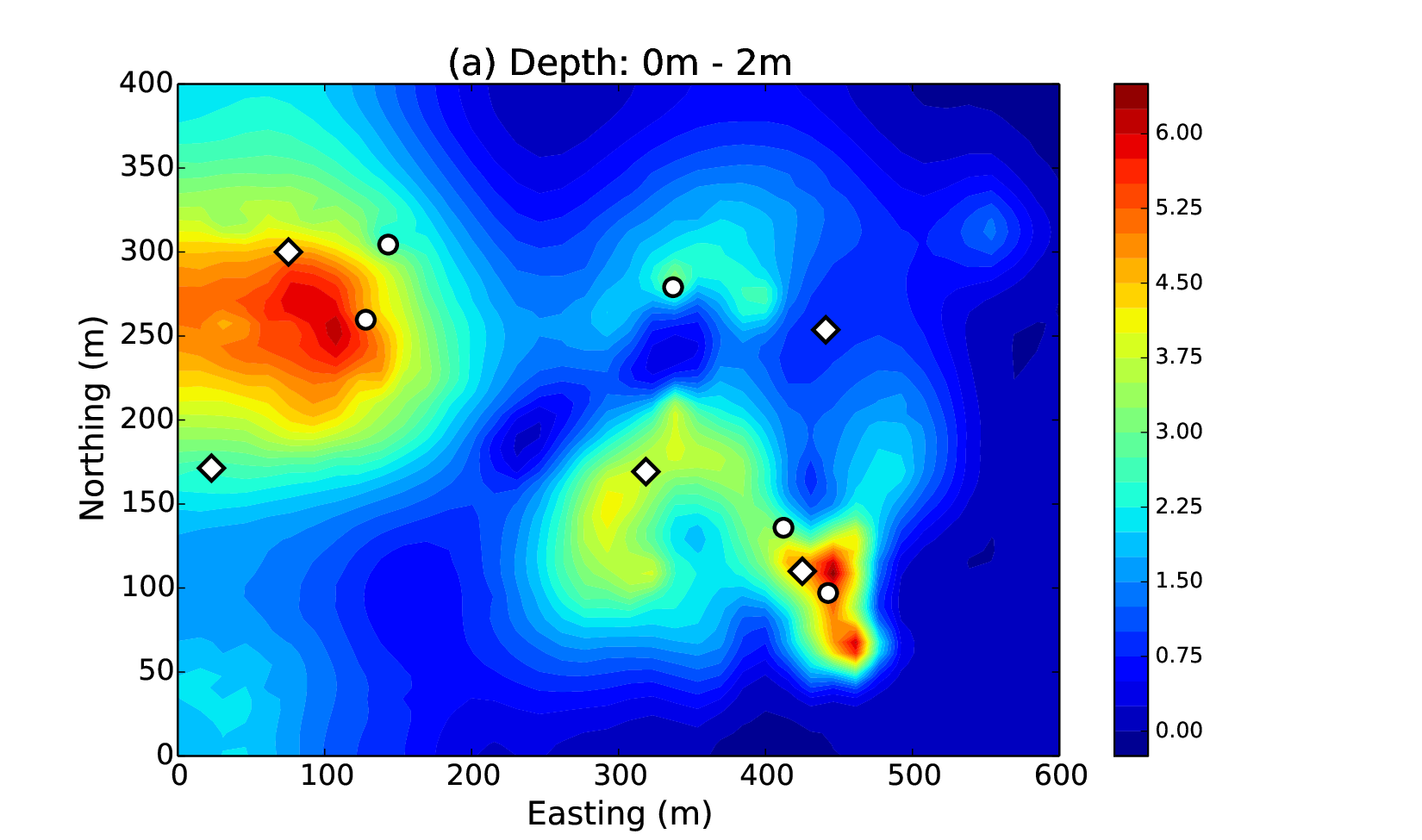}
\includegraphics[width = 6cm]{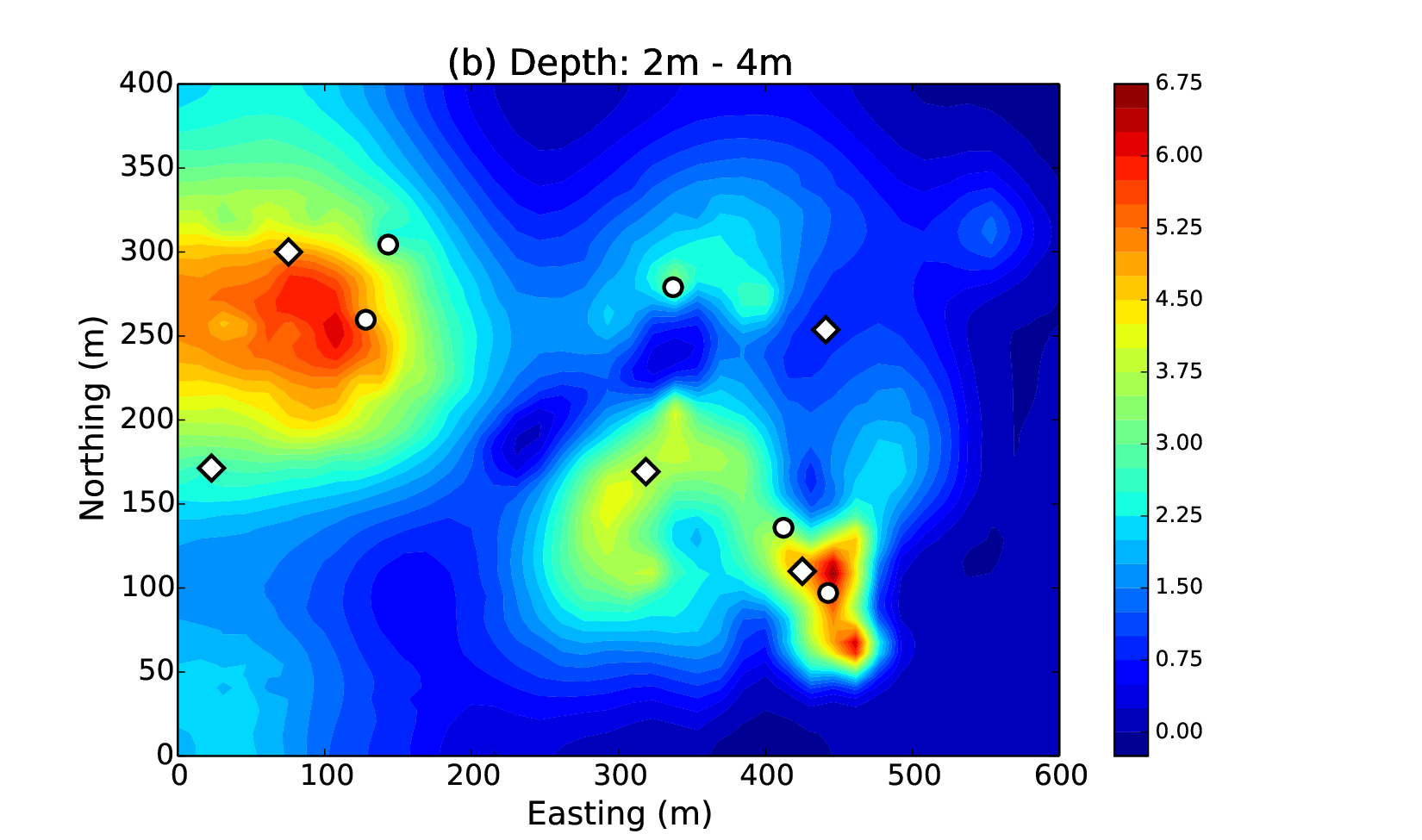}
\includegraphics[width = 6cm]{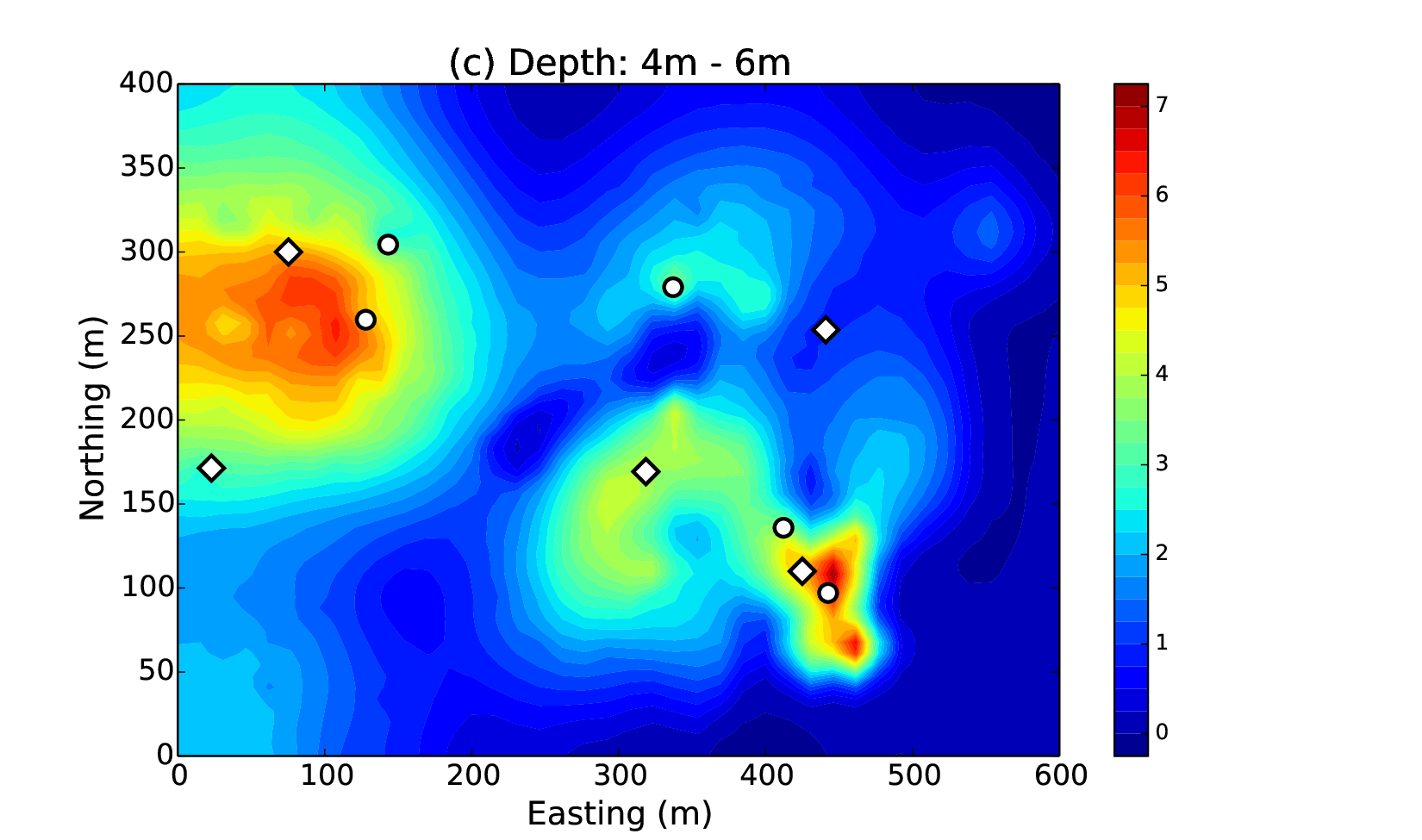}
\includegraphics[width = 6cm]{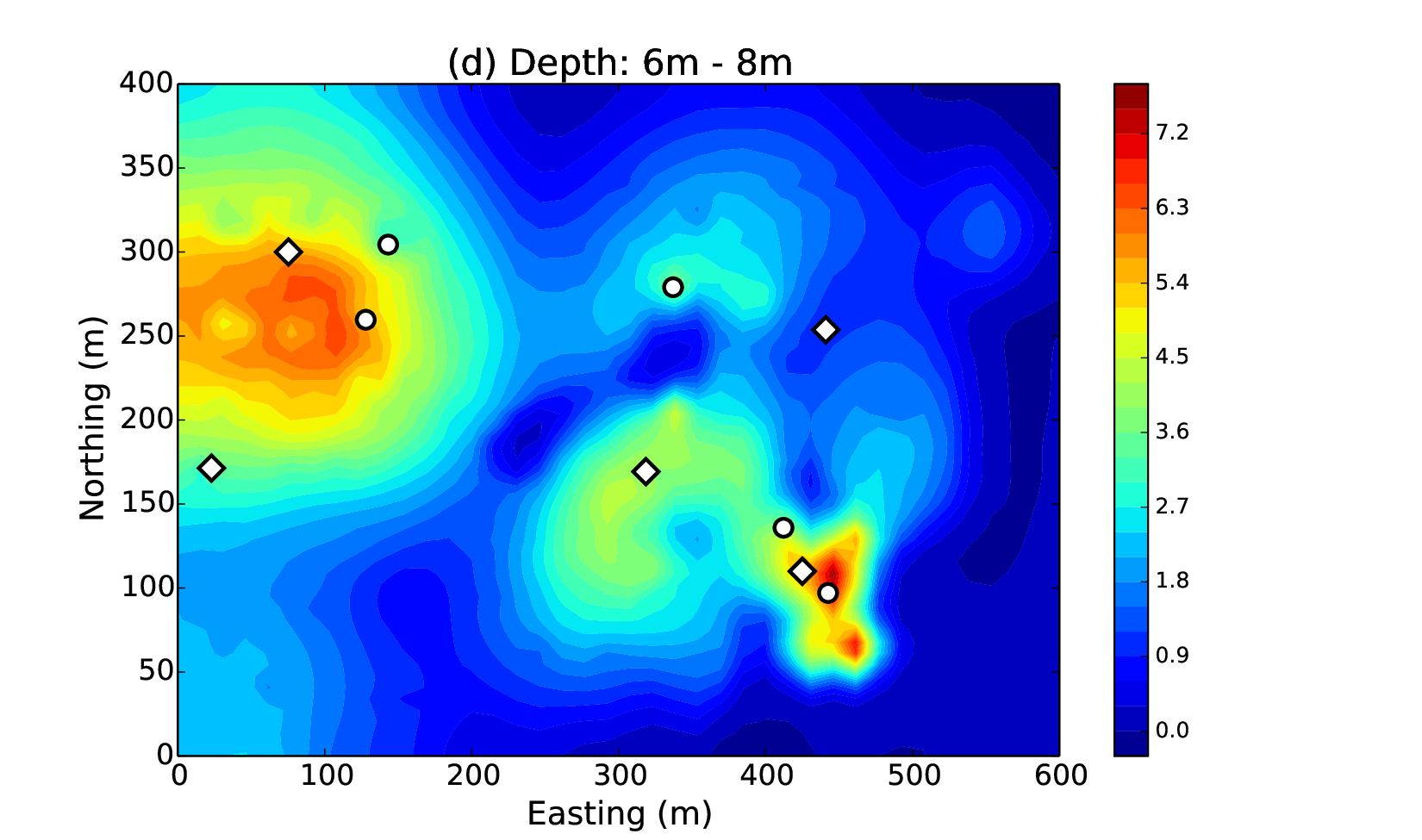}
\includegraphics[width = 6cm]{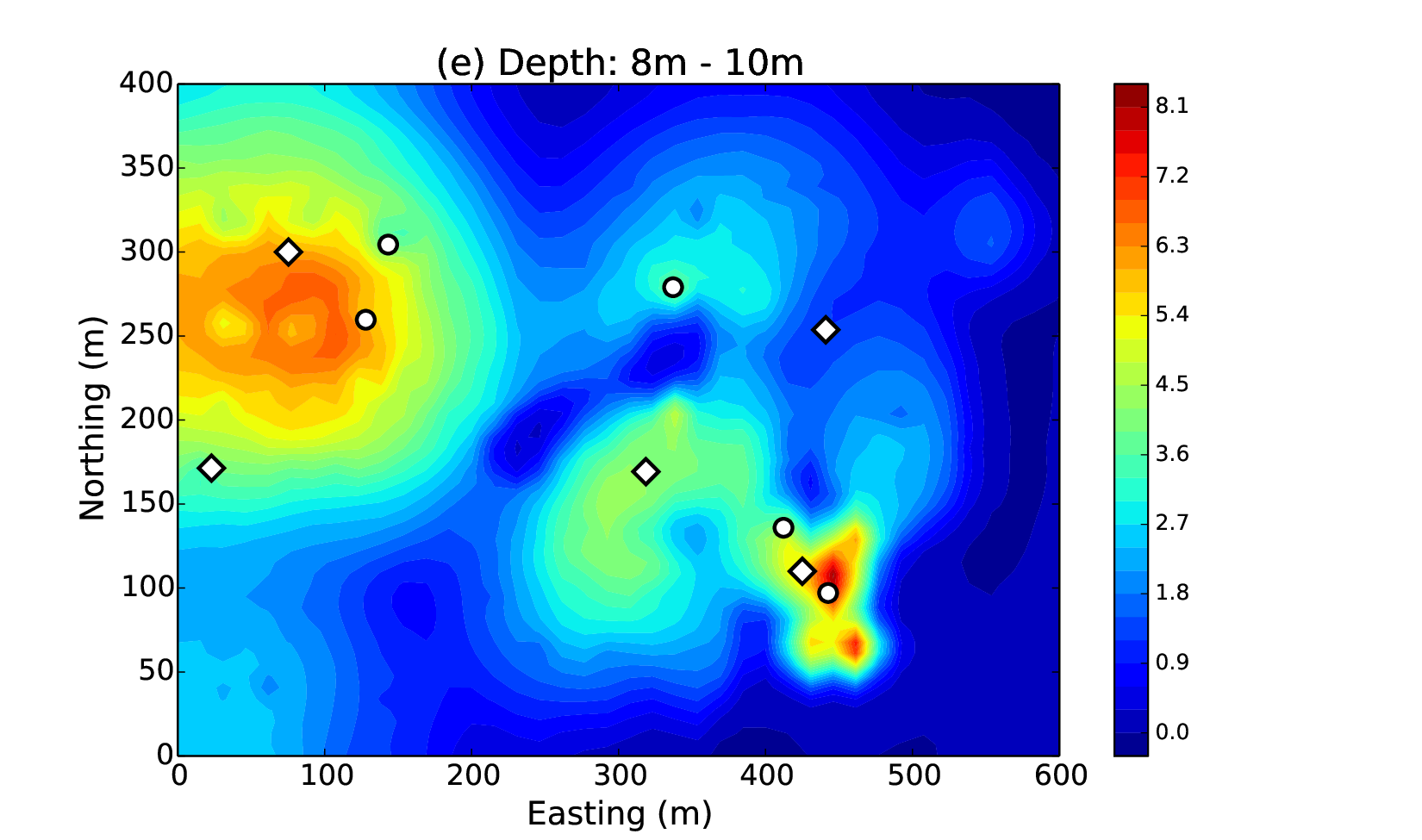}
\caption{Concentration map for the $5$ top layers of the domain used
  as the reference field in example 1. The '$\circ$' and  '$\diamond$' signs indicate the locations of
 $\bd_\circ$ and $\bd_\diamond$ designs respectively, from where
 measurements were taken. \label{fig:data_1}}
\end{figure}

\subsubsection{Case 1: Reference data generated from prior}

Here we assume a hypothetical situation where the real permeability
values are those displayed on Table~\ref{tab:logk}. Our model's output
for those values can be directly evaluated and the $5$ top
layers are displayed in Fig.~\ref{fig:data_1}. With this reference
field assumed to be our "reality'', we collect data from the locations
indicated at designs $\bd_{\circ}$ and $\bd_{\diamond}$ and use them
for our computations. 

\begin{table}
\caption{True values of $\bk$ and $\log(\bk)$ used for generating the
 data in case 1.}
\label{tab:logk}
\centering
\begin{tabular}{c | c | c}
$i$ & $\log\kappa_i$ & $\kappa_i$ (cm$^2$) \\
\hline
$1$ & $-22.195$ & $2.295\cdot 10^{-10}$\\
$2$ & $-20.791$ & $9.346\cdot 10^{-10}$\\ 
$3$ & $-21.567$ & $4.300\cdot 10^{-10}$\\
$4$ & $-25.193$ & $1.145\cdot 10^{-11}$\\
$5$ & $-25.272$ & $1.058\cdot 10^{-11}$\\
$6$ & $-24.610$ & $2.050\cdot 10^{-10}$\\
\end{tabular}
\end{table}

\begin{figure}[tbh]
\centering
\includegraphics[width = 5cm]{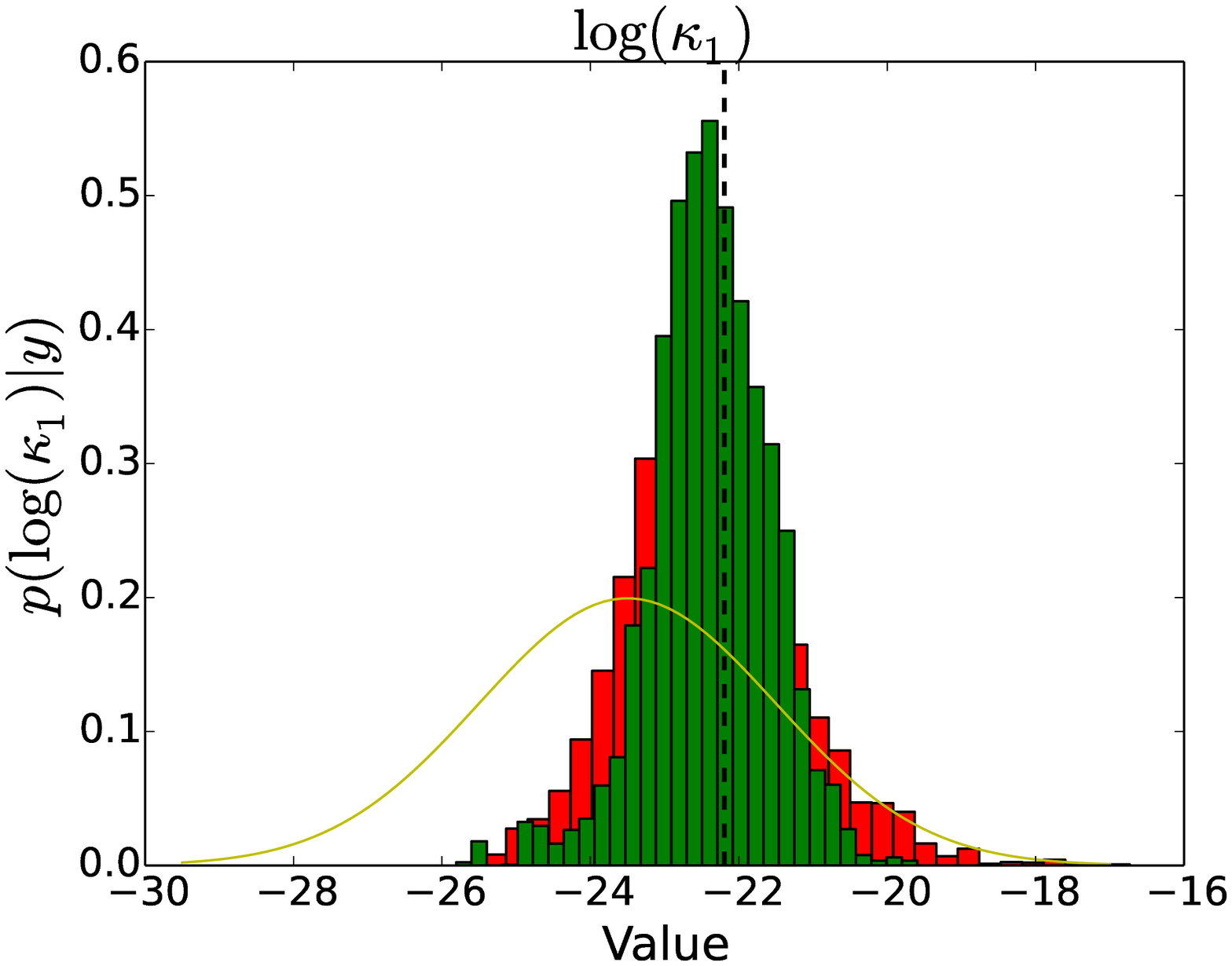}
\includegraphics[width = 5cm]{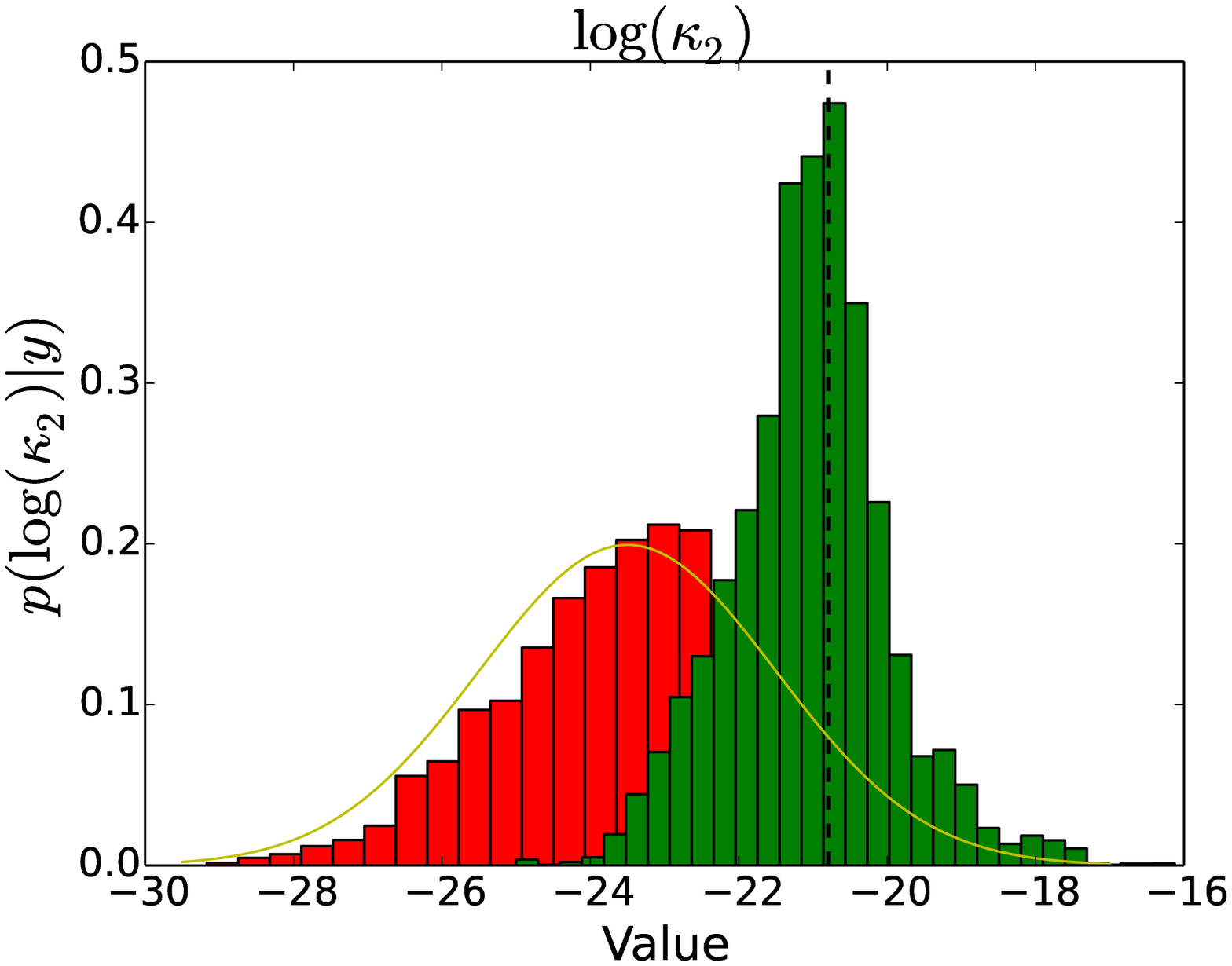}
\includegraphics[width = 5cm]{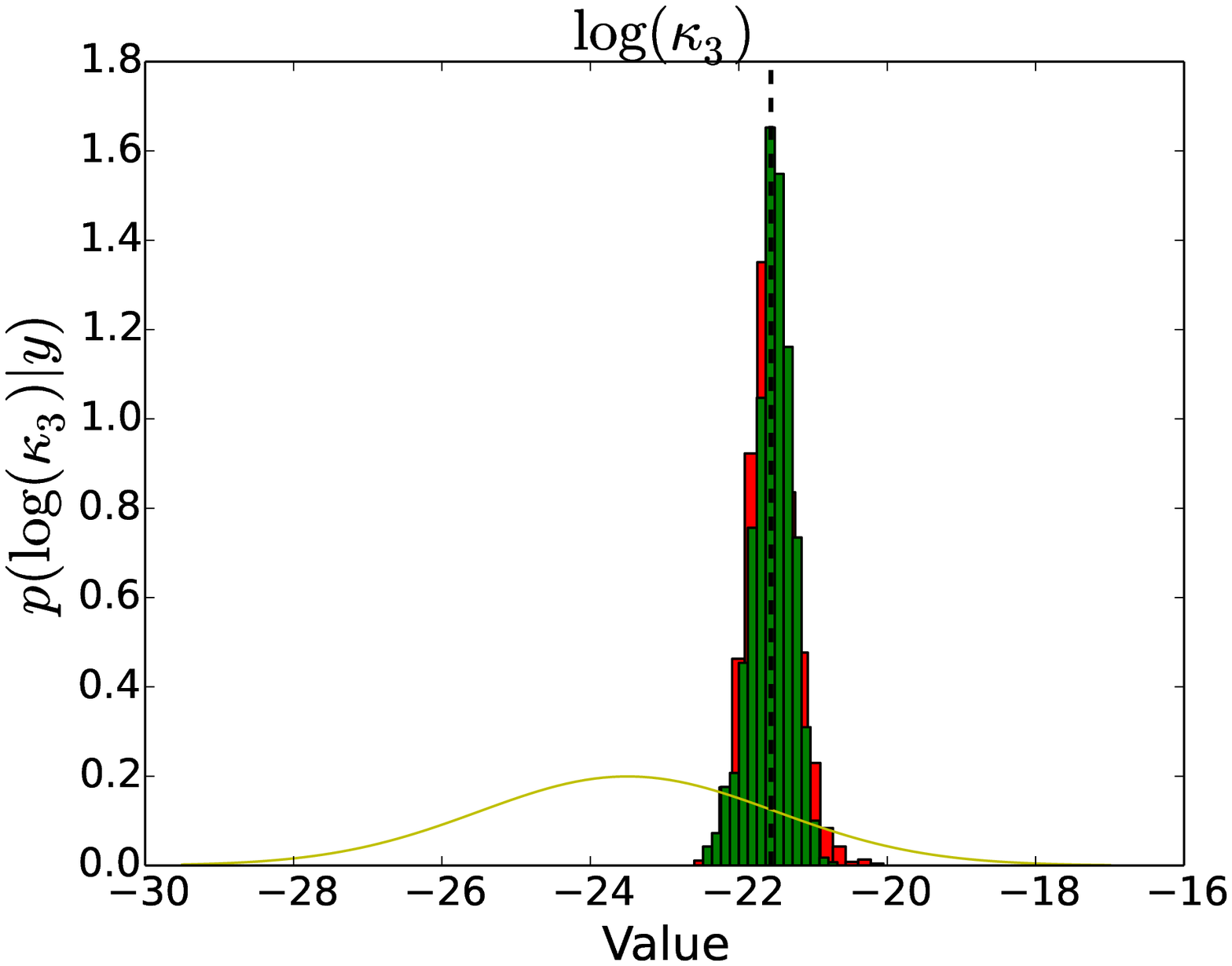}
\includegraphics[width = 5cm]{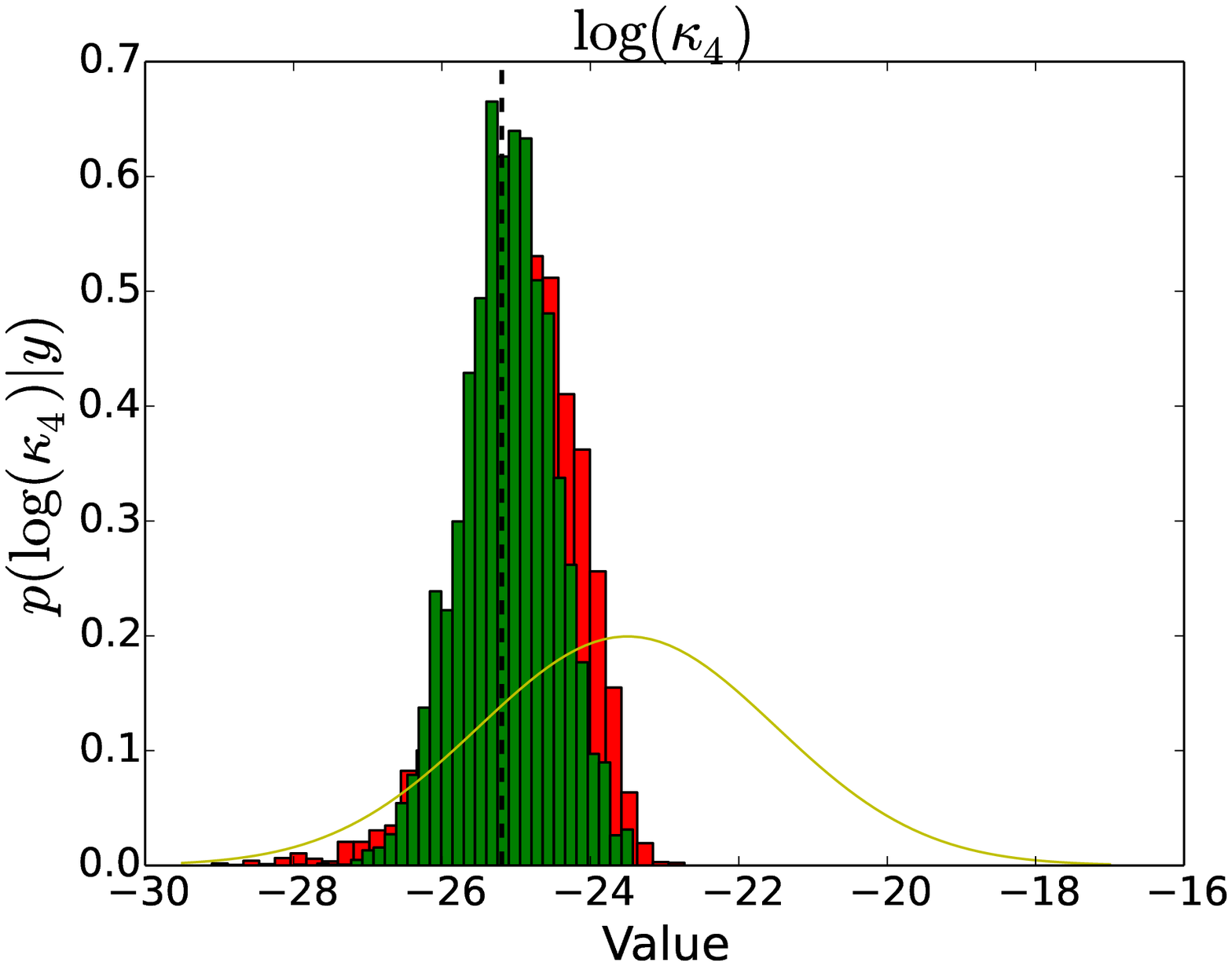}
\includegraphics[width = 5cm]{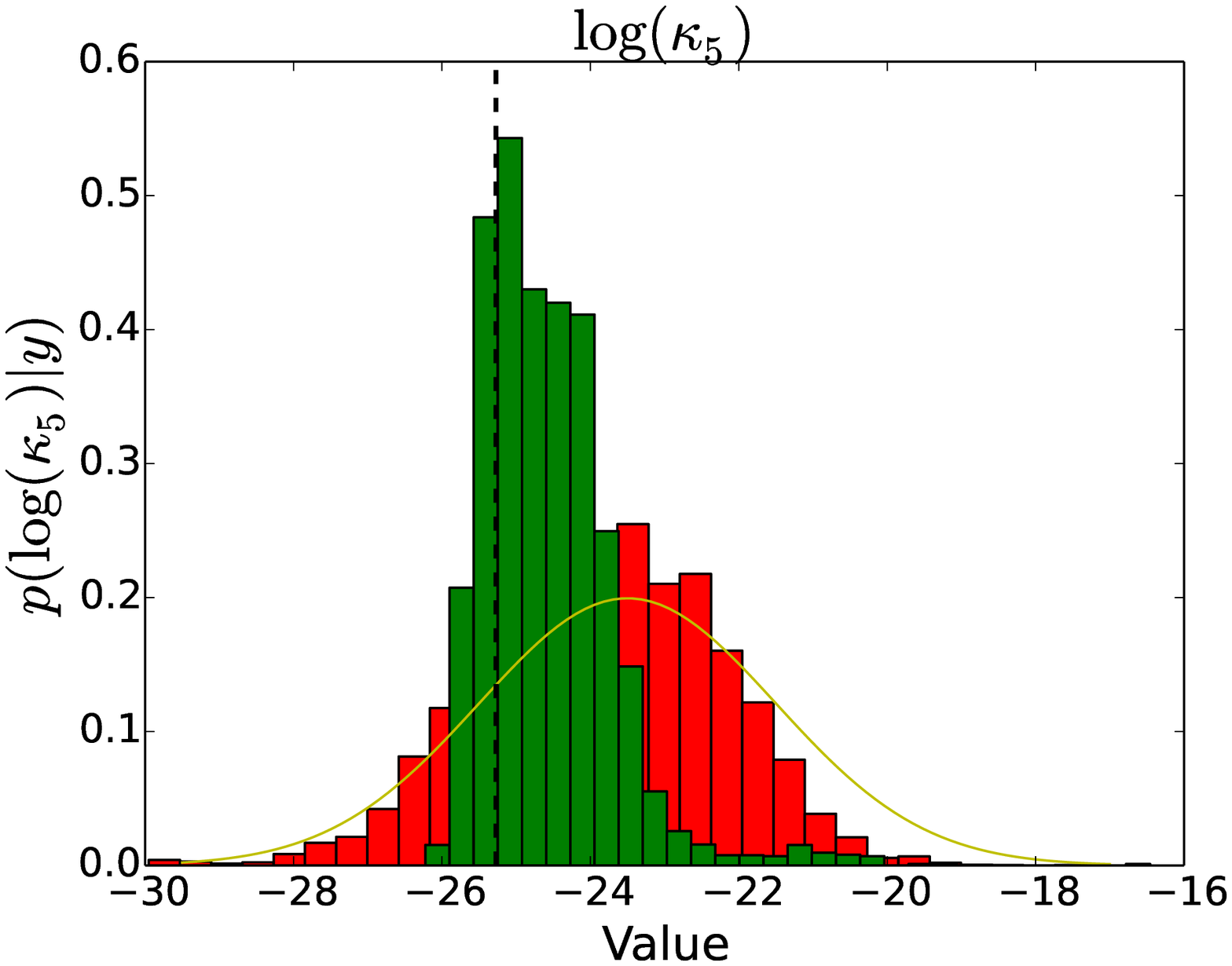}
\includegraphics[width = 5cm]{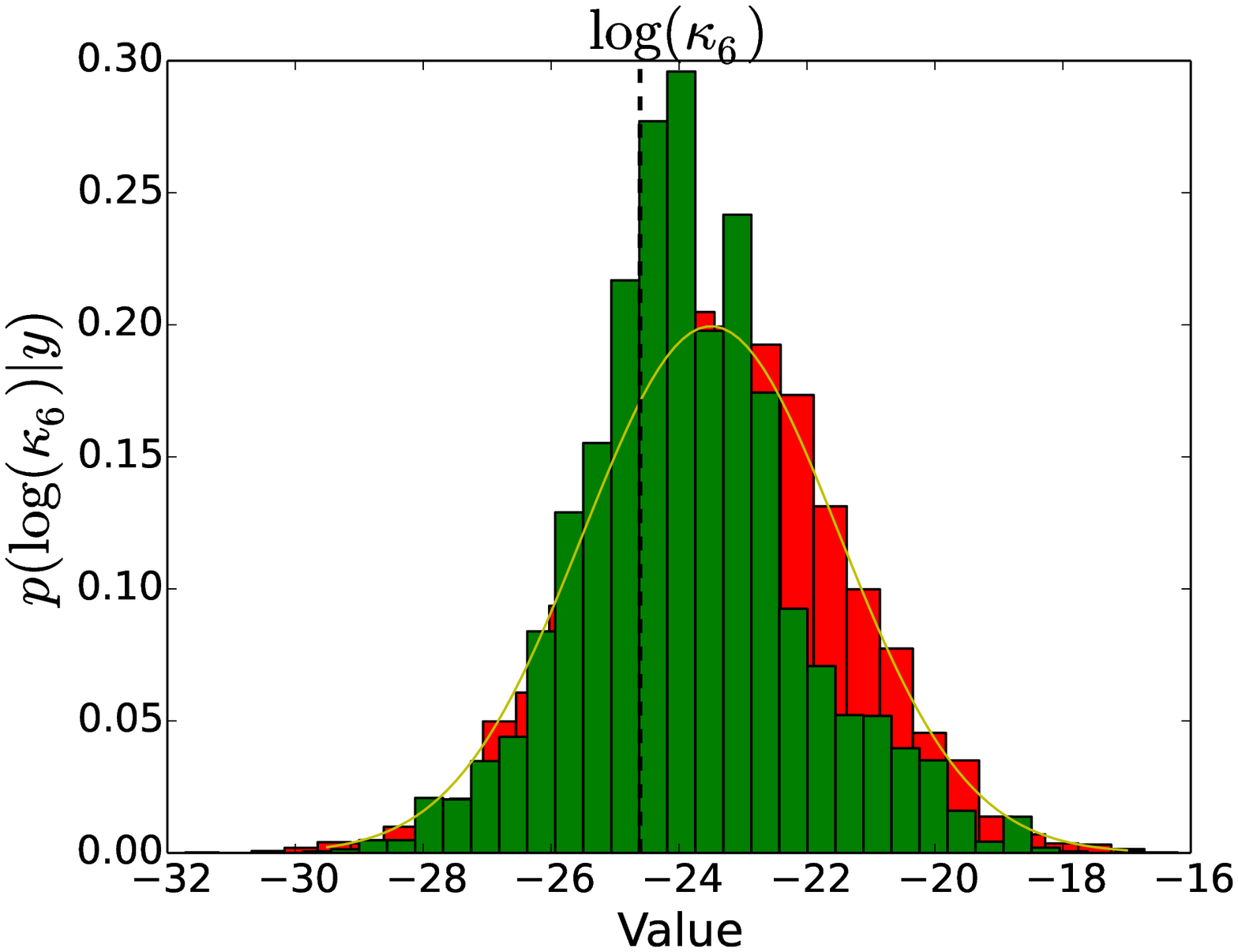}
\caption{Comparison of the histograms of the marginal posteriors
  of all $\log(\kappa_i)$, $i= 1, ..., 6$ for the designs $\bd_\circ$
  (green) and $\bd_\diamond$ (red). The red curve
  indicates their common prior distribution and the dashed black line
  indicates the true value used to generate the data. \label{fig:histo_1}}
\end{figure}

We generate $50000$ samples with a $10000$-sample burn-in period and we
retain a sample every $5$ steps.  After $8000$ samples are obtained, histograms of their
values are shown in Fig.~\ref{fig:histo_1}. Note that the
histograms display the marginal distribution of each $\log\kappa_i$,
$i = 1,..., 6$ which are no longer independent themselves, however
comparison of the histograms for each case together with their priors
and the true value gives us an idea about the different results
obtained for each design. As expected, we observe that $\bd_\circ$ provides
narrower posteriors that $\bd_\diamond$. This can be observed
particularly on $\log\kappa_2$ and $\log\kappa_5$ and in a smaller
scale on $\log\kappa_1$, $\log\kappa_3$ and  $\log\kappa_4$. The
posteriors of $\log\kappa_6$ for both designs do not display any
significant discrepancy from the prior but even in this case the one
corresponding to $\bd_\circ$ appear to be slightly decentralized
towards the true value. In addition, for $\bd_\diamond$ three posteriors,
namely $\log\kappa_2$,  $\log\kappa_5$ and $\log\kappa_6$ retain their
gaussian bell-shaped form and it seems that almost no new information has
been gained about their values. This is actually a consequence of the
fact that the corresponding soils (silty sand, sand with $10\%$ silt
and clay) are not present in the locations
where the data was collected from, for this design. Note that clay is
present in the lower layers of our domain and not much information is
gained about it from $\bd_\circ$ either. Another general characteristic
is that, even for the soils for which both designs provide similar posteriors, the maximum a
posteriori values of $\bd_\circ$ show excellent agreement with the true
values while those of $\bd_\diamond$ show a slight discrepancy which
makes them inappropriate for estimation. Overall, we conclude that the design
$\bd_\circ$ provides significantly better inference results that those
provided by $\bd_\diamond$ and this conclusion can be generalized for the comparison
of $\bd_\circ$ with any other design that achieves a lower $\hat{U}_L(\bd)$ value. 

\subsubsection{Case 2: Reference data from field measurements}

\begin{figure}[tbh]
\centering
\includegraphics[width = 6cm]{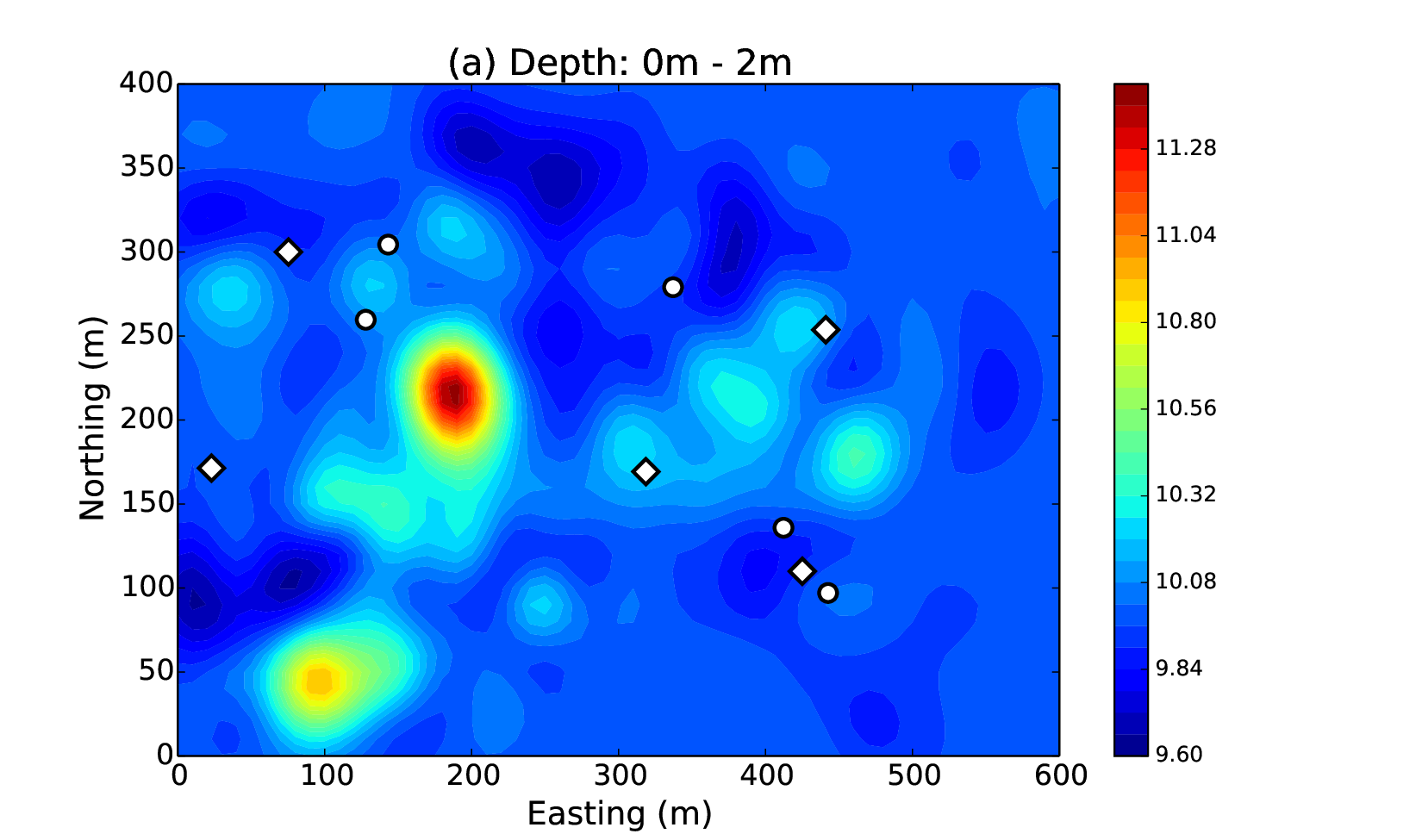}
\includegraphics[width = 6cm]{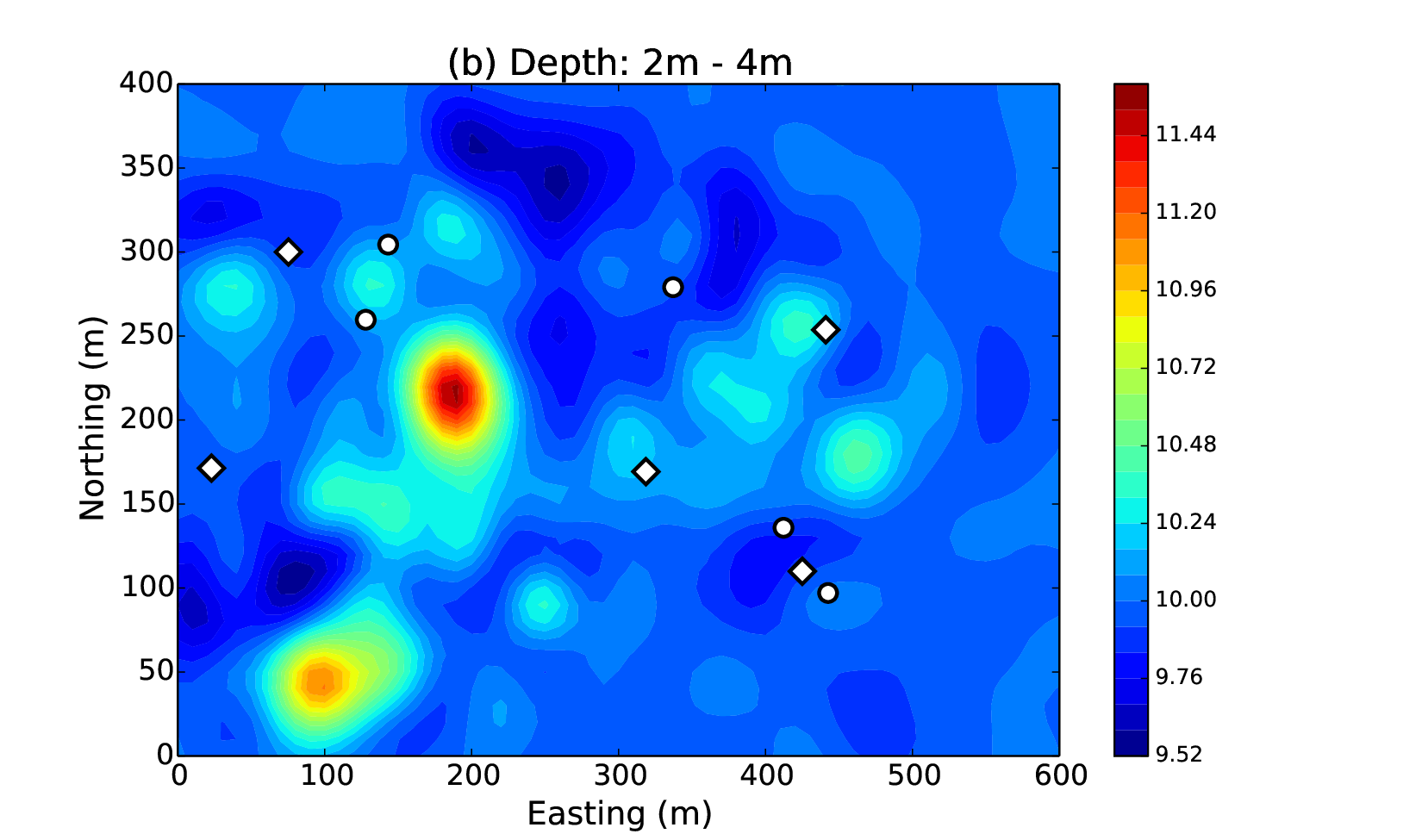}
\includegraphics[width = 6cm]{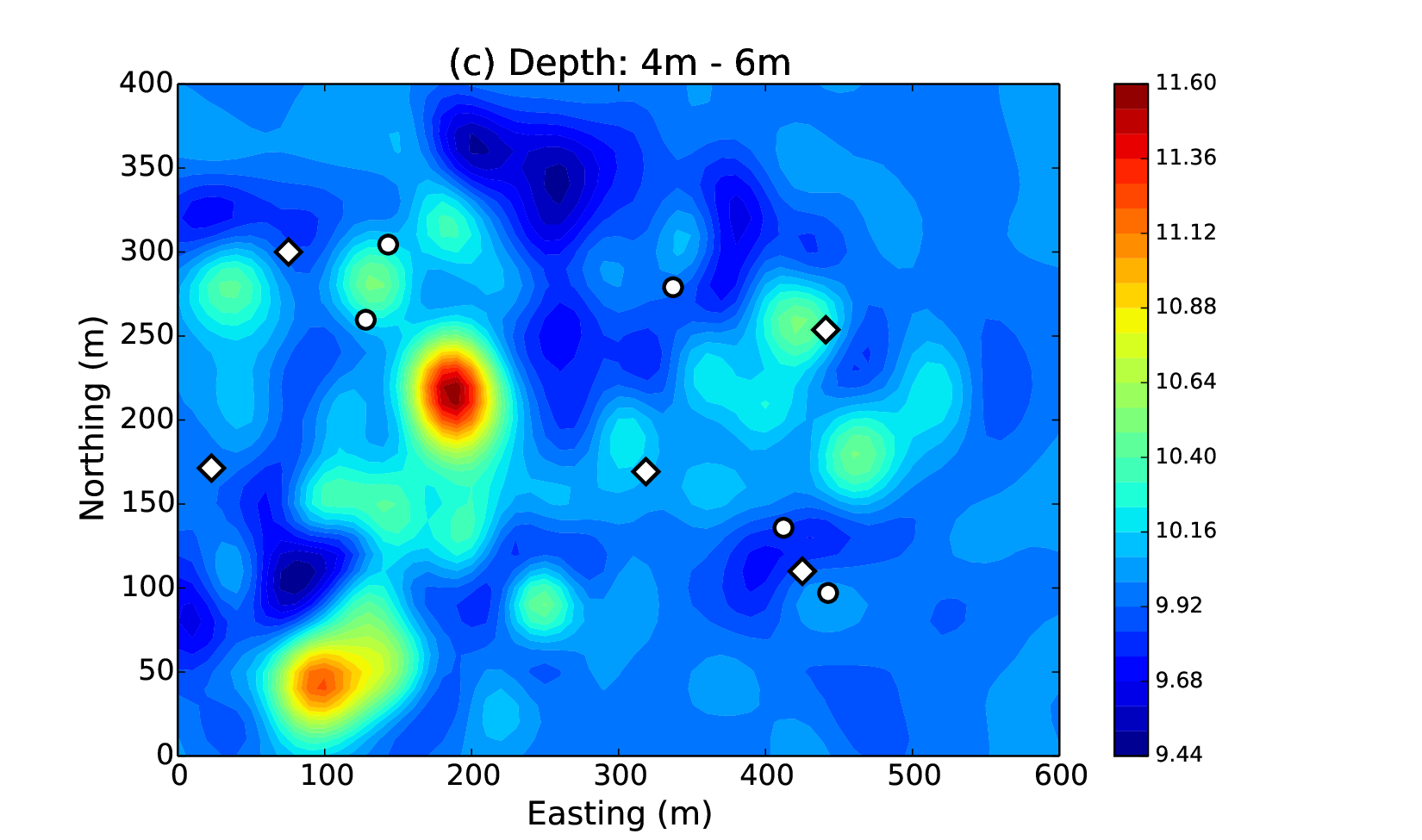}
\includegraphics[width = 6cm]{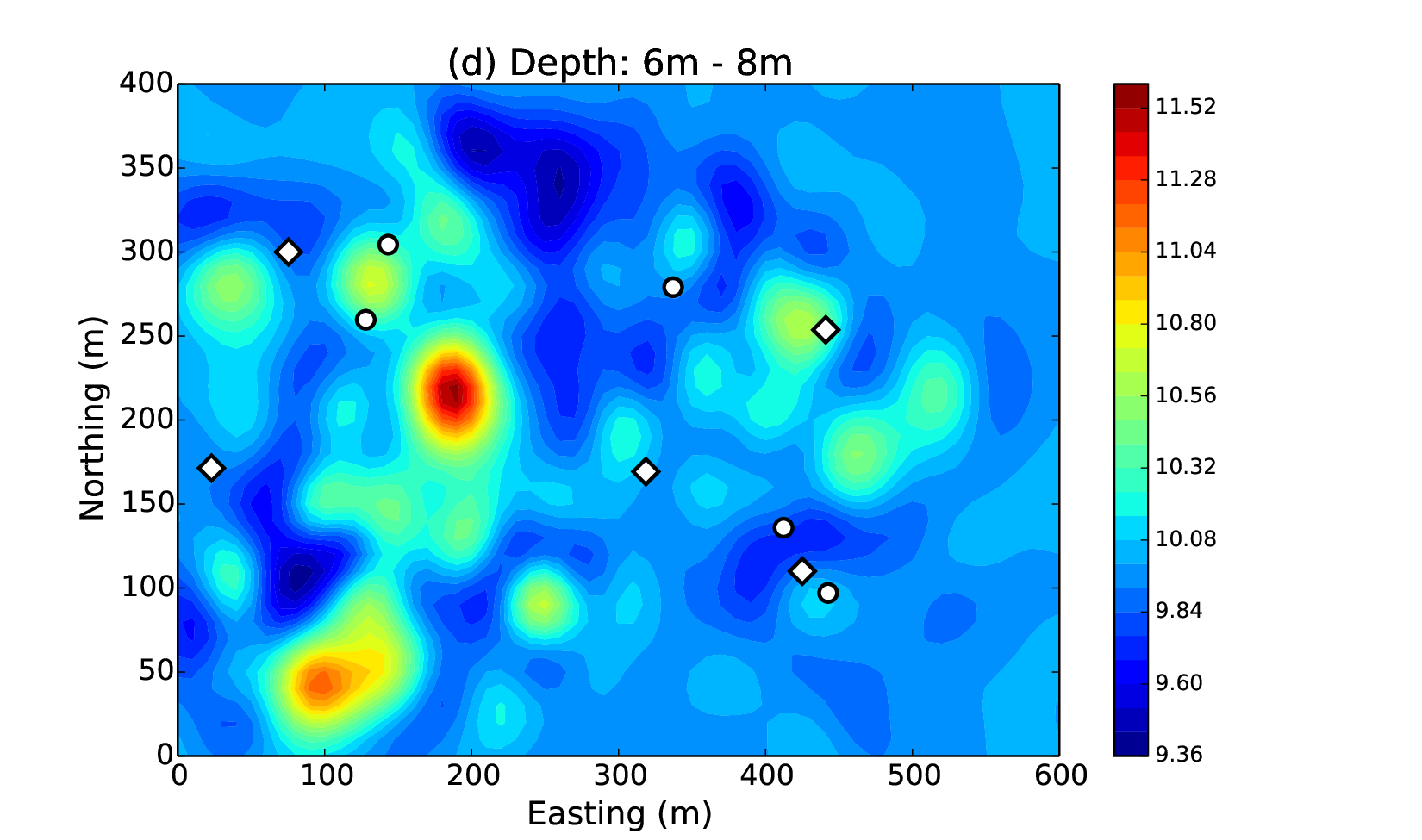}
\includegraphics[width = 6cm]{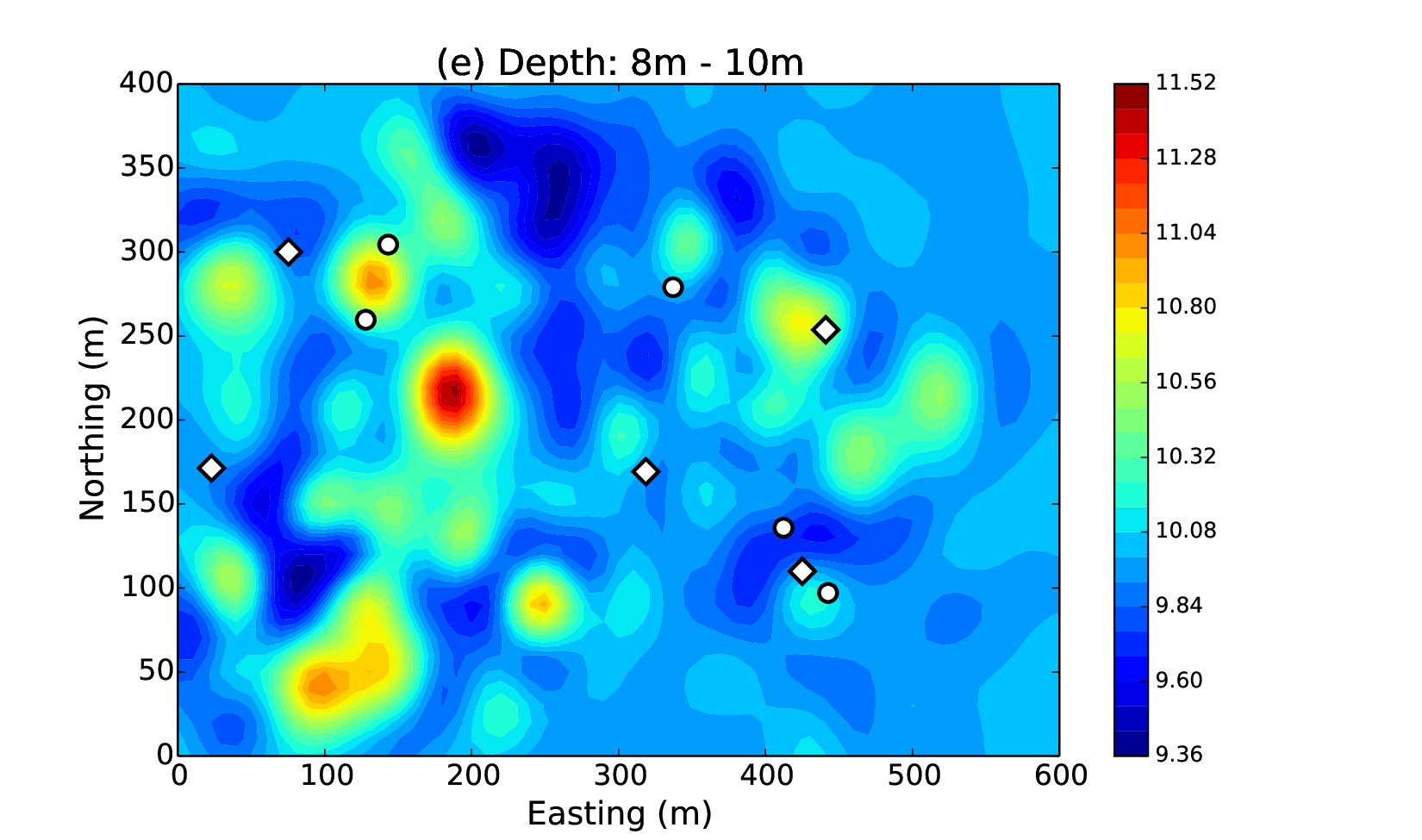}
\caption{Concentration map for the $5$ top layers of the domain obtained by GP
  regression on real data, used as our reference field in example
  2. The '$\circ$' and  '$\diamond$' signs indicate the locations of
 $\bd_\circ$ and $\bd_\diamond$ designs respectively, from where
 measurements were taken. \label{fig:data_2}}
\end{figure}

In this case, the data used in the likelihood
function which is incorporated in the acceptance probability through
Bayes' rule, is taken from a reference concentration field, that was generated by GP
regression based on measurements taken from the field data at the
specified locations as shown in the soil boring map in
Fig.~\ref{fig:world_map}. These measurements, denoted with $\bff$, consist of
$1240$ points and their locations form a $1240\times 3$ matrix $X$. Then, if
$\bff_{*}$ denotes the concentrations over the discretized domain, that is
$\bff_{*} \in \mathbb{R}^{24000}$, with corresponding locations $X_{*}$, then 
\begin{equation}
\bff_{*} | X_{*}, X, \bff \sim \calN\left(K(X_{*}, X)K(X,X)^{-1}\bff,
  K(X_{*}, X_{*}) - K(X_{*},X)K(X,X)^{-1}K(X,X_{*})\right),
\end{equation}
where $K(X_{*},X)$ is the covariance matrix with
$(K(X_{*},X))_{i,j} = \mathbf{k}(\bx_i,\bx_j)$, $\bx_i \in X_{*}$,
$\bx_j \in X$ and $\mathbf{k}(\cdot, \cdot)$ denotes some appropriately chosen
kernel (for more details on the derivation of this conditional
distribution one can see \cite{rasmussen}). In our case we chose a squared exponential kernel, given as 
\begin{equation}
\mathbf{k}(\bx, \bx') = \sigma^2\exp\left[ - \frac{1}{2}\sum_{n=1}^3\frac{(x_n - x'_n)^2}{\ell^2_n}\right]
\end{equation}
which was fit to the data with variance $\sigma^2 = 5$ and correlation lengths $(\ell_1, \ell_2,
\ell_3) = (30, 30, 10)$. Our reference field, consists of the mean
$\bar{\bff} = K(X_{*}, X)K(X,X)^{-1}\bff$ and the values of the $5$ top layers are shown
in Fig.~\ref{fig:data_2}.

\begin{figure}
\centering
\includegraphics[width = 5cm]{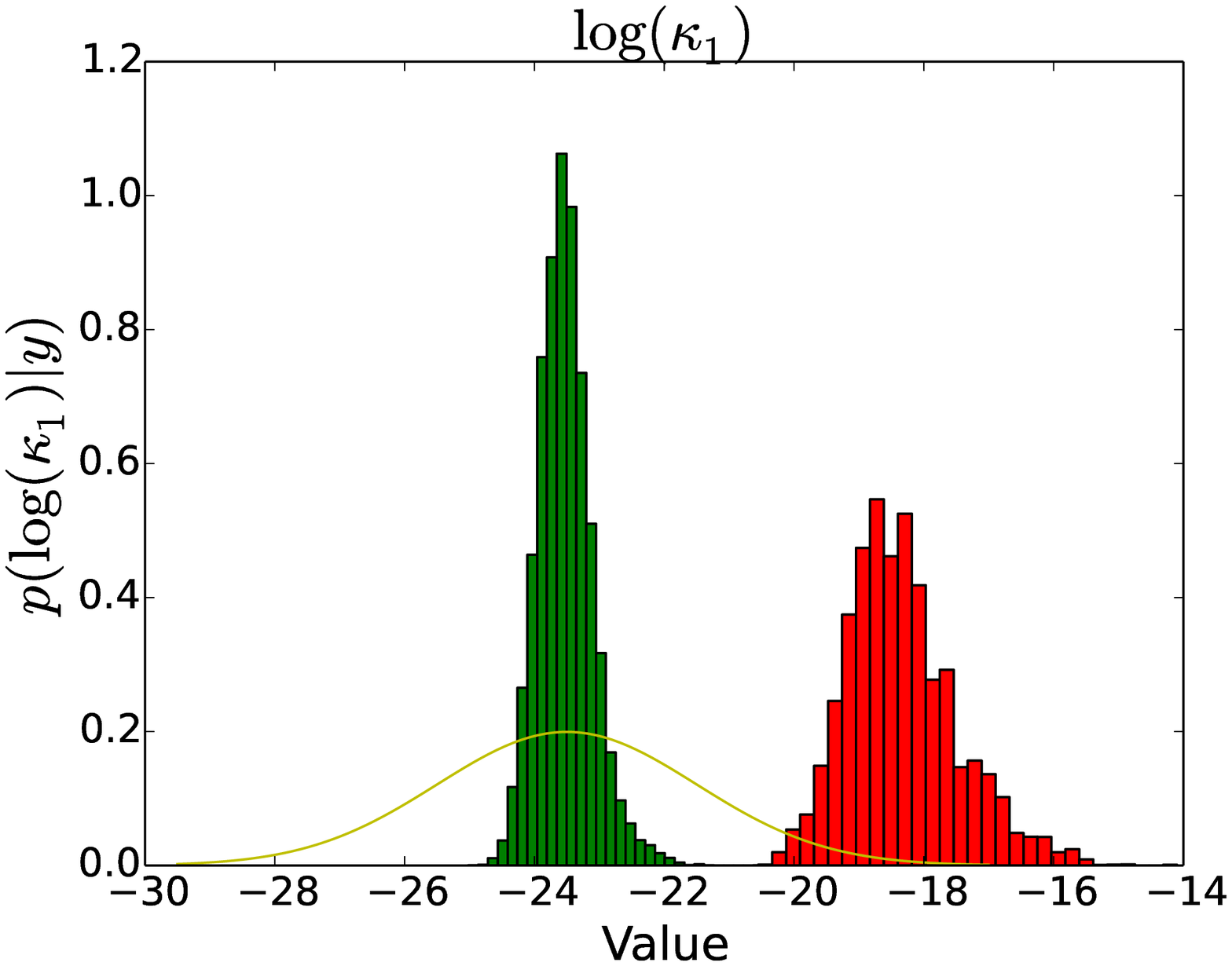}
\includegraphics[width = 5cm]{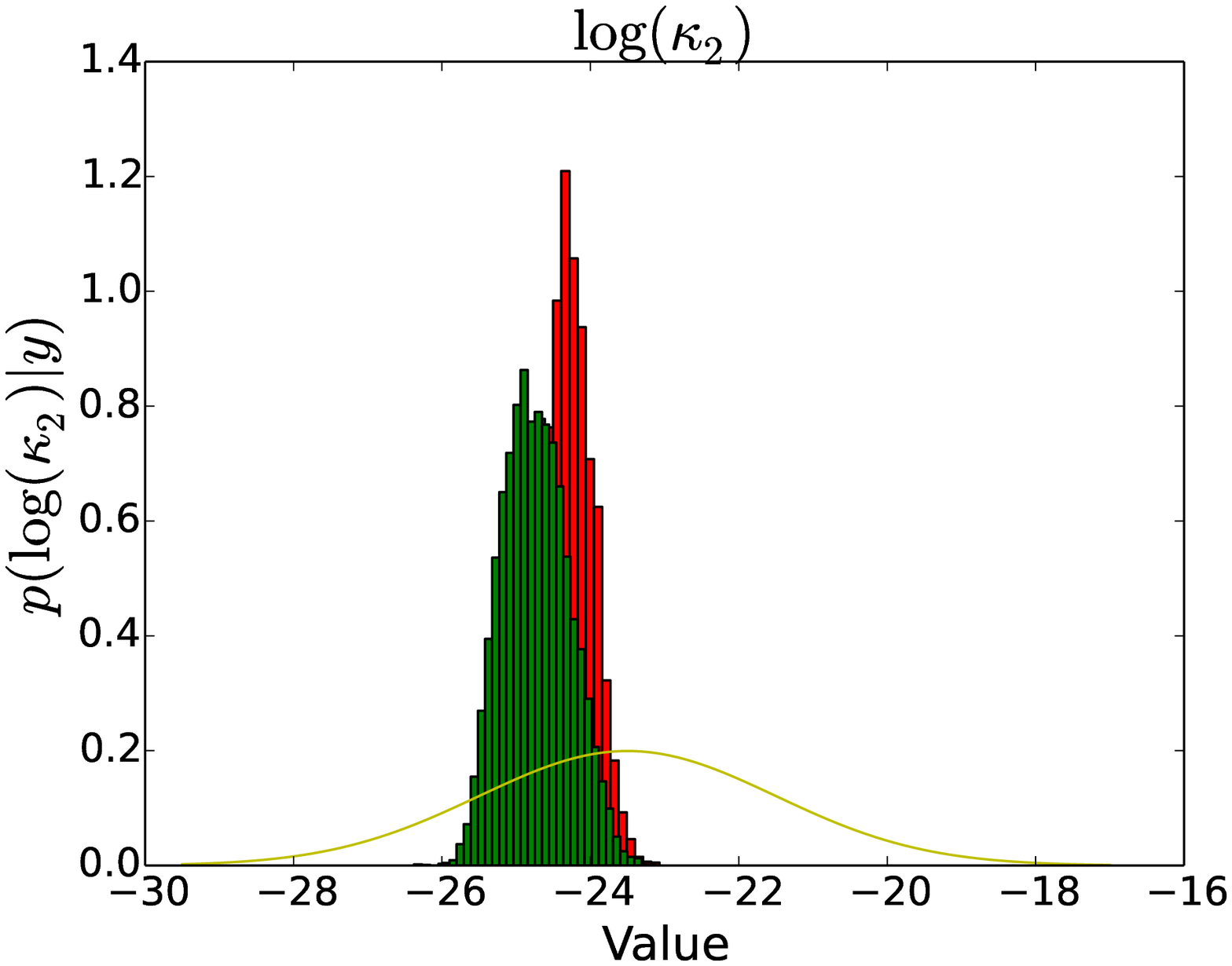}
\includegraphics[width = 5cm]{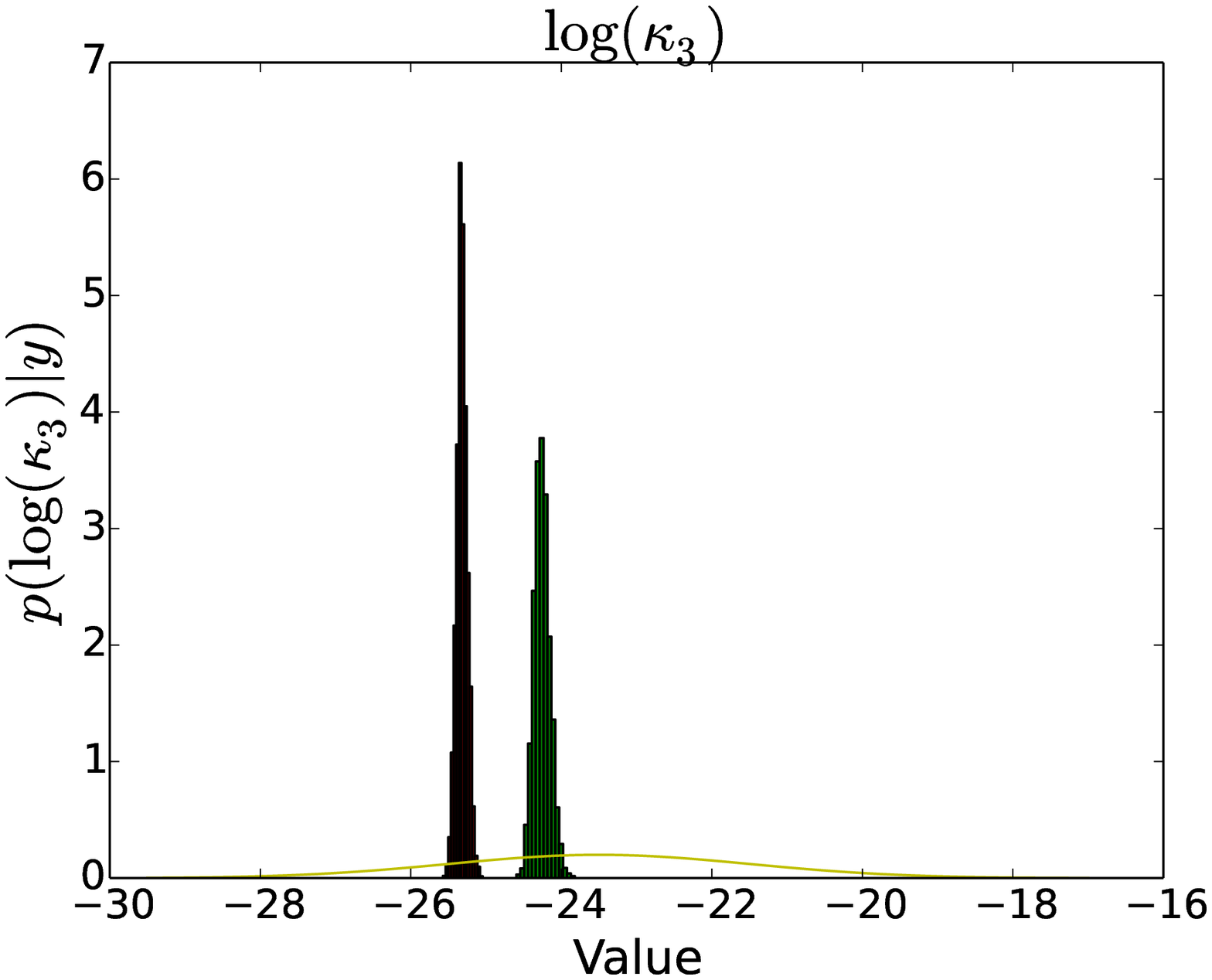}
\includegraphics[width = 5cm]{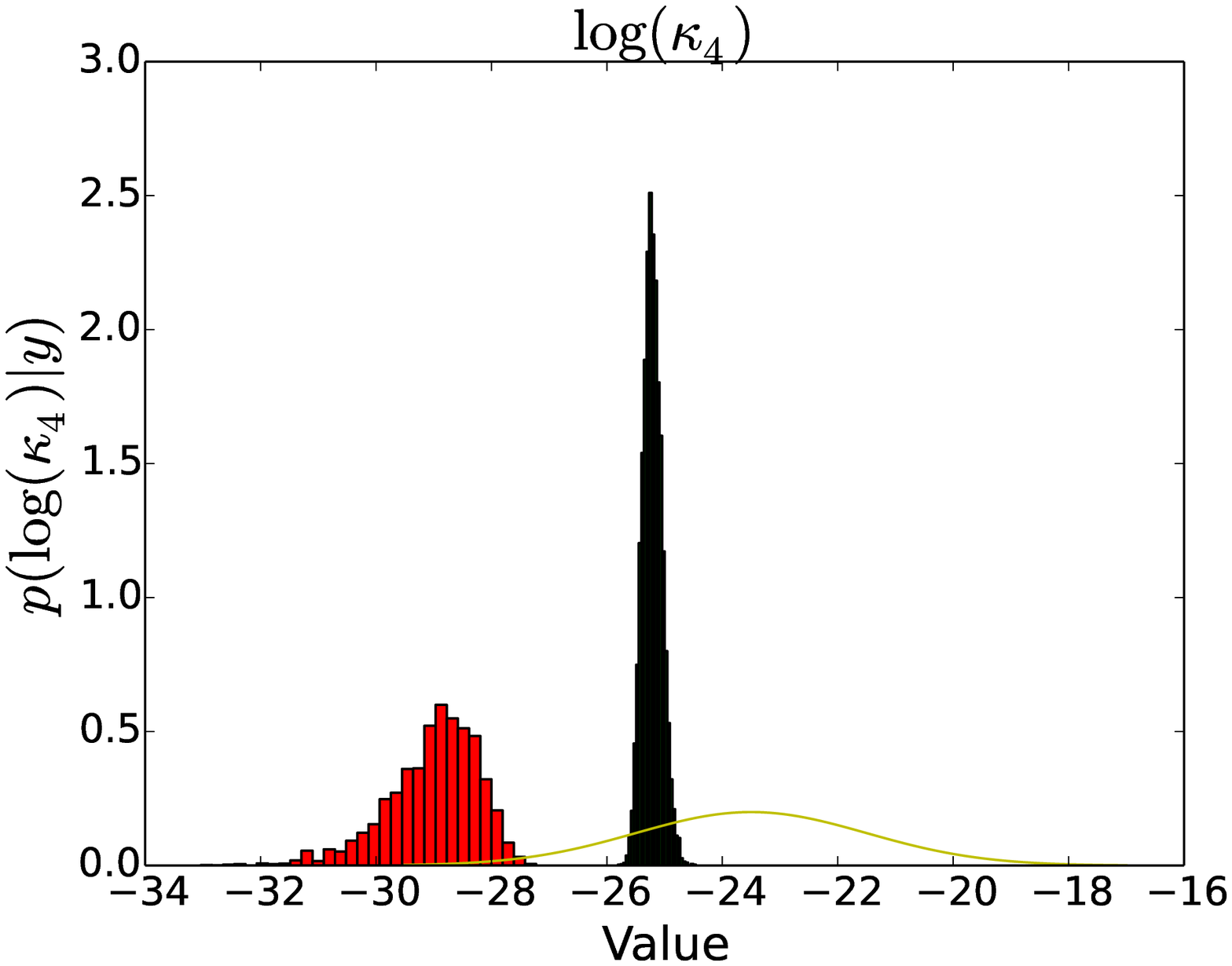}
\includegraphics[width = 5cm]{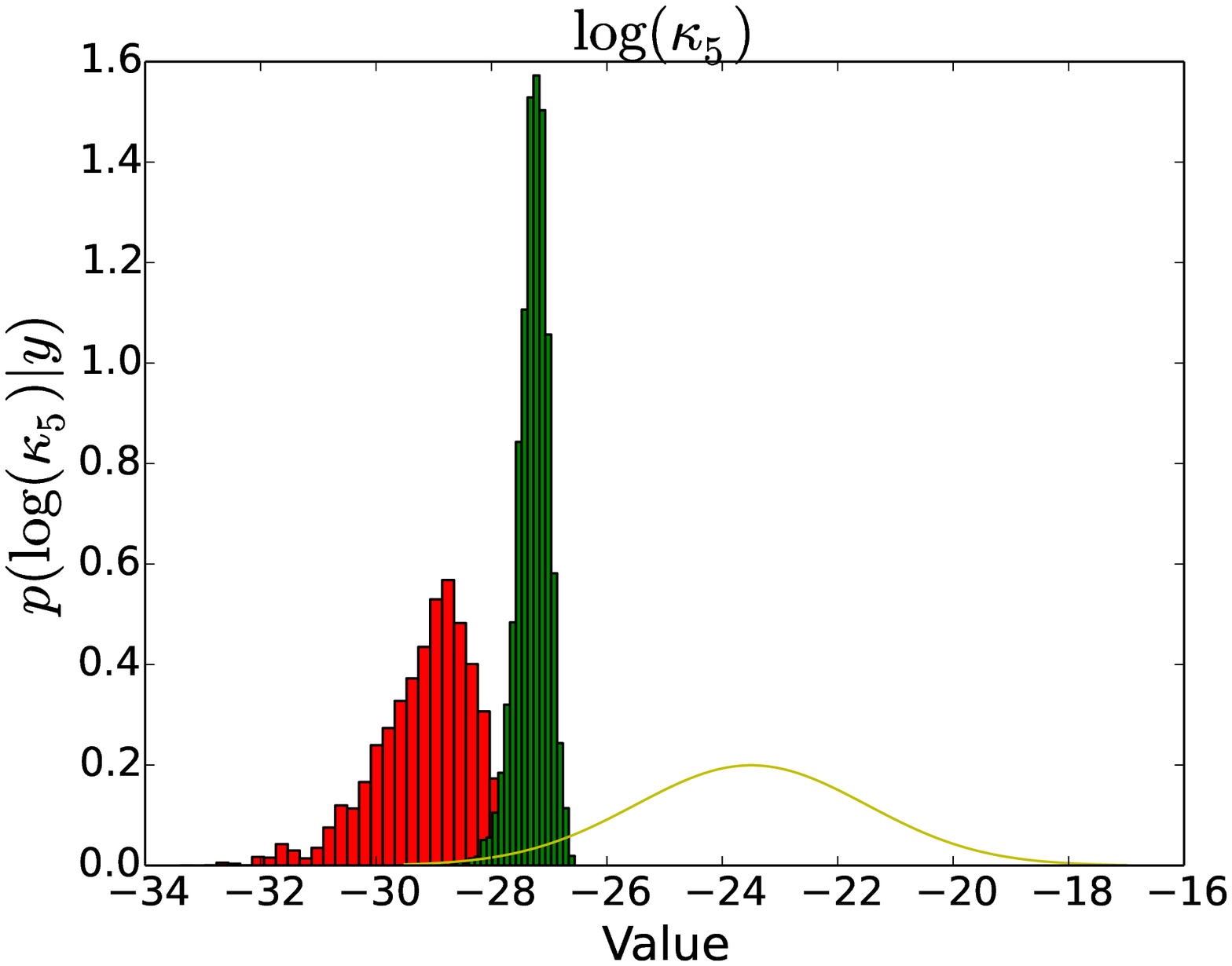}
\includegraphics[width = 5cm]{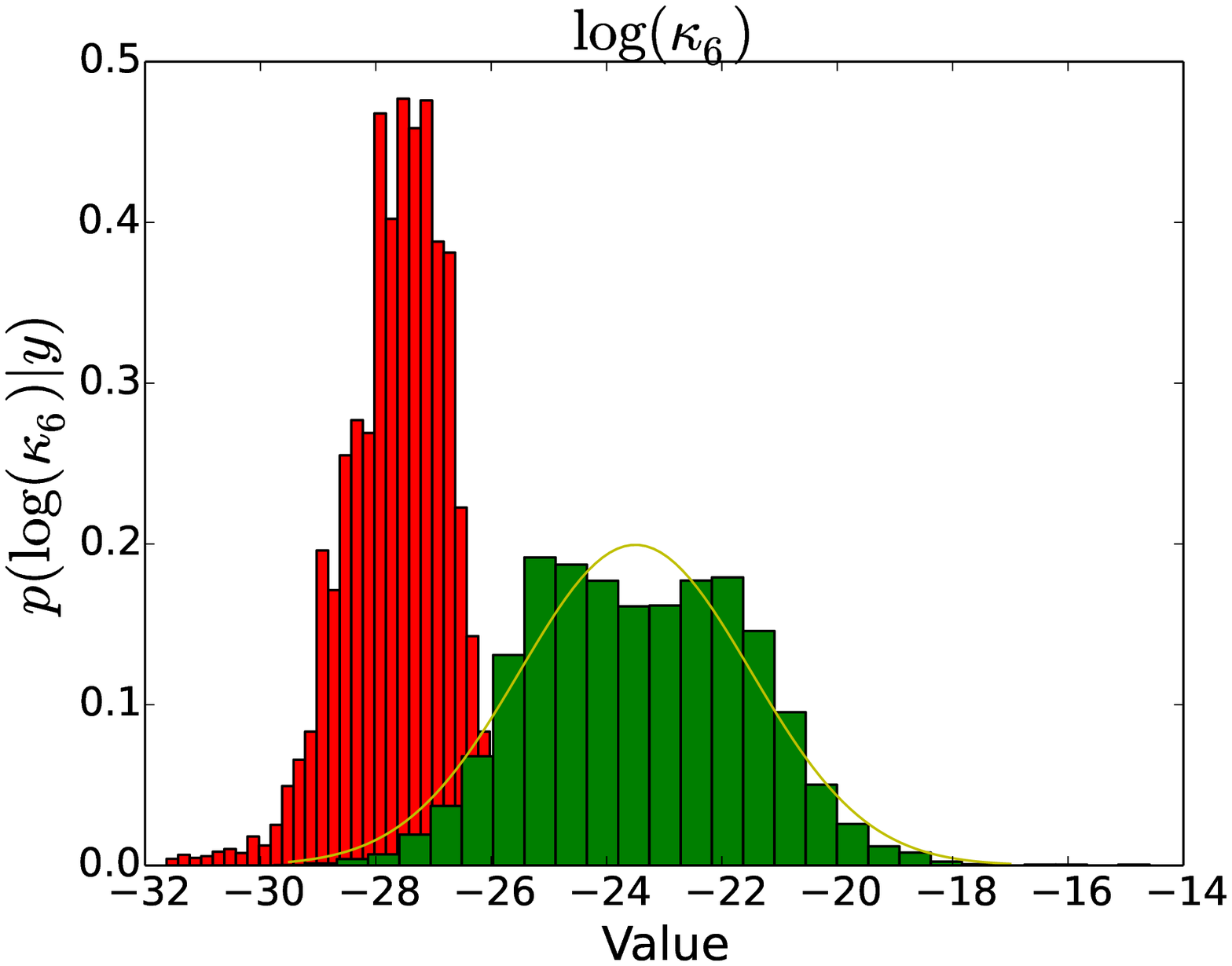}
\caption{Comparison of the histograms of the marginal posteriors
  of all $\log(\kappa_i)$, $i= 1, ..., 6$ for the designs $\bd_\circ$
  (green) and $\bd_\diamond$ (red). The red curve
  indicates their common prior distribution. \label{fig:histo_2}}
\end{figure}

Again, we set the posteriors with $p(\bk | \by, \bd_\circ)$ and
$p(\bk | \by, \bd_\diamond)$ as our target distributions and this time we
generate $100000$ samples with a $10000$-sample burn-in period and we
retain a sample every $5$ steps. After $18000$ samples are obtained,
histograms of the marginal distributions of each $\log\kappa_i$, $i = 1,...,6$, are shown in Fig.~\ref{fig:histo_2}. Again, we observe that $\bd_\circ$ provides
narrower posteriors than $\bd_\diamond$ for most of the cases. This can be observed
particularly on $\log\kappa_1$, $\log\kappa_4$ and $\log\kappa_5$. The
posterior of $\log\kappa_6$ for the optimal design is almost identical
to its prior which implies that nothing new has
been learned about its true value. As in the previous example, this is due
to the fact that the corresponding soil (clay) is not present in the locations
where the data was collected, for this design (recall that clay is
present in the lower layers of our domain). In this case, since our
data is obtained from a procedure other than evaluating our forward
model, it cannot be seen as a direct output of it and generally we do
not expect to observe uniqueness of the input parameters that can
potentially give rise to our measurements. This is reflected by the
fact that different designs give posteriors that seem to concentrate
around different values. Bayesian
inference however, still provides us with informative results as we observe the
posterior distributions to be much further away from the priors, thus
learning something about how close our model and our prior assumptions are
to the actual reality.

\section{Conclusions}
\label{sec:conclusions}

We presented an experimental design framework that focuses on
providing optimal solutions to the design problem in terms of
maximizing the information on model parameters from Bayesian
inference. This was achieved by employing an information theoretic
criterion, namely the expected relative entropy between the prior and
the posterior distributions of the unknown parameters. An additional
reformulation of the design criterion through the derivation of a
lower bound, was of great
importance in order to address biasedness issues as well as alleviate the
computational burden associated with optimization. 

The framework was applied to a real-world problem: The problem of
permeability identification in contaminated soils in the presence of
experimental field data. The forward
model in this case consisted of a system of differential equations
that describe contaminant transport in porous media, namely a
two-component, two-phase
flow-and-transport model that provides as its output the concentrations
of the pollutants at the time when data is collected. The
design parameters consist of multiple locations where the
concentrations will be measured. In the present setting, with the
analogous adopted code (TOUGH2) used as a black box simulator, no
derivative information was available to the optimization process. The
SPSA algorithm, in addition to addressing this issue also seemed to
accelerate convergence to the optimal solutions. 
To further reduce the computational cost, construction
of a model surrogate was of crucial importance since the
implementation of such a procedure would be impractical
otherwise. Finally, our methodology was validated after the
inference results showed that the data
collected at the optimal design locations were much more informative
than the data obtained from other arbitrarily chosen points, in the sense
that they resulted in much narrower posteriors, thus gaining more
knowledge about the true situation in the subsurface. We observe from
our analysis that, under limited resources, the performance of a
Bayesian update depends significantly on the location of data
acquisition. This highlights the need for optimal monitoring layouts
in order to manage environmental risks under economic constraints.

\appendix

\section{Comparison of the expected information gain estimates}
\label{sec:appendix_a}

\subsection{The double-loop Monte-Carlo estimate}
The expected information gain can be rearranged in the form 
\begin{equation*}
I(\bd) = \int_\calY \int_\calK \left\{ \log\left[p(\by | \bk,
    \bd)\right] - \log\left[ p(\by | \bd) \right] \right\} p(\by |
\bk, \bd) p(\bk) d\bk d\by
\end{equation*}
and can be evaluated by using Monte Carlo methods with 
\begin{equation*}
\hat{I}(\bd) = \frac{1}{N}\sum_{i = 1}^N \left\{ \log\left[ p(\by^{i} |
    \bk^{i}, \bd) \right]  - \log\left[ \frac{1}{M} \sum_{j=1}^M p(\by^i
    | \bk^{i,j}, \bd ) \right] \right\},
\end{equation*}
where $\{\bk^i\}$, $\{\bk^{ij}\}$, $i=1,..., N$, $j=1,..., M$ are i.i.d. samples from $p(\bk)$ and
$\{\by^{i}\}$ are i.i.d. samples from $p(\by | \bk^i, \bd)$. 

\subsection{Properties of the estimates}

The properties of $\hat{I}(\bd)$ are explored in \cite{ryan}. Precisely it
is shown that the variance is proportional to 
\begin{equation*}
var\left[\hat{I}(\bd) \right] \propto \frac{A(\bd)}{N}  +\frac{B(\bd)}{NM}
\end{equation*}
and the bias is proportional to 
\begin{equation*}
Bias\left[ \hat{I}(\bd) \right] \propto \frac{C(\bd)}{M},
\end{equation*}
where $A(\bd)$, $B(\bd)$ and $C(\bd)$ are terms that depend on the
likelihood and evidence function. Clearly, as mentioned above, $N$
controls the variance and $M$ controls the bias.  

For $\hat{U}^*_L(\bd)$ we have 
\begin{equation*}
var\left[\hat{U}^*_L(\bd) \right] = \frac{1}{NM} var\left[p(\by|
  \btheta, \bd)\right]
\end{equation*}
and the bias is trivially zero, whereas for $\hat{U}_L(\bd)$ we have
\begin{eqnarray*}
var\left[\hat{U}_L(\bd)\right] & = & var\left[\log
  \frac{1}{NM}\sum_{i,j = 1}^{N,M}p(\by^i|
  \btheta^{i,j}, \bd)\right] \\ & \approx & var
\left[ \log U^*_L(\bd) + \frac{\frac{1}{NM}\sum_{i,j =
        1}^{N,M}p(\by^i | \btheta^{i,j}, \bd) -
      U^*_L(\bd)}{U^*_L(\bd)} \right] \\
& = & var\left[ \frac{\frac{1}{NM}\sum_{i,j =
        1}^{N,M}p(\by^i | \btheta^{i,j}, \bd) -
      U^*_L(\bd)}{U^*_L(\bd)} \right] \\ & = &
\frac{1}{NM}\frac{var\left[p(\by | \btheta, \bd)\right]}{U^*_L(\bd)^2},
\end{eqnarray*}
where the second line follows after $1$st-order Taylor expansion about
$U^*_L(\bd)$ and 
\begin{eqnarray*}
Bias\left[\hat{U}_L(\bd)\right] & = & \E\left[\hat{U}_L(\bd) -
  U_L(\bd) \right] \\ & = & 
\E\left[ - \log \frac{1}{NM}\sum_{i,j =1}^{N,M} p(\by^i |
  \btheta^{i,j}, \bd) + \log U^*_L(\bd) \right] \\ & \approx &
\frac{1}{NM}\frac{var\left[p(\by | \btheta, \bd) \right]}{U^*_L(\bd)^2},
\end{eqnarray*}
where the third line follows after a $2$nd-order Taylor expansion.
Here the variance and the bias of the lower bound are
controlled by both $N$ and $M$.

\section{Governing equations for flow and transport in a porous
  medium}
\label{sec:appendix_b}

\subsection{Balance equations}

The general mass- and energy balance equations in a multicomponent (NK
components) multiphase problem are given as 
\begin{equation*}
\frac{d M^k}{dt} = \nabla \cdot \bF^{k} + q^k, \ \ k = 1,..., NK,
\end{equation*}
where $k$ is labeling the mass
component up to a total of $NK$, $M$ represents mass or energy per volume, $\bF$
represents mass or heat flux and $q$ denotes sinks or sources. In some
formulations, such as in TOUGH2, an integral expression is used in the form
\begin{equation*}
\frac{d}{dt} \int_{V_n}M^{k}dV_n =
\int_{\Gamma_n}\bF^{k}\cdot \bn d\Gamma_n +
\int_{V_n}q^{k}dV_n, \ \ k = 1, ..., NK
\end{equation*}
where $V_n$ is an arbitrary subdomain of the flow system and
$\Gamma_n$ its closed boundary surface, $\bn$
is a vector normal on the surface element $d\Gamma_n$ pointing inward
into $V_n$. If TOUGH2 is used for the numerical solution of the
governing equations, module EOS7r provides consistent characterization
of the constitutive equations, allowing a value of NK as high as 
$5$ (water, air, brine, parent radionuclide, daughter
radionuclide). In out problem, a zero source term is assumed resulting
in $q = 0$ and the contaminant sources in the model are taken to be
localized on the surface, at pre-assigned spatial locations shown in
Fig.~\ref{fig:sources}. In a TOUGH2 formalism, these sources are modeled as
injections in attached zero-volume blocks connected to the gridblocks
where the source is assumed to be.

The mass accumulation condition for the $k$th component is 
\begin{equation*}
M^{k} =  \phi \sum_{\beta}S_{\beta}\rho_{\beta}X_{\beta}^{k}
\end{equation*}
where the summation is taken over the fluid phases (liquid, gas,
non-aqueous phase liquids). Only liquid and gas phase are considered
in our case. The porosity is denoted by $\phi$, $S_{\beta}$ is the saturation of phase
$\beta$ (takes values from $0$ to $1$), $\rho_{\beta}$ is the density
of phase $\beta$ and $X_{\beta}^{k}$ is the mass fraction of
component $k$ present in phase $\beta$. 


The advective mass flux is given as 
\begin{equation*}
\bF^{k}_{adv} = \sum_{\beta}X_{\beta}^{k}\bF_{\beta}
\end{equation*}
and individual phase fluxed are given by a multiphase version of
Darcy's law:
\begin{equation*}
\bF_{\beta} = \rho_{\beta}\bu_{\beta} = - \kappa
\frac{\kappa_{r\beta}\rho_{\beta}}{\mu_{\beta}}(\nabla P_{\beta} - \rho_{\beta}\bg).
\end{equation*}
Here $u_{\beta}$ is the Darcy velocity (volume flux) in phase $\beta$,
$\kappa$ is the absolute permeability, $\kappa_{r\beta}$ is relative
permeability to phase $\beta$, $\mu_{\beta}$ is viscosity and 
\begin{equation*}
P_{\beta} = P + P_{c\beta} 
\end{equation*}
is the fluid pressure in phase $\beta$, which is the sum of the
pressure $P$ of a reference phase (usually taken to be the gas phase)
and the capillary pressure $P_{c\beta}$ ($\leq 0$), $\bg$ is the
vector of gravitational acceleration. Vapor pressure lowering due to
capillary and phase adsorption effects is modeled by Kelvin's equation 
\begin{equation*}
P_{\nu}(T, S_l) = f_{VPL}(T, S_l) P_{sat}(T)
\end{equation*}
where 
\begin{equation*}
f_{VPL} = \exp\left[ \frac{M_w P_{cl}(S_l)}{\rho_l R(T + 273.15)}\right]
\end{equation*} 
is the vapor pressure lowering factor. Phase adsorption is neglected. The saturated
vapor pressure of bulk aqueous phase is denoted by $P_{sat}$, $P_{cl}$ is the difference
between aqueous and gas phase pressures, $M_w$ is the molecular weight
of water and $R$ is the universal gas constant. Temperature $T$ is
assumed constant.

The absolute permeability of the gas phase increases at low pressures
according to the relation given by Klinkenberg
\begin{equation*}
\kappa = \kappa_{\infty}\left(1 + \frac{b}{P}\right)
\end{equation*}
where $\kappa_{\infty}$ is the permeability at "infinite" pressure and $b$
is the Klinkenberg parameter. In addition to Darcy's flow, mass flux
occurs by molecular diffusion according to
\begin{equation*}
\bff^k_{\beta} = - \phi\tau_0\tau_{\beta}\rho_{\beta}d_{\beta}^k
\nabla X_{\beta}^k.
\end{equation*}
Here $d_{\beta}^k$ is the molecular diffusion coefficient for
component $k$ in phase $\beta$, $\tau_0\tau_{\beta}$ is the tortuosity
which includes a porous medium dependent factor $\tau_0$ and a
coefficient that depends on phase saturation $S_{\beta}$,
$\tau_{\beta} = \tau_{\beta}(S_{\beta})$. Finally, the total mass flux is
finally given by 
\begin{equation*}
\bF^k = \bF_{abd} + \sum_{\beta}\bff_{\beta}^k.
\end{equation*}

\subsection{Relative permeability model}

The relative permeability is assumed to follow the van Genuchten-Mualem model
which for liquid is
\begin{eqnarray*}
\kappa_{rl} = \left\{\begin{array}{lc} \sqrt{S^*}\left[ 1 - \left(1 -
        S^{* 1 / \lambda} \right)^{\lambda}\right]^2, &  S_l <
    S_{ls}\\ 1, &  S_l \geq S_{ls} \end{array} \right.
\end{eqnarray*}
and for gas is 
\begin{eqnarray*}
\kappa_{rg} =  \left\{\begin{array}{lc} 1 - \kappa_{rl}, &
    if\ S_{gr} = 0 \\ \left(1 -
      \hat{S}\right)^2\left(1 - \hat{S}^2\right), & if \ S_{gr} >
    0 \end{array} \right.
\end{eqnarray*}
subject to restriction $0 \leq \kappa_{rl}, \kappa_{rg} \leq 1$. Here
$S^* = (S_l - S_{lr}) / (S_{ls} - S_{lr})$ and $\hat{S} = (S_l -
S_{lr})(1 - S_{lr} - S_{gr})$.

\subsection{Capillary pressure model}

For the capillary pressure we use the van Genuchten function
given as 
\begin{equation*}
P_{cap} = - P_0\left(S^{* -1/\lambda} - 1\right)^{1- \lambda}
\end{equation*}
subject to restrictions $-P_{\max} \leq P_{cap} \leq 0$.
Again $S^*$ is defined as for the relative permeability. 

\subsection{Parameters}

Table~\ref{tab:tough2_params} displays the nominal values assigned to
all parameters required according to the above governing
equations. It is important also to note the following for our
simulation:\\
1. The mobilities are upstream weighted. \\
2. The permeabilities are harmonic weighted. \\
3. Module EOS7r currently neglects brine and daughter radionuclide,
resulting in a 3-component flow. \\
4. Adsorption effects are neglected. \\
5. Biodegradation effects are neglected.

\bibliographystyle{plain}
\bibliography{references}

\begin{thebibliography}{10}

\bibitem{aquaveo}
Aquaveo, L.L.C., 2007.
\newblock {\em Groundwater Modeling System Version 6.5.6, build date, May 27,
  2009, UT, USA}.

\bibitem{atkinson}
A.C. Atkinson and A.N. Donev.
\newblock {\em Optimum experimental design with SAS}.
\newblock Oxford Statistical Science Series, Oxford University Press, 2007.

\bibitem{bau}
D.A. Ba\'{u} and A.S. Mayer.
\newblock Optimal design for pump-and-treat under uncertain hydraulic
  conductivity and plume distribution.
\newblock {\em Journal of Contaminant Hydrology}, 100:30--46, 2008.

\bibitem{chaloner}
K.~Chaloner and I.~Verdinelli.
\newblock Bayesian experimental design: a review.
\newblock {\em Statistical Science}, 10:273--304, 1995.

\bibitem{cohen}
A.~Cohen, M.A. Davenport, and D.~Leviatan.
\newblock On the stability and accuracy of leats squares approximations.
\newblock {\em Foundations of Computational Mathematics}, 13:819--834, 2013.

\bibitem{fedorov}
V.V. Fedorov.
\newblock {\em Theory of optimal experiments}.
\newblock Academic Press, New York, 1972.

\bibitem{freeze}
R.A. Freeze, B.~James, J.~Massmann, T.~Sperling, and L.~Smith.
\newblock Hydrogeological decision analysis, 4: The concept of data worth and
  its use in the development of site investigation strategies.
\newblock {\em Ground Water}, 30:574--588, 1992.

\bibitem{ghanem_sfem}
R.G. Ghanem and P.D. Spanos.
\newblock {\em Stochastic finite elements: A spectral approach, revised
  edition}.
\newblock Dover Publications Inc, 2012.

\bibitem{ghiocel}
D.M. Ghiocel and R.G. Ghanem.
\newblock Stochastic finite element analysis of seismic soil structure
  interaction.
\newblock {\em Journal of Engineering Mechanics}, 128:66--77, 2002.

\bibitem{haario}
H.~Haario, E.~Saksman, and J.~Tamminen.
\newblock An adaptive metropolis algorithm.
\newblock {\em Bernoulli}, 7:223--242, 2001.

\bibitem{Haddad-2013}
H.~Haddad-Zadegan, R.~Ghanem, and P.~Hajali.
\newblock Data worth analysis in spatial prediction and soil remediation.
\newblock In G.~Deodatis and B.~Ellingwood, editors, {\em ICOSSAR'13:
  International Conference on Structural Safety and Reliability}, Columbia
  University, June 16-20 2013.

\bibitem{doostan}
J.~Hampton and A.~Doostan.
\newblock Coherence motivated sampling and convergence analysis of
  least-squares polynomial chaos regression.
\newblock {\em arXiv preprint, arXiv:1410.1931}, 2014.

\bibitem{hastings}
W.K. Hastings.
\newblock Monte carlo sampling methods using markov chains and their
  applications.
\newblock {\em Biometrika}, 57:97--109, 1970.

\bibitem{hosder}
S.~Hosder, R.W. Walters, and M~Balch.
\newblock Efficient sampling for non-intrusive polynomial chaos applications
  with multiple uncertain input variables.
\newblock In {\em 9th AIAA Non-Deterministic Approaches Conference}, 2007.

\bibitem{marzouk_simul}
X.~Huan and Y.~Marzouk.
\newblock Simulation-based optimal bayesian experimental design for nonlinear
  systems.
\newblock {\em Journal of Computational Physics}, pages 288--317, 2013.

\bibitem{marzouk_gradient}
X.~Huan and Y.~Marzouk.
\newblock Gradient-based stochastic optimization methods in bayesian
  experimental design.
\newblock {\em International Journal for Uncertainty Quantification},
  4:479--510, 2014.

\bibitem{james}
B.R. James and R.A. Freeze.
\newblock The worth of data in predicting aquitard continuity in
  hydrogeological design.
\newblock {\em Water Resources Research}, 29:2049--2065, 1993.

\bibitem{gorelick}
B.R. James and S.M. Gorelick.
\newblock When enough is enough: The worth of monitoring data in aquifer
  remediation design.
\newblock {\em Water Resources Research}, 30:3499--3513, 1994.

\bibitem{kiefer}
J.~Kiefer and J.~Wolfowitz.
\newblock Stochastic estimation of a regression function.
\newblock {\em The Annals of Mathematical Statistics}, 23:462--466, 1952.

\bibitem{kullback}
S.~Kullback and R.A. Leibler.
\newblock On information and sufficiency.
\newblock {\em The Annals of Mathematical Statistics}, 22:79--86, 1951.

\bibitem{li}
H.~Li and D.~Zhang.
\newblock Probabilistic collocation method for flow in porous media:
  Comparisons with other stochastic methods.
\newblock {\em Water Resources Research}, 43:44--56, 2007.

\bibitem{lindley}
D.~V. Lindley.
\newblock On a measure of the information provided by an experiment.
\newblock {\em The Annals of Mathematical Statistics}, 27:986--1005, 1956.

\bibitem{long_quan}
Q.~Long, M.~Scavino, R.~Tempone, and S.~Wang.
\newblock Fast estimation of expected information gains for bayesian
  experimental designs based on laplace approximations.
\newblock {\em Computational Methods in Applied Mechanics and Engineering},
  259, 2013.

\bibitem{massmann}
J.~Massmann and R.A. Freeze.
\newblock Groundwater contamination from waste management sites: The
  interaction between risk-based engineering design and regulatory policy:1.
  methodology.
\newblock {\em Water Resources Research}, 23:351--367, 1987.

\bibitem{metropolis}
N.~Metropolis, A.W. Rosenbluth, M.N. Rosenbluth, A.H. Teller, and E.~Teller.
\newblock Equations of state calculations by fast computing machines.
\newblock {\em Journal of Chemical Physics}, 21:1087--1091, 1953.

\bibitem{norberg}
T.~Norberg and L.~Ros\'{e}n.
\newblock Calculating the optimal number of contaminant samples by means of
  data worth analysis.
\newblock {\em Environmetrics}, 17:705--719, 2006.

\bibitem{olden}
C.~Oldenburg and K.~Pruess.
\newblock {\em EOS7R: Radionuclide transport for TOUGH2}.
\newblock Berkeley, California, November 1995.
\newblock Report LBL-34868.

\bibitem{pruess}
K.~Pruess, C.~Oldenburg, and G.~Moridis.
\newblock {\em TOUGH2 User's guide, Version 2}.
\newblock Berkeley, California, 1999.
\newblock Report LBNL-43134.

\bibitem{rasmussen}
C.E. Rasmussen and C.K.I. Williams.
\newblock {\em Gaussian processes for machine learning}.
\newblock The MIT Press, 2006.

\bibitem{monro}
H.~Robbins and S.~Monro.
\newblock A stochastic approximation method.
\newblock {\em The Annals of Mathematical Statistics}, 22:400--407, 1951.

\bibitem{ryan}
K.J. Ryan.
\newblock Estimating expected information gains for experimental designs with
  application to the random fatigue-limit model.
\newblock {\em Journal of Computational and Graphical Statistics}, 12:585--603,
  2003.

\bibitem{sebastiani}
P.~Sebastiani and H.P. Wynn.
\newblock Maximum entropy sampling and optimal bayesian experimental design.
\newblock {\em Journal of the Royal Statistical Society. Series B: Statistical
  Methodology}, 62:145--157, 2000.

\bibitem{seber}
G.A.F. Seber and A.J. Lee.
\newblock {\em Linear regression analysis, second edition}.
\newblock John Wiley and Sons, 2003.

\bibitem{ghanem_soize}
C.~Soize and R.~Ghanem.
\newblock Physical systems with random uncertainties: Chaos representations
  with arbitrary probability measure.
\newblock {\em SIAM Journal of Scientific Computing}, 26(2):395--410, 2004.

\bibitem{spall_alg}
J.C. Spall.
\newblock Multivariate stochastic approximation using a simultaneous
  perturbation gradient approximation.
\newblock {\em IEEE Transactions on Automatic Control}, 37, 1992.

\bibitem{spall_cookbook}
J.C. Spall.
\newblock Implementation of the simultaneous perturbation algorithm for
  stochastic optimization.
\newblock {\em IEEE Transactions on Aerospace and Electronic Systems}, 34,
  1998.

\bibitem{vandenberg}
J.~van~den Berg, A.~Curtis, and J.~Trampert.
\newblock Optimal nonlinear bayesian experimental design: an application to
  amplitude versus offset experiments.
\newblock {\em Geophysical Journal International}, 155:411--421, 2003.

\bibitem{xiu}
D.~Xiu and G.E. Karniadakis.
\newblock The wiener-askey polynomial chaos for stochastic differential
  equations.
\newblock {\em SIAM Journal of Scientific Computing}, 24:619--644, 2002.

\bibitem{xiu_nonintrusive}
D.~Xiu and G.E. Karniadakis.
\newblock Modeling uncertainty in flow simulations via generalized polynomial
  chaos.
\newblock {\em Journal of Computational Physics}, 187:137--167, 2003.

\bibitem{pruess_mp}
K.~Zhang, Y.S. Wu, and K.~Pruess.
\newblock {\em User's guide for TOUGH2-MP - A massively parallel version of the
  TOUGH2 code}.
\newblock Berkeley, California, 2008.
\newblock Report LBNL-315E.

\bibitem{sahinidis}
Y.~Zhang and N.V. Sahinidis.
\newblock Uncertainty quantification in co$_2$ sequestration using surrogate
  models from polynomial chaos expansion.
\newblock {\em Industrial \& Engineering Chemistry Research}, 52:3121--3132,
  2012.

\end{thebibliography}




\end{document}